\pdfoutput=1
\documentclass[11pt]{article}

\usepackage[a4paper,margin=2cm]{geometry}

\usepackage[T1]{fontenc}
\usepackage[utf8]{inputenc}
\usepackage{amsmath,amssymb}
\usepackage{graphicx}
\usepackage{booktabs}
\usepackage{array}
\usepackage[hidelinks]{hyperref}
\hypersetup{pdftitle={An extended deep energy method for thermo-mechanical crack propagation},
            pdfauthor={Han Zhang, Mehrisadat Makki Alamdari, Babak Shahbodagh, Mohammad Vahab, Cosmin Anitescu, Timon Rabczuk, Elena Atroshchenko}}
\usepackage{placeins}

\usepackage{caption}
\usepackage{authblk}

\usepackage{pgfplots}
\pgfplotsset{compat=1.18}
\usetikzlibrary{arrows.meta,positioning,shapes.geometric,calc}

\title{An extended deep energy method for thermo-mechanical crack propagation}

\author[a]{Han Zhang\thanks{E-mail: han.zhang7@unsw.edu.au}}
\author[a]{Mehrisadat Makki Alamdari}
\author[a]{Babak Shahbodagh}
\author[b]{Mohammad Vahab}
\author[c]{Cosmin Anitescu}
\author[c]{Timon Rabczuk}
\author[a]{Elena Atroshchenko\thanks{Corresponding author. E-mail: e.atroshchenko@unsw.edu.au}}
\affil[a]{University of New South Wales, Sydney, NSW, Australia}
\affil[b]{Central Queensland University, Melbourne, VIC, Australia}
\affil[c]{Bauhaus-Universit\"{a}t Weimar, Weimar, Germany}
\date{}

\begin{document}

\maketitle

\begin{abstract}
Thermo-mechanical fracture couples transient heat conduction on a cracked domain with a crack that grows as the temperature and the displacement evolve. Recently, neural energy solvers have been proposed for phase-field modeling of fracture and later extended to represent a sharp crack through the network input. However, heat conduction on the cracked domain and crack propagation under the resulting thermal stresses have not yet been treated together in these solvers. We present an extended deep energy method for thermo-mechanical crack propagation in which the crack remains a sharp polyline throughout. The temperature and the displacement are represented by two networks that receive the crack through a scalar embedding function, discontinuous across the crack and smooth elsewhere. Both fields can therefore jump across the crack without a regularization length, and the displacement is further enriched near the tip by the Williams expansion with trainable amplitudes. The temperature is obtained by minimizing an incremental functional of transient heat conduction and the displacement by minimizing the thermoelastic potential energy, in a staggered sequence at every load step. The energies are estimated by Monte Carlo integration on points that are stratified over square background elements, densified near the tip and redrawn during training, which removes the spurious solutions that a fixed point set admits. Crack growth is decided by classical fracture mechanics criteria evaluated on the trained fields. The stress intensity factors (SIFs) are extracted by the interaction integral, which under thermal load includes the area term of Wilson and Yu, and each extraction is checked by a sweep of the contour radius before the crack may advance. The crack advances at the maximum hoop stress angle when the energy release rate of a kink in that direction reaches the critical value at the crack-tip temperature. The method is assessed on a stationary thermal edge crack, on single-edge notched tension and shear tests under thermal load, the latter with a functionally graded material, and on a notched cruciform specimen. The extracted SIF of the stationary crack agrees with the published value to $0.11\%$, initiation in the shear test agrees with an independent sharp-crack finite element solution to within one load step, and the predicted crack paths of the cruciform follow the published solutions under mechanical, thermal and combined loading.
\end{abstract}


\clearpage

\section{Introduction}
\label{sec:introduction}

Fracture under combined thermal and mechanical loading is a common failure mode in engineering practice, in the thermal shock of ceramics, in hot cracking during additive manufacturing~\cite{ruan2023thermomechanical}, and in thermally loaded components of nuclear, aerospace and energy systems. A temperature gradient alone can generate stresses large enough to initiate and drive a crack, and the resulting paths are often curved and hard to anticipate. In quenching experiments on thin ceramic plates the crack patterns are periodic and hierarchical, and their geometry depends on the severity of the thermal load~\cite{shao2010}. Predicting both the structural response and the crack path under such loading is therefore a standing task of computational fracture mechanics.

The theory of brittle fracture rests on the energy balance of Griffith~\cite{griffith1921}, the stress intensity factors (SIFs) of Irwin~\cite{irwin1957} and the asymptotic crack-tip fields of Williams~\cite{williams1957}. The $J$-integral of Rice~\cite{rice1968} connects the energetic and the stress intensity descriptions and remains the standard tool for reading the crack driving force from a computed solution. It is path independent for a homogeneous elastic body without body forces or thermal strain, and under a non-uniform temperature field it has to be completed by an area term, which Section~\ref{sec:crack_propagation} recalls. Within Linear Elastic Fracture Mechanics (LEFM), the direction of growth under mixed-mode loading is commonly taken from the maximum hoop stress criterion of Erdogan and Sih~\cite{erdogan1963} or from the minimum strain energy density criterion of Sih~\cite{sih1974}. Cohesive zone models in the spirit of Barenblatt~\cite{barenblatt1962} and Dugdale~\cite{dugdale1960} add a process zone governed by a traction-separation law, and they became the basis of the cohesive element methods for crack propagation~\cite{xu1994,camacho1996}.

Whatever theory is adopted, a sharp-crack computation has to represent a moving discontinuity within a fixed discretization. Early finite element treatments required the mesh to conform to the crack faces and to be rebuilt whenever the crack advanced~\cite{bittencourt1996}, which becomes cumbersome for curved paths that are not known in advance. Meshfree methods such as the Element-Free Galerkin method relaxed the meshing burden and were applied to crack growth early on~\cite{belytschko1994}. The Partition of Unity framework~\cite{melenk1996} then led to the Extended Finite Element Method (XFEM)~\cite{belytschko1999,moes1999}, which enriches the approximation with a Heaviside function across the crack and with the Williams functions around the tip, so that the displacement jump and the $\sqrt{r}$ singularity are both captured on a mesh that does not conform to the crack. Some variants track the crack geometry with level set functions~\cite{stolarska2001}. Fries and Belytschko review the method and its variants in~\cite{fries2010}.

Two extensions of the sharp-crack route to thermal loading are relevant here. Duflot~\cite{duflot2008} derived temperature enrichment functions for insulated and conducting cracks and extracted mixed-mode SIFs under thermal load with interaction integrals~\cite{yau1980}, and KC and Kim~\cite{kckim2008} gave the interaction integrals for thermal fracture of graded materials, whose auxiliary fields the graded case requires. Enrichment is not the only route to a sharp crack under thermal load. Ai and Augarde~\cite{ai2019cracking} model thermoelastic fracture with an adaptive cracking particle method that uses no enrichment functions and reaches the accuracy of the enriched methods with fewer degrees of freedom. All of these methods keep the crack sharp and connect directly to the classical propagation criteria, which is the property retained here. Their cost is the explicit tracking of the crack geometry, and topological events such as branching and merging need dedicated treatment.

The energy minimization formulation of Francfort and Marigo~\cite{francfort1998} avoids the tracking altogether. Its regularized implementation by Bourdin et al.~\cite{bourdin2000} replaces the sharp crack by a diffuse damage band whose width is set by a regularization length. Thermodynamically consistent phase-field formulations and operator split solvers followed from Miehe et al.~\cite{miehe2010,miehe2010cmame}, the related gradient damage models were analyzed by Pham et al.~\cite{pham2011}, energy decompositions that restrict damage to tensile states~\cite{amor2009} and Ginzburg-Landau type evolution equations~\cite{kuhn2010} widened the modeling options, and the framework was extended to dynamic fracture~\cite{borden2012}. The reviews of Ambati et al.~\cite{ambati2015} and Wu et al.~\cite{wu2020} give the whole picture.

For thermo-mechanical fracture, Miehe et al.~\cite{miehe2015} extended the phase-field model to thermoelastic solids, and Bourdin et al.~\cite{bourdin2014} and Sicsic et al.~\cite{sicsic2014} showed that such models reproduce the periodic and hierarchical crack patterns of quenched ceramic plates. More recent work brought monolithic solvers for thermoelastic fracture~\cite{mandal2021fracture}, hybrid formulations for thermoelastic crack propagation~\cite{tangella2022hybrid}, models with temperature-dependent fracture properties for hot cracking in additive manufacturing~\cite{ruan2023thermomechanical}, and applications to Functionally Graded Materials (FGMs)~\cite{hirshikesh2019}, whose composition is engineered for severe thermal environments and at the same time decides where and how a crack grows. In the phase-field description, initiation, curved growth, merging and branching all emerge from the energy minimization without any tracking of the crack, and two of these solutions serve as references below. This description has two costs. The damage band has to be resolved by elements much smaller than the regularization length, which makes large problems expensive, and fracture mechanics quantities such as the SIFs are recovered only indirectly from the diffuse fields. The method proposed in this paper keeps the sharp crack and the direct access to the SIFs, and shares with the phase-field description the principle of obtaining every field by minimizing an energy.

Over the same period, neural network solvers for partial differential equations have matured. Physics-Informed Neural Networks (PINNs)~\cite{raissi2019} represent the solution by a network trained to minimize the residual of the governing equations, with software frameworks~\cite{lu2021deepxde} and a growing body of applications reviewed by Karniadakis et al.~\cite{karniadakis2021}. In solid mechanics they have been used for forward and inverse problems~\cite{haghighat2021}, and residual formulations have been applied to heat conduction and conjugate heat transfer~\cite{cai2021}. For boundary value problems the energy itself can serve as the loss. The Deep Ritz Method~\cite{e2018deepritz} introduced this idea for variational problems in general, and the Deep Energy Method (DEM)~\cite{samaniego2020,nguyenthanh2020} applied it to computational mechanics. In both, the potential energy is minimized directly and the derivatives come from automatic differentiation. An energy loss involves lower derivatives than a strong-form residual and includes the natural boundary conditions of the functional, and the essential conditions can be imposed exactly through distance functions~\cite{sukumar2022}. Weak form residuals~\cite{kharazmi2021} and mixed formulations with the stresses as additional outputs~\cite{fuhg2022} improve the accuracy near stress concentrations. The energy still has to be integrated numerically over the domain, so these solvers dispense with a mesh in the sense of connectivity but not with a set of integration points.

The DEM has been combined with phase-field modeling of brittle fracture by Goswami et al., with transfer learning across the load increments~\cite{goswami2020tl} and with adaptive schemes for fourth-order models~\cite{goswami2020fourth}, and studied systematically by Manav et al.~\cite{manav2024}. Operator learning variants predict crack paths across loading scenarios with DeepONets~\cite{goswami2022deeponet}, Physics-Informed Neural Operators (PINOs)~\cite{li2024pino} combine operator learning with physics constraints, and the energy minimization has since been extended to higher-order anisotropic crack density functionals~\cite{plunge2026} and to a second physical field~\cite{dean2026multiphysics}. These works show that a network can reproduce phase-field solutions of brittle fracture by energy minimization, and they also show what the combination costs. A network that takes the coordinates as its input is globally smooth and learns the smooth, slowly varying part of a field much faster than its steep, localized part~\cite{rahaman2019,wang2022ntk,wang2021gradient}, whereas the phase-field solution is concentrated in a narrow band with steep gradients, so accurate training needs dense sampling in the band and often a specialized architecture, such as Fourier features~\cite{tancik2020}, Sinusoidal Representation Networks (SIRENs)~\cite{sitzmann2020} or Kolmogorov-Arnold Networks (KANs)~\cite{liu2024kan}. A continuous network of the coordinates alone cannot represent the displacement jump across a sharp crack at all, whatever its size. For that class of approximation the diffuse description is therefore a necessity, and the network spends most of its capacity on the localized field it approximates worst.

Several recent works return to a discrete crack inside the neural approximation. One route decomposes the domain and assigns separate networks to the two sides of the discontinuity, following the Extended PINN (XPINN) formulation~\cite{jagtap2020xpinn}, and enriches them with the XFEM crack-tip functions, as Lotfalian et al.~\cite{lotfalian2025xpinn} do for several cracks, but the decomposition becomes cumbersome once the crack is curved and growing. The other route keeps a single network and encodes the crack in its input. Discontinuity-Embedded Neural Networks (DENNs)~\cite{zhao2025denns} add to the input a scalar embedding function that is discontinuous across the crack and smooth elsewhere, so that points on opposite faces map to well separated inputs and a displacement jump can be represented across the crack, its magnitude being left to the training. The Discontinuity-Embedded Deep Energy Method~\cite{zhao2024dedem} places this crack in an energy formulation, and the Extended Deep Energy Method of Wang et al.~\cite{wang2025xdem} adds the Williams near-tip functions with trainable amplitudes, in analogy with XFEM. These formulations recover a sharp crack, need no regularization length and give direct access to the crack-tip quantities. They have been applied to isothermal elastic fracture, and in the present work they are extended to thermal loading.

Thermo-mechanical fracture presents several challenges. A transient heat conduction problem has to be solved on the cracked domain, the crack faces are adiabatic so the temperature jumps across them, the near-tip fields under thermal load have their own structure~\cite{duflot2008}, and the fracture resistance may depend on the temperature at the tip~\cite{ruan2023thermomechanical}. So far the two ingredients, a temperature field solved by a network and an embedded sharp crack propagated from the trained fields, have been treated in separate studies. Coupled thermo-mechanics of a graded material has been solved with PINNs without a crack~\cite{raj2021pinnfgm}, and the embedded crack formulations have a crack without a temperature field. Where cracked bodies under thermal load have been treated with networks, the crack has been held fixed, the temperature prescribed rather than solved, and the analysis has stopped at the SIFs, in the energy-informed peridynamic network of Yu and Zhou~\cite{yuzhou2023nonlocal} and in the enriched graded plates of Yadav et al.~\cite{yadav2026mpinn}. A network solver that solves the temperature on the cracked domain and grows the crack under a fracture criterion evaluated on the trained fields has not been reported.

In this paper we develop an extended deep energy method for thermo-mechanical crack propagation, in which the crack is a sharp polyline that both fields receive through one embedding function built from the current crack geometry, as in the discontinuity-embedded networks~\cite{zhao2025denns,zhao2024dedem}. Two networks share this input, one for the temperature and one for the displacement. The transient heat conduction equation is advanced in time by backward Euler, recast as an incremental functional and minimized, so the temperature follows from the same kind of minimization that delivers the displacement, and the two minimizations are staggered at every load step in the manner of the operator splits of phase-field computations~\cite{miehe2010cmame}. The embedding lets the temperature take different values on the two crack faces, and the zero flux across the faces is the natural condition of the thermal functional, supplemented by a weak consistency term. The displacement includes the Williams expansion~\cite{williams1957} near the tip with trainable amplitudes, as in XFEM and, for the isothermal case, in~\cite{wang2025xdem}, while the temperature receives no tip enrichment, for the reasons given in Section~\ref{sec:neural_energy_formulation}. The energies are estimated on integration points that are redrawn during training and densified near the tip.

Crack growth follows LEFM. The SIFs are extracted by the interaction integral~\cite{yau1980,shih1986}, which under thermal load includes the area term of Wilson and Yu~\cite{wilson1979} in addition to the contour term. The area term does not appear in the isothermal form of the integral and is left out, for example, by Ghaffari et al.~\cite{ghaffari2016xfem}. Without it the extracted SIFs stay finite but drift with the contour radius~\cite{wilson1979}. Since the complete integral is path independent, a drift with the radius marks an unreliable reading, and in the present work every extraction is therefore read over a sweep of contour radii and refused when it drifts beyond a set tolerance. The direction of growth follows the maximum hoop stress criterion~\cite{erdogan1963}, the criterion is evaluated on the energy released by the kink about to be taken~\cite{cotterell1980,nuismer1975}, which keeps it sensitive to the sign of the thermal load, and the resistance is a temperature-dependent critical energy release rate at the crack-tip temperature.

The embedded crack and the near-tip enrichment are inherited from the works cited above, and the growth criteria are the established relations named. What is new is their combination with a temperature field solved on the cracked domain, the propagation of the crack from the trained fields, and the numerical procedures that make the extraction reliable: the evaluation of the area term on the trained fields, the sweep of the contour radius, and the resampling of the integration points.

The method is assessed on four benchmarks. A stationary thermal edge crack with a published SIF~\cite{wang2023cmt} verifies the extraction. A single-edge notched tension test under three thermal loads~\cite{tangella2022hybrid} and its functionally graded shear counterpart are compared with a phase-field reference and, for the shear test, with an independent sharp-crack finite element solution. A notched cruciform specimen~\cite{mandal2021fracture} is compared with the published crack paths of six independent solutions under either load alone and of two under the combined load. Across these cases the method reproduces the crack paths and the influence of the thermal load on initiation and peak load, and the crack remains a sharp geometric entity from which the SIFs and the crack-tip temperature are read directly.

The paper is organized as follows. Section~\ref{sec:formulation} states the governing equations and the energy functionals. Section~\ref{sec:crack_representation} describes the crack geometry and the embedding function. Section~\ref{sec:neural_energy_formulation} gives the two networks, their losses and the staggered solution. Section~\ref{sec:crack_propagation} sets out the extraction of the SIFs, its admissibility checks and the propagation criterion. Section~\ref{section:numerical_examples} presents the four benchmarks, and Section~\ref{sec:conclusions} concludes. The numerical parameters, the cost and the reference solutions are given in the appendices.

\section{Governing equations and energy functional}
\label{sec:formulation}

Consider a cracked thermoelastic body subjected to transient thermal loading. The temperature field evolves in time through heat conduction, while the mechanical response is evaluated as a quasi-static equilibrium problem at each loading or time step. For a fixed crack geometry we treat the thermo-mechanical coupling in a one-way sense, in which the temperature field induces thermal expansion and therefore modifies the mechanical stress state, whereas heat generation due to deformation and the feedback of mechanical fields into the thermal balance are neglected. The mechanical solution nevertheless decides whether the crack advances, and the advanced crack changes the domain on which the next temperature field is solved, so the two fields are coupled through the crack geometry. This treatment is appropriate for quasi-static brittle fracture, where deformations remain small and slow, so the temperature changes produced by straining are negligible compared with those imposed by the thermal loading~\cite{duflot2008}.

\begin{figure}[htbp]
  \centering
  \includegraphics[width=0.6376\textwidth]{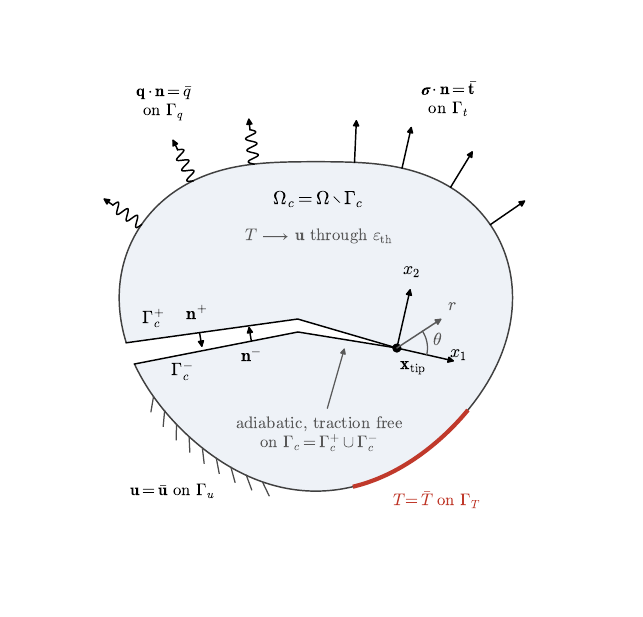}
  \caption{The thermo-mechanical problem on a cracked body. The crack $\Gamma_c=\Gamma_c^{+}\cup\Gamma_c^{-}$ has adiabatic, traction-free faces, whose outward normals $\mathbf{n}^{\pm}$ point into the opening, and the active tip $\mathbf{x}_{\mathrm{tip}}$. Temperature and displacement are prescribed on $\Gamma_T$ and $\Gamma_u$, heat flux and traction on $\Gamma_q$ and $\Gamma_t$. For a fixed crack the temperature enters the mechanical problem through the thermal strain $\boldsymbol{\varepsilon}_{\mathrm{th}}$, and the mechanical field acts on the heat conduction problem only through the crack it advances.}
  \label{fig:problem_setting}
\end{figure}

The crack is an internal geometrical discontinuity of the body. The thermal and mechanical fields are solved on the cracked domain, and the crack faces carry their own boundary conditions, as sketched in Fig.~\ref{fig:problem_setting}. It is held fixed while those fields are solved, and advanced only by the growth rule of Section~\ref{sec:crack_propagation}. Table~\ref{tab:main_symbols} collects the main symbols.

\begin{table}[htbp]
  \centering
  \caption{Main symbols. The load or time step index $n$ appears as a superscript. Descriptive subscripts are upright ($\boldsymbol{\varepsilon}_{\mathrm{th}}$, $T_{\mathrm{tip}}$). The mass density is written $\varrho$ so that $\rho$ can denote the crack embedding function, and $\|\cdot\|$ is the Euclidean norm.}
  \label{tab:main_symbols}
  \small
  \begin{tabular}{@{}l>{\raggedright\arraybackslash}p{5.55cm}@{\quad}l>{\raggedright\arraybackslash}p{5.75cm}@{}}
    \toprule
    Symbol & Meaning & Symbol & Meaning \\
    \midrule
    $\Omega$, $\Omega_c$ & body, cracked domain & $\rho(\mathbf{x})$ & crack embedding function \\
    $\Gamma_c^{\pm}$ & crack faces & $\tilde{\mathbf{x}}$ & crack-aware network input \\
    $\mathbf{u}$, $T$ & displacement, temperature & $\boldsymbol{\theta}_u$, $\boldsymbol{\theta}_T$ & network parameters \\
    $\boldsymbol{\varepsilon}$, $\boldsymbol{\varepsilon}_{\mathrm{e}}$, $\boldsymbol{\varepsilon}_{\mathrm{th}}$ & total, elastic, thermal strain & $a_I$, $a_{II}$ & trainable enrichment amplitudes \\
    $\boldsymbol{\sigma}$, $\psi_{\mathrm{e}}$ & stress, elastic energy density & $(r,\theta)$, $\beta$ & tip polar frame, tangent angle \\
    $\varrho$, $C_p$, $k$ & density, specific heat, conductivity & $K_I$, $K_{II}$, $J$ & extracted SIFs, $J$-integral \\
    $\alpha$, $T_0$ & expansion coefficient, reference temperature & $\Gamma_r$, $M^{\mathrm{th}}$ & extraction contour, thermal area term \\
    $E$, $\nu$, $E'$ & elastic constants, effective modulus & $p$ & path-independence exponent \\
    $\Pi_T^n$, $\Pi_u^n$ & incremental energy functionals & $\theta_c$ & kink angle \\
    $G_c(T)$ & critical energy release rate & $k_I$, $k_{II}$ & SIFs at the kinked tip \\
    $T_{\mathrm{tip}}^{n}$ & crack-tip temperature & $G(\theta_c)$ & energy release rate of the kink \\
    $\Delta a$, $\delta$ & crack increment, separation margin & $\chi$, $\chi_c$ & criterion ratio, threshold \\
    $\mathbf{q}$, $\bar{q}$, $Q$ & heat flux, prescribed normal flux, heat source & $\bar{\mathbf{t}}$, $\mathbf{b}$ & prescribed traction, body force \\
    $\mathbf{n}$, $\mathbf{n}^{\pm}$ & outward normal, on the crack faces & $\mathbf{p}_j$, $\mathbf{t}_j$, $\mathbf{n}_j$ & crack points, segment tangent and normal \\
    $\mathbf{D}_u$, $D_{\mathrm{tip}}$ & liftings of the correction and the enrichment & $\alpha_\rho$, $\ell_\rho$, $\delta_c$ & embedding decay, endpoint overlap \\
    \bottomrule
  \end{tabular}
\end{table}

\subsection{Cracked domain and primary fields}
\label{subsec:geometry}

Let the body occupy an open bounded domain $\Omega \subset \mathbb{R}^{2}$ containing an explicitly represented crack. The crack has two faces, $\Gamma_c^{+}$ and $\Gamma_c^{-}$, which coincide with the crack curve as point sets and are distinguished by the side from which they are approached, and $\Gamma_c=\Gamma_c^{+}\cup\Gamma_c^{-}$ denotes the crack as a whole. The cracked domain is
\begin{equation}
  \Omega_c = \Omega \setminus \Gamma_c .
\end{equation}
In two dimensions, the crack is an internal curve of zero measure. Hence, the energy integrals may be evaluated over the background domain $\Omega$, provided that the admissible fields are allowed to have distinct traces on the two crack faces.

The primary unknowns are the displacement field $\mathbf{u}$ and the temperature field $T$,
\begin{equation}
  \mathbf{u}(\mathbf{x},t)
  =
  \{u(\mathbf{x},t),v(\mathbf{x},t)\}^{\mathrm{T}},
  \qquad
  T = T(\mathbf{x},t),
  \qquad
  \mathbf{x} \in \Omega_c ,
  \quad
  t \in [0,t_f] ,
\end{equation}
where $t_f$ denotes the final time of the loading program. Inertial effects are not included in the mechanical formulation.

\subsection{Transient heat conduction}
\label{subsec:thermal}

The temperature field is governed by transient heat conduction. For an isotropic material, the heat flux $\mathbf{q}$ is described by Fourier's law,
\begin{equation}
  \mathbf{q}
  =
  -k\nabla T ,
  \label{eq:fourier_law}
\end{equation}
where $k$ denotes the thermal conductivity. Combining Fourier's law with the local energy balance gives
\begin{equation}
  \varrho C_p
  \frac{\partial T}{\partial t}
  -
  \nabla\cdot
  \left(
    k\nabla T
  \right)
  =
  Q
  \qquad
  \text{in } \Omega_c \times (0,t_f],
  \label{eq:transient_heat_equation}
\end{equation}
where $\varrho$ is the mass density, $C_p$ is the specific heat capacity, and $Q$ is a volumetric heat source. No internal heat is generated in the numerical examples, that is, $Q=0$.

The external boundary is partitioned for the thermal problem as
\begin{equation}
  \partial\Omega
  =
  \Gamma_T \cup \Gamma_q,
  \qquad
  \Gamma_T \cap \Gamma_q = \varnothing ,
  \label{eq:thermal_boundary_partition}
\end{equation}
where the temperature is prescribed on $\Gamma_T$ and the normal heat flux $\mathbf{q}\cdot\mathbf{n}=-k\,\partial T/\partial n$ is prescribed on $\Gamma_q$, with $\mathbf{n}$ the outward unit normal. Along the crack, both faces are treated as adiabatic boundaries. The thermal boundary and initial conditions are then
\begin{equation}
  T
  =
  \bar{T}
  \quad
  \text{on } \Gamma_T,
  \qquad
  \mathbf{q}\cdot\mathbf{n}
  =
  \bar{q}
  \quad
  \text{on } \Gamma_q,
  \qquad
  \mathbf{q}\cdot\mathbf{n}^{\pm}
  =
  0
  \quad
  \text{on } \Gamma_c^{\pm},
  \label{eq:thermal_boundary_conditions}
\end{equation}
where $\mathbf{n}^{\pm}$ denotes the outward unit normal of the body on the crack face $\Gamma_c^{\pm}$, which points into the opening and is given in terms of the crack geometry in Section~\ref{subsec:explicit_crack_geometry}, together with the initial condition
\begin{equation}
  T(\mathbf{x},0)
  =
  T_{\mathrm{ini}}(\mathbf{x})
  \qquad
  \text{in } \Omega_c .
  \label{eq:thermal_initial_condition}
\end{equation}

The heat equation is advanced in time with the implicit backward Euler scheme, which is unconditionally stable for linear heat conduction~\cite{hughes2000fem}. The time step can therefore follow the load steps of the incremental analysis, and each step reduces to the minimization of a single functional, as shown below. Let $T^n$ and $T^{n-1}$ denote the temperatures at two consecutive time steps, and let $\Delta t=t_n-t_{n-1}$. The time discrete heat equation is then
\begin{equation}
  \varrho C_p
  \frac{T^n-T^{n-1}}{\Delta t}
  -
  \nabla\cdot
  \left(
    k\nabla T^n
  \right)
  =
  Q^n
  \qquad
  \text{in } \Omega_c .
  \label{eq:heat_backward_euler}
\end{equation}

Equation~\eqref{eq:heat_backward_euler} can be obtained as the Euler-Lagrange equation of the incremental thermal functional
\begin{equation}
  \Pi_T^n[T^n]
  =
  \int_{\Omega_c}
  \left[
    \frac{\varrho C_p}{2\Delta t}
    \left( T^n - T^{n-1} \right)^2
    +
    \frac{k}{2}
    \left\| \nabla T^n \right\|^2
    -
    Q^n T^n
  \right]
  d\Omega
  +
  \int_{\Gamma_q}
    \bar{q}^{\,n} T^n
  d\Gamma ,
  \label{eq:thermal_incremental_functional}
\end{equation}
in the sense that the stationarity condition of \eqref{eq:thermal_incremental_functional} with respect to $T^n$ recovers \eqref{eq:heat_backward_euler} together with the natural thermal boundary conditions.

\subsection{Thermoelastic constitutive response}
\label{subsec:constitutive}

At each loading or time step, the current temperature field produces thermal expansion and consequently contributes to the stress state. Under the small strain assumption, the total strain is defined as
\begin{equation}
  \boldsymbol{\varepsilon}
  =
  \frac{1}{2}
  \left(
    \nabla\mathbf{u}
    +
    \nabla\mathbf{u}^{\mathrm{T}}
  \right),
  \label{eq:small_strain}
\end{equation}
and the thermal strain, measured relative to the stress-free reference temperature $T_0$, is
\begin{equation}
  \boldsymbol{\varepsilon}_{\mathrm{th}}
  =
  \alpha
  \left(
    T-T_0
  \right)
  \mathbf{I},
  \label{eq:thermal_strain}
\end{equation}
where $\alpha$ is the coefficient of thermal expansion and $\mathbf{I}$ is the second-order identity tensor. The elastic strain is obtained by subtracting the thermal strain from the total strain,
\begin{equation}
  \boldsymbol{\varepsilon}_{\mathrm{e}}
  =
  \boldsymbol{\varepsilon}
  -
  \boldsymbol{\varepsilon}_{\mathrm{th}} .
  \label{eq:elastic_strain}
\end{equation}

The elastic strain energy density is defined as
\begin{equation}
  \psi_{\mathrm{e}}
  =
  \frac{1}{2}
  \boldsymbol{\varepsilon}_{\mathrm{e}}
  :
  \mathbb{C}
  :
  \boldsymbol{\varepsilon}_{\mathrm{e}},
  \label{eq:elastic_energy_density}
\end{equation}
where $\mathbb{C}$ is the fourth-order elastic stiffness tensor. The corresponding Cauchy stress follows from the Duhamel-Neumann thermoelastic law,
\begin{equation}
  \boldsymbol{\sigma}
  =
  \mathbb{C}
  :
  \boldsymbol{\varepsilon}_{\mathrm{e}} .
  \label{eq:thermoelastic_stress}
\end{equation}
For an isotropic solid, this relation reduces to
\begin{equation}
  \boldsymbol{\sigma}
  =
  \lambda
  \mathrm{tr}
  \left(
    \boldsymbol{\varepsilon}_{\mathrm{e}}
  \right)
  \mathbf{I}
  +
  2\mu
  \boldsymbol{\varepsilon}_{\mathrm{e}},
  \label{eq:isotropic_thermoelastic_stress}
\end{equation}
where $\lambda$ and $\mu$ are the Lam\'e constants, related to the Young's modulus $E$ and the Poisson's ratio $\nu$ by $\lambda = E\nu/[(1+\nu)(1-2\nu)]$ and $\mu = E/[2(1+\nu)]$. Plane strain is assumed in the numerical examples unless stated otherwise. Under plane strain the out-of-plane thermal expansion is suppressed by the constraint $\varepsilon_{zz}=0$ and is carried by $\sigma_{zz}$ instead. Eliminating $\sigma_{zz}$ from the in-plane relations shows that the in-plane problem is that of \eqref{eq:thermal_strain} with the coefficient of thermal expansion replaced by the effective value $(1+\nu)\alpha$. In what follows $\alpha$ is understood as this effective value whenever plane strain is assumed, whereas the material tables list $\alpha$ itself. The energy then differs from the three-dimensional one only by a term that depends on the temperature alone, which affects neither the displacement nor the stresses. The material parameters ($E$, $\nu$, $\alpha$, and the thermal properties) may also vary with position, as in Functionally Graded Materials (FGMs). The formulation carries over unchanged, with all parameters evaluated pointwise, and this case is exercised in the graded shear test of Section~\ref{subsec:sens_fgm}.

In component form, the elastic strain energy density can be written as
\begin{equation}
  \psi_{\mathrm{e}}
  =
  \frac{1}{2}
  \left(
    \varepsilon_{\mathrm{e},xx}\sigma_{xx}
    +
    \varepsilon_{\mathrm{e},yy}\sigma_{yy}
    +
    2\varepsilon_{\mathrm{e},xy}\sigma_{xy}
  \right),
  \label{eq:elastic_energy_component}
\end{equation}
which is equivalent to \eqref{eq:elastic_energy_density} under the two-dimensional tensorial strain convention.

\subsection{Quasi-static mechanical equilibrium}
\label{subsec:mechanical_equilibrium}

Once the temperature field $T^n$ at time $t_n$ is known, the mechanical response is obtained from quasi-static equilibrium,
\begin{equation}
  \nabla\cdot\boldsymbol{\sigma}
  +
  \mathbf{b}
  =
  \mathbf{0}
  \qquad
  \text{in } \Omega_c ,
  \label{eq:mechanical_equilibrium}
\end{equation}
where $\mathbf{b}$ denotes the body force. The external boundary is partitioned for the mechanical problem as
\begin{equation}
  \partial\Omega
  =
  \Gamma_u \cup \Gamma_t,
  \qquad
  \Gamma_u \cap \Gamma_t = \varnothing ,
  \label{eq:mechanical_boundary_partition}
\end{equation}
where displacement is prescribed on $\Gamma_u$ and traction is prescribed on $\Gamma_t$. The boundary and crack-face conditions are
\begin{equation}
  \mathbf{u}
  =
  \bar{\mathbf{u}}
  \quad
  \text{on } \Gamma_u,
  \qquad
  \boldsymbol{\sigma}\cdot\mathbf{n}
  =
  \bar{\mathbf{t}}
  \quad
  \text{on } \Gamma_t,
  \qquad
  \boldsymbol{\sigma}\cdot\mathbf{n}^{\pm}
  =
  \mathbf{0}
  \quad
  \text{on } \Gamma_c^{\pm} .
  \label{eq:mechanical_boundary_conditions}
\end{equation}
The traction-free condition on the crack faces describes an open crack without contact. We retain this open crack assumption throughout. Loading states that close the crack locally would require a unilateral contact condition.

For a fixed crack and the temperature field $T^n$, the displacement field $\mathbf{u}^n$ minimizes the mechanical potential energy
\begin{equation}
  \Pi_u^n[\mathbf{u}^n;T^n]
  =
  \int_{\Omega_c}
    \psi_{\mathrm{e}}
  d\Omega
  -
  W_{\mathrm{ext}}^n,
  \label{eq:mechanical_potential}
\end{equation}
where $\psi_{\mathrm{e}}$ is evaluated with the elastic strain formed from $\mathbf{u}^n$ and $T^n$, and the external work is
\begin{equation}
  W_{\mathrm{ext}}^n
  =
  \int_{\Omega_c}
    \mathbf{b}\cdot\mathbf{u}^n
  d\Omega
  +
  \int_{\Gamma_t}
    \bar{\mathbf{t}}^{\,n}\cdot\mathbf{u}^n
  d\Gamma .
  \label{eq:external_work}
\end{equation}
The examples of Section~\ref{section:numerical_examples} introduce their loading through essential boundary conditions, on the displacement, the temperature, or both. If neither body forces nor surface tractions are prescribed, then $W_{\mathrm{ext}}^n=0$, and the mechanical problem reduces to the minimization of the stored elastic energy over the admissible displacement fields. When the essential conditions leave a rigid body mode free, the minimizer is defined only up to that motion, and the free mode is removed by an additional point constraint.

\section{Explicit crack representation}
\label{sec:crack_representation}

The crack is represented explicitly as a sharp geometrical entity. Rather than splitting the physical domain into separate subdomains, the crack is encoded through an additional scalar input to the neural approximation. This allows a single global network to distinguish points located on opposite sides of the crack, even when their Euclidean coordinates nearly coincide. This construction follows the embedding function idea introduced for Discontinuity-Embedded Neural Networks (DENNs) by Zhao and Shao~\cite{zhao2025denns} and adopted in Deep Energy Method formulations~\cite{zhao2024dedem,wang2025xdem}. As a result, the approximation can represent a displacement jump across the crack while retaining a single global neural representation over the entire domain.

\subsection{Crack geometry}
\label{subsec:explicit_crack_geometry}

The crack is described by an ordered set of control points
\begin{equation}
  \mathcal{P}
  =
  \left\{
  \mathbf{p}_{0},
  \mathbf{p}_{1},
  \ldots,
  \mathbf{p}_{N_c}
  \right\},
  \qquad
  \mathbf{p}_{j}\in\mathbb{R}^{2},
\end{equation}
and is represented as the polyline passing through these points,
\begin{equation}
  \Gamma_c
  =
  \bigcup_{j=0}^{N_c-1}
  \Gamma_c^j ,
\end{equation}
where the $j$th segment is defined by
\begin{equation}
  \Gamma_c^j
  =
  \left\{
  \mathbf{x}
  =
  \mathbf{p}_{j}
  +
  \xi
  \left(
    \mathbf{p}_{j+1}
    -
    \mathbf{p}_{j}
  \right),
  \quad
  0\leq \xi \leq 1
  \right\}.
\end{equation}

Each segment is associated with a length and a unit direction vector,
\begin{equation}
  l_j
  =
  \left\|
    \mathbf{p}_{j+1}
    -
    \mathbf{p}_{j}
  \right\|,
  \qquad
  \mathbf{t}_j
  =
  \frac{
    \mathbf{p}_{j+1}
    -
    \mathbf{p}_{j}
  }{
    l_j
  },
\end{equation}
and the corresponding unit normal is chosen as
\begin{equation}
  \mathbf{n}_j
  =
  \left\{
    -t_{j,2},
    t_{j,1}
  \right\}^{\mathrm{T}} .
  \label{eq:segment_normal}
\end{equation}
The face $\Gamma_c^{+}$ is the one toward which $\mathbf{n}_j$ points and $\Gamma_c^{-}$ the other. The outward normals of the body on the two faces are therefore opposite, $\mathbf{n}^{+}=-\mathbf{n}_j$ and $\mathbf{n}^{-}=\mathbf{n}_j$ along segment $j$, and both point into the opening, as drawn in Fig.~\ref{fig:problem_setting}. The arc length along the crack is given by
\begin{equation}
  s_0=0,
  \qquad
  s_j
  =
  \sum_{m=0}^{j-1} l_m,
  \qquad
  L_c
  =
  \sum_{m=0}^{N_c-1} l_m,
\end{equation}
where $L_c$ denotes the total crack length.

The active crack tip is taken as the last point of the current crack path,
\begin{equation}
  \mathbf{x}_{\mathrm{tip}}
  =
  \mathbf{p}_{N_c},
\end{equation}
and its tangent and normal are inherited from the final segment,
\begin{equation}
  \mathbf{t}_{\mathrm{tip}}
  =
  \mathbf{t}_{N_c-1},
  \qquad
  \mathbf{n}_{\mathrm{tip}}
  =
  \left\{
    -t_{\mathrm{tip},2},
    t_{\mathrm{tip},1}
  \right\}^{\mathrm{T}} .
\end{equation}
These two vectors define the local crack-tip frame used for both the near-tip enrichment of Section~\ref{sec:neural_energy_formulation} and the crack-tip post-processing of Section~\ref{sec:crack_propagation}.

When crack-face quantities are required, the crack is sampled along its arc length and the sampling points are shifted slightly along the local normal. Let $\mathbf{x}_{c}(s)$ be the point of the polyline at arc length $s\in[0,L_c]$ and $\mathbf{n}(s)$ the normal $\mathbf{n}_j$ of the segment that contains it. The two face samples, one on the side of each face, are
\begin{equation}
  \mathbf{x}^{+}(s)
  =
  \mathbf{x}_{c}(s)
  +
  \epsilon_c \mathbf{n}(s),
  \qquad
  \mathbf{x}^{-}(s)
  =
  \mathbf{x}_{c}(s)
  -
  \epsilon_c \mathbf{n}(s),
\end{equation}
where $\epsilon_c$ is a small numerical offset. The shifted points serve only to evaluate crack-face quantities, namely the flux consistency term of Section~\ref{subsec:transient_thermal_energy_loss} and the displacement-jump diagnostic of Section~\ref{subsec:crack_tip_quantities}, and the crack geometry remains the polyline $\Gamma_c$. The offset is chosen between two limits. It should be larger than the single-precision noise floor of the trained fields, and small enough that the jump sampled across the crack is not attenuated by the normal decay of the embedding of Section~\ref{subsec:crack_embedding}. The two samples should also lie on opposite sides of the embedded discontinuity, which holds along the straight stretches of the polyline. Close to a corner the blending of Section~\ref{subsec:crack_embedding} displaces the discontinuity slightly from the polyline, and a pair of samples there can fall on one side and read no jump. The values used in each example are listed in Table~\ref{tab:discretization}, and Appendix~\ref{app:discretization} reports the attenuation they incur and the extent and the consequences of the corner effect.

\subsection{Crack embedding function}
\label{subsec:crack_embedding}

The crack geometry enters the neural approximation through a scalar crack embedding function $\rho(\mathbf{x})$, which is constructed to be discontinuous across the crack and smooth in the intact region. Ideally it satisfies~\cite{zhao2025denns}
\begin{equation}
  \lim_{\mathbf{x}\rightarrow\mathbf{x}_{c}^{-}}
  \rho(\mathbf{x})
  \neq
  \lim_{\mathbf{x}\rightarrow\mathbf{x}_{c}^{+}}
  \rho(\mathbf{x}),
  \label{eq:crack_function_condition_jump}
\end{equation}
at every point $\mathbf{x}_{c}$ of the crack, where $\mathbf{x}\rightarrow\mathbf{x}_{c}^{\pm}$ denotes the one-sided limit taken from the side of the crack face $\Gamma_c^{\pm}$, together with
\begin{equation}
  \lim_{\mathbf{x}\rightarrow\mathbf{x}_{0}}
  \rho(\mathbf{x})
  =
  \rho(\mathbf{x}_{0}),
  \qquad
  \lim_{\mathbf{x}\rightarrow\mathbf{x}_{0}}
  \nabla\rho(\mathbf{x})
  =
  \nabla\rho(\mathbf{x}_{0}),
  \qquad
  \mathbf{x}_{0}\in\Omega\setminus\Gamma_c .
  \label{eq:crack_function_condition_smooth}
\end{equation}
The first condition provides the separation required to distinguish the two crack faces, while the second condition prevents the introduction of artificial discontinuities away from the crack. The implemented function given below meets these conditions up to three local departures, in a strip of width $\delta_c$ ahead of the tip, in a neighborhood of each corner of the polyline, and on the lines normal to a segment through its endpoints, which are described in Appendix~\ref{app:discretization}.

For a single crack segment, the crack embedding function can be factorized~\cite{zhao2025denns,wang2025xdem} as
\begin{equation}
  \rho(\mathbf{x})
  =
  f_1(\mathbf{x}) f_2(\mathbf{x}) ,
  \label{eq:rho_xdem_form}
\end{equation}
where $f_1$ is a sign component and $f_2$ is a localization component. The sign component is defined as
\begin{equation}
  f_1(\mathbf{x})
  =
  \operatorname{sgn}
  \left(
    D_s(\mathbf{x};\Gamma_{cl})
  \right),
  \label{eq:f1_sign}
\end{equation}
where $D_s(\mathbf{x};\Gamma_{cl})$ is the signed distance to the extended crack line $\Gamma_{cl}$, the infinite straight line obtained by extending the crack segment beyond its two endpoints, and
\begin{equation}
  \operatorname{sgn}(a)
  =
  \begin{cases}
    -1, & a\leq 0,\\
    1, & a>0 .
  \end{cases}
  \label{eq:sign_function}
\end{equation}
Let $\hat{\mathbf{x}}$ be the closest point to $\mathbf{x}$ on the extended crack line and let $\mathbf{n}$ be the segment normal $\mathbf{n}_j$ of \eqref{eq:segment_normal}, so that the signed distance is positive on the side of $\Gamma_c^{+}$ and negative on the side of $\Gamma_c^{-}$. The signed and unsigned distances are then written in terms of the Euclidean distance as
\begin{equation}
  D_s(\mathbf{x};\Gamma_{cl})
  =
  \operatorname{sgn}
  \left[
    \mathbf{n}\cdot
    \left(
      \mathbf{x}-\hat{\mathbf{x}}
    \right)
  \right]
  \left\|
    \mathbf{x}-\hat{\mathbf{x}}
  \right\|,
  \label{eq:signed_distance}
\end{equation}
\begin{equation}
  D(\mathbf{x};\Gamma_{cl})
  =
  \left\|
    \mathbf{x}-\hat{\mathbf{x}}
  \right\| .
  \label{eq:unsigned_distance}
\end{equation}

The localization component is constructed from the signed distance functions associated with the two endpoint extension lines,
\begin{equation}
  f_2(\mathbf{x})
  =
  \operatorname{ReLU}^{2}
  \left(
    D_s(\mathbf{x};\Gamma_{c1})
    D_s(\mathbf{x};\Gamma_{c2})
  \right)
  \exp
  \left(
    D(\mathbf{x};\Gamma_{cl})
  \right),
  \label{eq:f2_localisation}
\end{equation}
where $\Gamma_{c1}$ and $\Gamma_{c2}$ are auxiliary extension lines at the two crack endpoints. The squared ReLU factor confines the embedding to the crack segment, while the distance-dependent factor controls its spatial variation.

Equation~\eqref{eq:rho_xdem_form} may also be interpreted as a weighted sign function~\cite{zhao2025denns},
\begin{equation}
  \rho(\mathbf{x})
  =
  \mathcal{W}(\psi(\mathbf{x}))
  \operatorname{sgn}
  \left(
    \phi(\mathbf{x})
  \right),
  \label{eq:weighted_sign_form}
\end{equation}
where $\phi$ is the signed distance to the extended crack surface, $\psi$ is a signed distance measure associated with the crack tips or endpoint extensions, and $\mathcal{W}$ is a smooth weight. A convenient choice is
\begin{equation}
  \mathcal{W}(\psi)
  =
  \operatorname{ReLU}^{2}(\psi)
  =
  \begin{cases}
    \psi^2, & \psi>0,\\
    0, & \psi\leq 0 .
  \end{cases}
  \label{eq:relu_weight}
\end{equation}
This form separates the two roles of the embedding. The sign term distinguishes the two faces of the crack, while the weight localizes the discontinuity to the crack and leaves the function elsewhere as regular as the weight itself. The implementation given next follows this form.

For the propagating crack considered here, the same weighted sign structure is evaluated on the current polyline $\mathbf{p}_0,\dots,\mathbf{p}_{N_c}$ of Section~\ref{subsec:explicit_crack_geometry}, with segment tangents $\mathbf{t}_j$, normals $\mathbf{n}_j$, lengths $l_j$ and arc lengths $s_j$ at the segment starts, $j=0,\dots,N_c-1$. For a point $\mathbf{x}$, each segment $j$ supplies a signed distance to its line, the position $t_j$ of a foot point limited to the segment, and the squared distance $r_j^2$ to that foot point,
\begin{equation}
  d_j(\mathbf{x}) = \mathbf{n}_j\cdot(\mathbf{x}-\mathbf{p}_j),
  \qquad
  t_j(\mathbf{x}) = \min\!\left(l_j,\max\!\left(0,\mathbf{t}_j\cdot(\mathbf{x}-\mathbf{p}_j)\right)\right),
  \qquad
  r_j^2(\mathbf{x}) = \left\|\mathbf{x}-\mathbf{p}_j-t_j\mathbf{t}_j\right\|^2,
  \label{eq:segment_projection}
\end{equation}
and the segment values are blended with Gaussian weights on the squared distance to the segment,
\begin{equation}
  w_j(\mathbf{x}) = \frac{\exp\!\left(-r_j^2/\sigma^2\right)}{\sum_k \exp\!\left(-r_k^2/\sigma^2\right)},
  \qquad
  \tilde{d}(\mathbf{x}) = \sum_j w_j d_j,
  \label{eq:segment_blending}
\end{equation}
\begin{equation}
  \tilde{s}(\mathbf{x}) = \sum_j w_j\left(s_j+t_j\right) + \max\!\left(0,\mathbf{t}_{N_c-1}\cdot(\mathbf{x}-\mathbf{p}_{N_c})\right) - \max\!\left(0,-\mathbf{t}_0\cdot(\mathbf{x}-\mathbf{p}_0)\right),
  \label{eq:arc_length_extension}
\end{equation}
with $\sigma$ half the shortest segment length. The blending gives one representative signed distance $\tilde{d}$ and one projected arc length $\tilde{s}$ that are continuous across the segment junctions, where taking the nearest segment alone would introduce discontinuities. The two end terms of \eqref{eq:arc_length_extension} extend the arc length beyond the two endpoints along the end tangents, so that $\tilde{s}$ is defined on the whole domain. In terms of these two fields, the implemented crack embedding function reads
\begin{equation}
  \rho(\mathbf{x})
  =
  S(\mathbf{x})\mathcal{W}_{c}(\mathbf{x}),
  \qquad
  S(\mathbf{x})
  =
  \operatorname{sgn}
  \left(
    \tilde{d}(\mathbf{x})
  \right),
  \label{eq:rho_polyline}
\end{equation}
where the side indicator $S$ plays the role of the sign term in \eqref{eq:weighted_sign_form} and $\mathcal{W}_{c}$ is a continuous localization weight,
\begin{equation}
  \mathcal{W}_{c}(\mathbf{x})
  =
  \left[
    \min
    \left(
      1,\,
      \max
      \left(
        0,\,
        \frac{L_c+\delta_c-\tilde{s}(\mathbf{x})}{L_c+\delta_c}
      \right)
    \right)
  \right]^{2}
  \exp
  \left(
    -\alpha_{\rho}
    \left(
      \frac{|\tilde{d}(\mathbf{x})|}{\ell_{\rho}}
    \right)^{p_{\rho}}
  \right),
  \label{eq:implemented_localisation_weight}
\end{equation}
where $\delta_c$ is a small endpoint overlap, and $\alpha_{\rho}$, $p_{\rho}$ and the reference length $\ell_{\rho}$ control the normal decay, with the values of Table~\ref{tab:discretization}. The first factor is the squared window of \eqref{eq:relu_weight} expressed in arc-length coordinates. It is held at one at the first control point $\mathbf{p}_0$ and behind it, decreases along the crack, and vanishes at a distance $\delta_c$ beyond the active tip, so that the embedded jump closes continuously toward the tip, mirroring the physical crack opening, while the near-tip enrichment of Section~\ref{sec:neural_energy_formulation} supplies the opening in the immediate tip neighborhood. The overlap keeps the embedded jump open up to the tip itself. In exchange, the network can place a small discontinuity over the distance $\delta_c$ ahead of the tip, where the exact fields are continuous, and its size is measured in Appendix~\ref{app:references}.

\begin{figure}[htbp]
  \centering
  \includegraphics[width=\textwidth]{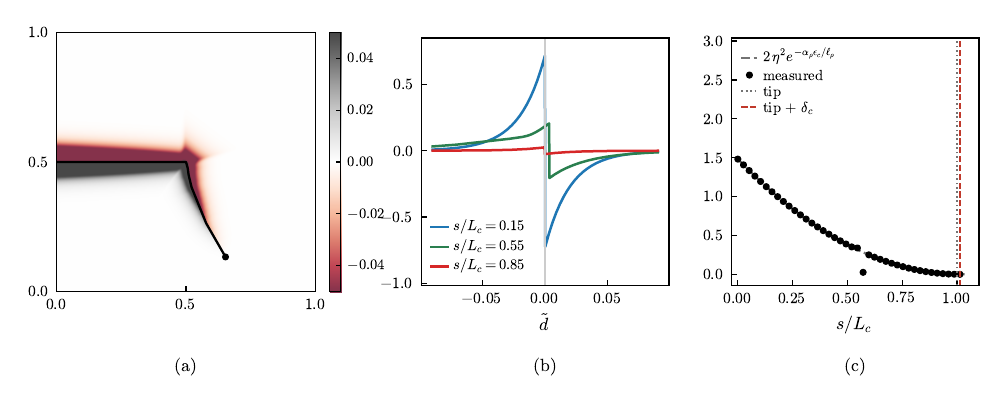}
  \caption{The crack embedding function \eqref{eq:rho_polyline}--\eqref{eq:implemented_localisation_weight} on the crack at load step 70 of a heated graded shear run of Section~\ref{subsec:sens_fgm} ($\alpha_\rho=50$, $p_\rho=1$, $\delta_c=0.0125$). (a) $\rho$ changes sign across the crack and decays away from it, the color scale clipped at $\pm0.05$ so that the sign change stays visible toward the tip. (b) Transverse profiles at three stations $s/L_c$, with $s$ the arc length. The jump is largest where the crack meets the boundary and nearly closed at the tip. (c) The embedded jump sampled at the offset $\pm\epsilon_c=0.006$, as markers, against $2\eta^{2}e^{-\alpha_\rho\epsilon_c/\ell_\rho}$, as the dashed line, where $\eta$ is the arc-length window of \eqref{eq:implemented_localisation_weight} before squaring. The zero just past the notch corner is the corner effect of Section~\ref{subsec:explicit_crack_geometry}.}
  \label{fig:crack_embedding}
\end{figure}

This construction is the weighted sign form \eqref{eq:weighted_sign_form} with $\phi=\tilde{d}$ and $\mathcal{W}=\mathcal{W}_c$, and for a single segment it reduces to the factorization \eqref{eq:rho_xdem_form} up to two changes. The exponential in \eqref{eq:implemented_localisation_weight} decays away from the crack where that of \eqref{eq:f2_localisation} grows, which keeps the embedding bounded as the crack propagates without changing the jump across it, and the endpoint window is written in arc length rather than through the two endpoint lines, which is what lets it follow a polyline. Which face receives the positive sign of $S$ is a convention tied to the orientation of the segment normals. Fig.~\ref{fig:crack_embedding} evaluates the implemented function on a kinked polyline. The sign change follows the crack, the transverse decay follows $\alpha_\rho$, and the measured jump tracks the window formula and closes at $\delta_c$ beyond the tip.

The crack-aware network input is then
\begin{equation}
  \tilde{\mathbf{x}}
  =
  \left\{
    x,
    y,
    \rho(\mathbf{x})
  \right\}^{\mathrm{T}} .
  \label{eq:crack_aware_input}
\end{equation}
Both the displacement and temperature networks use this augmented input, so that the two fields are represented with respect to the same crack geometry.

\subsection{Local crack-tip frame}
\label{subsec:local_crack_tip_coordinate}

A local coordinate frame is attached to the active crack tip in order to evaluate the near-tip enrichment. The tip tangent is written as
\begin{equation}
  \mathbf{t}_{\mathrm{tip}}
  =
  \left\{
    \cos\beta,
    \sin\beta
  \right\}^{\mathrm{T}},
\end{equation}
where $\beta$ is the angle between the tip tangent and the global $x$ axis. The corresponding normal is
\begin{equation}
  \mathbf{n}_{\mathrm{tip}}
  =
  \left\{
    -\sin\beta,
    \cos\beta
  \right\}^{\mathrm{T}} .
\end{equation}
For a point $\mathbf{x}=\{x,y\}^{\mathrm{T}}$, its relative position with respect to the active tip is
\begin{equation}
  \mathbf{r}
  =
  \mathbf{x}
  -
  \mathbf{x}_{\mathrm{tip}},
\end{equation}
and its projections onto the tangent and normal are
\begin{equation}
  x_1
  =
  \mathbf{r}\cdot\mathbf{t}_{\mathrm{tip}},
  \qquad
  x_2
  =
  \mathbf{r}\cdot\mathbf{n}_{\mathrm{tip}}.
\end{equation}
The local polar coordinates are therefore
\begin{equation}
  r
  =
  \sqrt{x_1^2+x_2^2},
  \qquad
  \theta
  =
  \operatorname{atan2}(x_2,x_1).
  \label{eq:local_tip_coordinates}
\end{equation}
These coordinates are used in the crack-tip asymptotic enrichment described in Section~\ref{sec:neural_energy_formulation}.

\section{Enriched deep energy formulation}
\label{sec:neural_energy_formulation}

The formulation follows the Deep Energy Method (DEM). The temperature and the displacement are represented by separate crack-aware networks. The incremental functionals \eqref{eq:thermal_incremental_functional} and \eqref{eq:mechanical_potential} are the losses, and they are minimized with derivatives obtained by automatic differentiation~\cite{samaniego2020,nguyenthanh2020,goswami2020tl}.

The networks take the physical coordinates of the example, which their first layer maps affinely onto $[-1,1]^2$. Every length in the formulation, the embedding decay, the face offsets, the contour radius and the enrichment decay, is therefore stated in physical units, as Table~\ref{tab:discretization} lists them. In what follows $x$ and $y$ denote the physical coordinates, $L_d$ the width of the bounding box of the specimen and $L_c$ the current crack length. The displacement output is scaled by a reference displacement $U_{\mathrm{ref}}$ and the temperature output by a reference increment $\Delta T_{\mathrm{ref}}$, each of the order of the applied quantity, and all reported values are converted back. Scalings of this kind are routine in network-based solvers, and the expressions below are written in physical variables.

The crack embedding introduced in Section~\ref{sec:crack_representation} is supplied as an additional input to both networks through the crack-aware input $\tilde{\mathbf{x}}$ of \eqref{eq:crack_aware_input}.
The displacement network is a multilayer perceptron with hyperbolic tangent activations. The temperature network has the same form except for its first layer, which applies a sinusoidal map with map parameter $\omega_0=30$, in the spirit of Sinusoidal Representation Networks (SIRENs)~\cite{sitzmann2020}, to help resolve steep thermal gradients. Nothing below depends on that choice, and the layer widths are listed in Table~\ref{tab:discretization}.

Essential boundary conditions, when present, are imposed through hard constraints~\cite{samaniego2020,sukumar2022}. The network correction is multiplied by a lifting function that vanishes on the constrained boundary, so the prescribed values are satisfied exactly. For prescribed values on the lower and upper edges of a unit square, one may take
\begin{equation}
  B(y)
  =
  y(1-y),
  \qquad
  0\leq y\leq 1 .
  \label{eq:boundary_lifting_function}
\end{equation}
Other boundary layouts require a corresponding choice of lifting function and prescribed extension.

\subsection{Temperature approximation}
\label{subsec:temperature_approximation_energy}

Let $\widehat{T}_{\boldsymbol{\theta}_T}(\tilde{\mathbf{x}})$ denote the output of the thermal network, where $\boldsymbol{\theta}_T$ collects its trainable weights and biases and $\tilde{\mathbf{x}}$ is the crack-aware input \eqref{eq:crack_aware_input}. The temperature at time step $t_n$ is approximated as
\begin{equation}
  T(\mathbf{x},t_n)
  =
  \bar{T}(\mathbf{x},t_n)
  +
  D_T(\mathbf{x})\,
  \widehat{T}_{\boldsymbol{\theta}_T}(\tilde{\mathbf{x}}),
  \label{eq:temperature_general_approximation}
\end{equation}
where $\bar{T}$ is a prescribed extension satisfying the essential thermal boundary conditions and $D_T$ is a lifting function that vanishes on the prescribed thermal boundary. If the reference temperature is imposed on the lower edge and a temperature increment is imposed on the upper edge, one may choose, with $B$ from \eqref{eq:boundary_lifting_function},
\begin{equation}
  \bar{T}(\mathbf{x},t_n)
  =
  T_0+y\, \Delta T_n,
  \qquad
  D_T(\mathbf{x})
  =
  B(y),
\end{equation}
which gives
\begin{equation}
  T(\mathbf{x},t_n)
  =
  T_0
  +
  y\, \Delta T_n
  +
  B(y)\,
  \widehat{T}_{\boldsymbol{\theta}_T}(\tilde{\mathbf{x}}).
  \label{eq:temperature_approximation}
\end{equation}
This construction enforces $T=T_0$ at $y=0$ and $T=T_0+\Delta T_n$ at $y=1$. Here, $T_0$ is the reference temperature and $\Delta T_n$ is the prescribed temperature increment at time step $t_n$. Other thermal boundary conditions can be accommodated by modifying $\bar{T}$ and $D_T$.

The temperature receives no near-tip enrichment, although its asymptotic behavior at the tip matches that of the displacement, $\sqrt{r}$ with an $r^{-1/2}$ gradient. The reason lies in what is evaluated from each field. The displacement enters the interaction integral of Section~\ref{subsec:crack_tip_quantities} through its strains on the contour, so an error in its $\sqrt{r}$ term propagates directly into $K_I$ and $K_{II}$, which is why the enrichment of Section~\ref{subsec:displacement_approximation_energy} supplies that term. The temperature enters the growth model only through its values near the tip, through the thermal strain in the mechanical loss, and through its gradient in the area term \eqref{eq:thermal_domain_term} of the interaction integral, where the singular gradient is integrable. Thermal enrichment functions are available if the singular gradient itself has to be resolved~\cite{duflot2008}. The choice is tested on the stationary benchmark of Section~\ref{subsec:wang_verification} and against an independent conduction solution in Section~\ref{subsec:sens_fgm}.

The jump across the adiabatic faces is the one thermal feature a smooth network cannot represent, and the shared embedding admits it by giving the two faces distinct inputs. The zero normal flux on the faces is not imposed through the embedding. It is the natural condition of the thermal functional of Section~\ref{subsec:thermal}, which the minimization satisfies in the weak sense, supplemented by the small consistency term of Section~\ref{subsec:transient_thermal_energy_loss}.

\subsection{Thermal energy loss}
\label{subsec:transient_thermal_energy_loss}

\begin{figure}[htbp]
  \centering
  \includegraphics[width=0.46\textwidth]{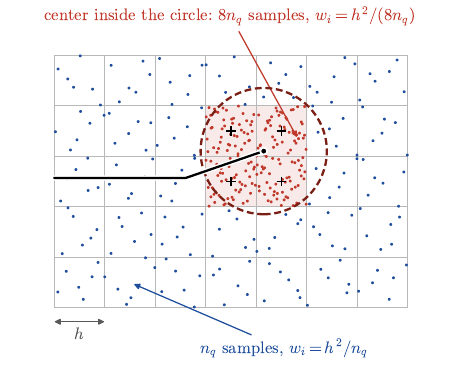}
  \caption{The integration scheme. The bounding box of the specimen is partitioned into square background elements of side $h$, each with $n_q$ Latin hypercube samples of weight $h^2/n_q$. An element whose center, marked by a cross, lies within a set radius of the crack tip, the dashed circle, receives a multiple of $n_q$ samples, each weight divided by the same multiple, so that every element still integrates to its own area. The multiple is $8$ in the sketch and on the notched tests and $48$ on the cruciform. The set is redrawn at a fixed interval during training.}
  \label{fig:integration_scheme}
\end{figure}

At time step $t_n$, the thermal network is trained by minimizing the time discrete heat conduction functional. Let
\begin{equation}
  \mathcal{Q}_T
  =
  \left\{
    \mathbf{x}_i^T,
    w_i^T
  \right\}_{i=1}^{N_T^q}
\end{equation}
denote the thermal integration points and weights. The integrals are estimated by Monte Carlo integration on a background mesh, sketched in Fig.~\ref{fig:integration_scheme}. The bounding box of the specimen is partitioned into square elements of side $h$, and each element receives a fixed number $n_q$ of samples drawn within it by Latin hypercube sampling, so that the samples are stratified by element. The weight of a sample is the area of its element divided by the number of samples in that element, $w_i^T=h^2/n_q$, so that every element integrates to its own area and the sum in \eqref{eq:thermal_loss} is an unbiased estimate of the functional. The mesh has no connectivity and no shape functions, and its role is to stratify the samples. With every element sampled, the variance of the estimate is no larger than that of unstratified sampling over the whole domain~\cite{caflisch1998}, and no draw leaves a region wider than $h$ without a point, which bounds the size of a feature that the loss cannot observe. The element is also the unit in which the near-tip densification of Fig.~\ref{fig:integration_scheme} is stated. Concentrating integration points near the tip, where the integrand varies most rapidly, is common practice in related methods, such as the dedicated quadrature for the tip enrichment of XFEM~\cite{bechet2005} and the local refinement along the crack in the Deep Energy Method for phase-field modeling of brittle fracture~\cite{goswami2020fourth}. Let $T_i^{n-1}$ be the temperature at the previous time step evaluated at the same integration points, and define
\begin{equation}
  T_i^n
  =
  T(\mathbf{x}_i^T,t_n).
\end{equation}

The discrete thermal loss is the quadrature evaluation of the incremental thermal functional \eqref{eq:thermal_incremental_functional} with $Q=0$ and $\bar{q}=0$, supplemented by a crack-face consistency term,
\begin{equation}
  \mathcal{L}_T^n
  =
  \sum_{i=1}^{N_T^q}
  w_i^T
  \left[
    \frac{\varrho C_p}{2\Delta t}
    \left( T_i^n - T_i^{n-1} \right)^2
    +
    \frac{k}{2}
    \left\| \nabla T_i^n \right\|^2
  \right]
  +
  \mathcal{L}_{q,c}^{n}.
  \label{eq:thermal_loss}
\end{equation}
If volumetric sources or prescribed fluxes are present, the corresponding linear terms of \eqref{eq:thermal_incremental_functional} are restored.

The integration points are redrawn periodically during training on the same background elements. The resampling is essential for the thermal loss. With a fixed point set the optimizer can develop a spurious transition layer that threads between the integration points, the kind of incorrect, approximately zero-energy solution reported by Manav et al.~\cite{manav2024}, which arises when the representation contains more localized detail than the integration rule can observe. Redrawing the points at a fixed interval removes the fixed gaps in which such a layer could hide. The same resampling is applied to the mechanical integration set $\mathcal{Q}_u$, where it plays a milder stabilizing role.

The crack faces are adiabatic, and the zero-flux condition is naturally associated with the energy statement. As a numerical safeguard, the small heat flux consistency term in \eqref{eq:thermal_loss} is defined as
\begin{equation}
  \mathcal{L}_{q,c}^{n}
  =
  \frac{\lambda_q}{N_q}
  \sum_{j=1}^{N_q}
  \left[
    \mathbf{q}^{n}
    \left(
      \mathbf{x}_{q,j}
    \right)
    \cdot
    \mathbf{n}_{q,j}
  \right]^2 ,
  \label{eq:adiabatic_crack_loss}
\end{equation}
where $\mathbf{x}_{q,j}$ are points sampled on the crack faces, $\mathbf{n}_{q,j}$ are the associated face normals, and
\begin{equation}
  \mathbf{q}^{n}
  =
  -k\nabla T(\mathbf{x},t_n).
\end{equation}
The weight $\lambda_q$ is kept small so that this term acts only as a weak consistency regularizer, rather than as the primary enforcement mechanism.

The trainable thermal parameters are the network weights and biases,
\begin{equation}
  \boldsymbol{\vartheta}_T
  =
  \left\{
    \boldsymbol{\theta}_T
  \right\} .
  \label{eq:thermal_trainable_set}
\end{equation}
The thermal problem is therefore written as
\begin{equation}
  \boldsymbol{\vartheta}_T^n
  =
  \operatorname*{arg\,min}_{\boldsymbol{\vartheta}_T}
  \mathcal{L}_T^n .
  \label{eq:thermal_minimization}
\end{equation}

\subsection{Enriched displacement approximation}
\label{subsec:displacement_approximation_energy}
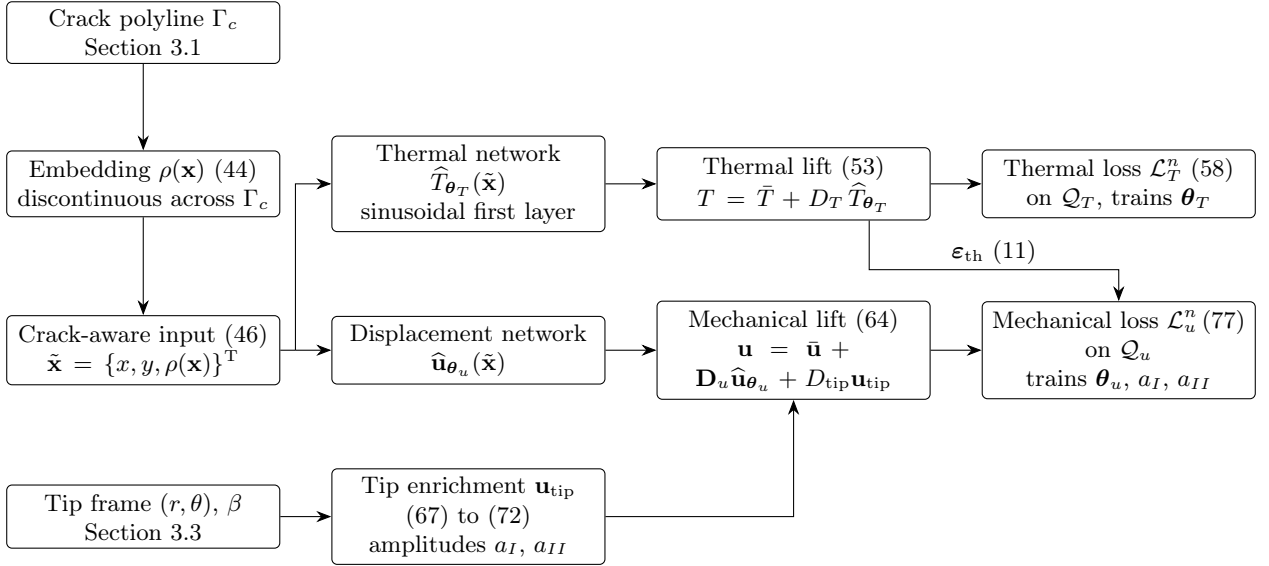
\begin{figure}[htbp]
\centering
\begin{tikzpicture}[
  font=\footnotesize,
  box/.style={rectangle, rounded corners=2pt, draw, align=center, inner xsep=3pt, inner ysep=3pt, text width=34mm},
  line/.style={-{Stealth[length=2.0mm]}}
]
\node[box] (poly) at (0, 1.1) {Crack polyline $\Gamma_c$\\ Section~\ref{subsec:explicit_crack_geometry}};
\node[box] (rho) at (0, -0.9) {Embedding $\rho(\mathbf{x})$~\eqref{eq:rho_polyline}\\ discontinuous across $\Gamma_c$};
\node[box] (input) at (0, -3.1) {Crack-aware input~\eqref{eq:crack_aware_input}\\ $\tilde{\mathbf{x}}=\{x,y,\rho(\mathbf{x})\}^{\mathrm{T}}$};
\node[box] (frame) at (0, -5.3) {Tip frame $(r,\theta)$, $\beta$\\ Section~\ref{subsec:local_crack_tip_coordinate}};
\node[box] (netT) at (4.3, -0.9) {Thermal network\\ $\widehat{T}_{\boldsymbol{\theta}_T}(\tilde{\mathbf{x}})$\\ sinusoidal first layer};
\node[box] (netU) at (4.3, -3.1) {Displacement network\\ $\widehat{\mathbf{u}}_{\boldsymbol{\theta}_u}(\tilde{\mathbf{x}})$};
\node[box] (enr) at (4.3, -5.3) {Tip enrichment $\mathbf{u}_{\mathrm{tip}}$\\ \eqref{eq:local_tip_enrichment_compact} to \eqref{eq:mechanical_tip_decay}\\ amplitudes $a_I$, $a_{II}$};
\node[box] (liftT) at (8.6, -0.9) {Thermal lift~\eqref{eq:temperature_general_approximation}\\ $T=\bar{T}+D_T\,\widehat{T}_{\boldsymbol{\theta}_T}$};
\node[box] (liftU) at (8.6, -3.1) {Mechanical lift~\eqref{eq:displacement_general_approximation}\\ $\mathbf{u}=\bar{\mathbf{u}}+\mathbf{D}_u\widehat{\mathbf{u}}_{\boldsymbol{\theta}_u}+D_{\mathrm{tip}}\mathbf{u}_{\mathrm{tip}}$};
\node[box] (lossT) at (12.9, -0.9) {Thermal loss $\mathcal{L}_T^n$~\eqref{eq:thermal_loss}\\ on $\mathcal{Q}_T$, trains $\boldsymbol{\theta}_T$};
\node[box] (lossU) at (12.9, -3.1) {Mechanical loss $\mathcal{L}_u^n$\,\eqref{eq:mechanical_loss_vector}\\ on $\mathcal{Q}_u$\\ trains $\boldsymbol{\theta}_u$, $a_I$, $a_{II}$};
\draw[line] (poly) -- (rho);
\draw[line] (rho) -- (input);
\coordinate (fork) at ($(input.east)+(0.2,0)$);
\draw (input.east) -- (fork);
\draw[line] (fork) |- (netT.west);
\draw[line] (fork) -- (netU.west);
\draw[line] (frame) -- (enr);
\draw[line] (netT) -- (liftT);
\draw[line] (netU) -- (liftU);
\draw[line] (enr.east) -| (liftU.south);
\draw[line] (liftT) -- (lossT);
\draw[line] (liftU) -- (lossU);
\draw[line] ($(liftT.south)+(1.0,0)$) -- ++(0,-0.65) -| node[pos=0.25, above, inner sep=2pt]{$\boldsymbol{\varepsilon}_{\mathrm{th}}$~\eqref{eq:thermal_strain}} (lossU.north);
\end{tikzpicture}
\caption{Structure of the enriched approximation. The crack polyline defines the embedding, which enters both networks through the crack-aware input. Both network outputs are lifted onto the essential conditions, and the displacement is enriched near the tip by the Williams modes with trainable amplitudes. The thermal and mechanical losses are minimized in sequence, as set out in Section~\ref{subsec:staggered_training_procedure}, the trained temperature entering the mechanical loss through the thermal strain.}
\label{fig:enrichment_anatomy}
\end{figure}

Once the temperature field has been determined, the displacement field is represented by a separate crack-aware network with trainable parameters $\boldsymbol{\theta}_u$. Two elements of what follows are inherited. The first is the embedded crack input itself, which allows a single global network to represent a displacement jump and is the construction of the discontinuity-embedded networks of Zhao and Shao~\cite{zhao2025denns} and of the energy formulation built on them~\cite{zhao2024dedem}. The second is the asymptotic enrichment with trainable amplitudes, classical in XFEM and combined with an embedded crack in an isothermal energy setting by Wang et al.~\cite{wang2025xdem}. The network output is written as
\begin{equation}
  \widehat{\mathbf{u}}_{\boldsymbol{\theta}_u}(\tilde{\mathbf{x}})
  =
  \left[
    \widehat{u}_{\boldsymbol{\theta}_u}(\tilde{\mathbf{x}}),
    \widehat{v}_{\boldsymbol{\theta}_u}(\tilde{\mathbf{x}})
  \right]^{\mathrm{T}},
\end{equation}
where $\tilde{\mathbf{x}}$ is again the crack-aware input \eqref{eq:crack_aware_input}. The displacement approximation at time step $t_n$ is defined as
\begin{equation}
  \mathbf{u}(\mathbf{x},t_n)
  =
  \bar{\mathbf{u}}(\mathbf{x},t_n)
  +
  \mathbf{D}_u(\mathbf{x})\,\widehat{\mathbf{u}}_{\boldsymbol{\theta}_u}(\tilde{\mathbf{x}})
  +
  D_{\mathrm{tip}}(\mathbf{x})\,\mathbf{u}_{\mathrm{tip}}(\mathbf{x}).
  \label{eq:displacement_general_approximation}
\end{equation}
Here, $\bar{\mathbf{u}}$ is a prescribed extension satisfying the essential displacement boundary conditions, $\mathbf{D}_u$ is the lifting of the network correction, a scalar function or a diagonal matrix when the two displacement components are constrained differently, $\mathbf{u}_{\mathrm{tip}}$ is the near-tip enrichment defined below, and $D_{\mathrm{tip}}$ is a scalar lifting that vanishes on every constrained boundary. As in the thermal case \eqref{eq:temperature_general_approximation}, the liftings keep the essential conditions exactly satisfied. The enrichment receives one scalar lifting for both components, which preserves the angular structure of the Williams modes near the tip, and the matrix form of $\mathbf{D}_u$ is therefore confined to the network correction. For example, if the displacements $\bar{\mathbf{u}}_{\mathrm{b}}^n$ and $\bar{\mathbf{u}}_{\mathrm{t}}^n$ are prescribed on the lower and upper edges, respectively, the extension may be chosen as
\begin{equation}
  \bar{\mathbf{u}}(\mathbf{x},t_n)
  =
  (1-y)\bar{\mathbf{u}}_{\mathrm{b}}^n
  +
  y\, \bar{\mathbf{u}}_{\mathrm{t}}^n,
  \label{eq:displacement_extension}
\end{equation}
with $\mathbf{D}_u=D_{\mathrm{tip}}=B(y)$ from \eqref{eq:boundary_lifting_function}, so that the trainable correction and the enrichment vanish on the prescribed boundaries. The displacement approximation then becomes
\begin{equation}
  \mathbf{u}(\mathbf{x},t_n)
  =
  (1-y)\bar{\mathbf{u}}_{\mathrm{b}}^n
  +
  y\, \bar{\mathbf{u}}_{\mathrm{t}}^n
  +
  B(y)
  \left[
    \widehat{\mathbf{u}}_{\boldsymbol{\theta}_u}(\tilde{\mathbf{x}})
    +
    \mathbf{u}_{\mathrm{tip}}(\mathbf{x})
  \right].
  \label{eq:displacement_lifted_approximation}
\end{equation}
Other displacement boundary conditions can be handled by modifying $\bar{\mathbf{u}}$, $\mathbf{D}_u$ and $D_{\mathrm{tip}}$ accordingly. The whole approximation, from the crack polyline to the two losses, is drawn in Fig.~\ref{fig:enrichment_anatomy}.

The crack-tip contribution is introduced to represent the leading asymptotic displacement behavior near the crack tip. In the local crack-tip coordinate system, indicated by the subscript $\ell$, the enriched displacement field is written as
\begin{equation}
  \mathbf{u}_{\ell}^{\mathrm{tip}}(r,\theta)
  =
  a_I \boldsymbol{\Phi}_I(r,\theta)
  +
  a_{II} \boldsymbol{\Phi}_{II}(r,\theta),
  \label{eq:local_tip_enrichment_compact}
\end{equation}
where $(r,\theta)$ are the local polar coordinates centered at the crack tip, and $a_I$ and $a_{II}$ are trainable scalar amplitudes associated with the mode I and mode II asymptotic fields. These amplitudes have the dimensions of SIFs but are not used as such, since their trained values also absorb the local value of the lifting at the tip and the decay introduced below. The SIFs used for propagation are extracted from the converged fields by the interaction integral of Section~\ref{sec:crack_propagation}, on a contour away from the tip. The corresponding functions $\boldsymbol{\Phi}_I$ and $\boldsymbol{\Phi}_{II}$ are the Williams mode I and mode II displacement fields~\cite{williams1957}. Both carry the factor $\sqrt{r}$, so their gradients carry $r^{-1/2}$ and it is the enrichment that supplies the singular strain the network cannot form on its own:
\begin{equation}
  \boldsymbol{\Phi}_I(r,\theta)
  =
  \frac{1}{2\mu}
  \sqrt{\frac{r}{2\pi}}
  \begin{bmatrix}
    \cos\frac{\theta}{2}
    \left(
      \kappa-1
      +
      2\sin^2\frac{\theta}{2}
    \right)
    \\
    \sin\frac{\theta}{2}
    \left(
      \kappa+1
      -
      2\cos^2\frac{\theta}{2}
    \right)
  \end{bmatrix},
  \label{eq:mode_I_tip_function}
\end{equation}
and
\begin{equation}
  \boldsymbol{\Phi}_{II}(r,\theta)
  =
  \frac{1}{2\mu}
  \sqrt{\frac{r}{2\pi}}
  \begin{bmatrix}
    \sin\frac{\theta}{2}
    \left(
      2+\kappa+\cos\theta
    \right)
    \\
    \cos\frac{\theta}{2}
    \left(
      2-\kappa-\cos\theta
    \right)
  \end{bmatrix}.
  \label{eq:mode_II_tip_function}
\end{equation}
Here, $\mu$ is the shear modulus and $\kappa$ is the Kolosov constant, with $\kappa=3-4\nu$ for plane strain and $\kappa=(3-\nu)/(1+\nu)$ for plane stress.

\begin{figure}[htbp]
  \centering
  \includegraphics[width=0.42\textwidth]{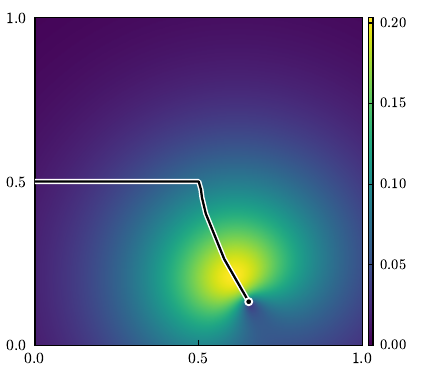}
  \caption{Magnitude of the mode I enrichment with its decay, computed from \eqref{eq:mode_I_tip_function} and \eqref{eq:mechanical_tip_decay} with the tension test's $m_u=5$ and $q_u=1$, on the kinked crack of Fig.~\ref{fig:crack_embedding}.}
  \label{fig:enrichment_decay}
\end{figure}

The local enriched displacement is rotated back to the global coordinate system by
\begin{equation}
  \mathbf{u}_{\mathrm{tip}}^{0}(\mathbf{x})
  =
  \mathbf{R}(\beta)
  \mathbf{u}_{\ell}^{\mathrm{tip}}(r,\theta),
  \label{eq:global_tip_enrichment_raw}
\end{equation}
where $\beta$ is the angle between the local crack-tip tangent and the global $x$ axis. Since the Williams expansion is intended only to enrich the near-tip response, the contribution is localized by a decay function,
\begin{equation}
  \mathbf{u}_{\mathrm{tip}}(\mathbf{x})
  =
  \mathcal{D}_u(r)
  \mathbf{u}_{\mathrm{tip}}^{0}(\mathbf{x}),
  \label{eq:mechanical_tip_contribution}
\end{equation}
with
\begin{equation}
  \mathcal{D}_u(r)
  =
  \exp
  \left[
    -m_u
    \frac{L_d}{L_c}
    r^{q_u}
  \right],
  \label{eq:mechanical_tip_decay}
\end{equation}
where $L_c$ is the current crack length, $L_d$ the width of the bounding box, and $m_u$ and $q_u$ are prescribed localization parameters. Scaling the exponent by $L_d/L_c$ widens the region in which the enrichment acts as the crack grows, so that the represented near-tip zone remains proportional to the crack size. The decay localizes the enrichment without giving it a finite support. The angle $\theta$ of \eqref{eq:local_tip_enrichment_compact} is measured from the tip tangent, so the Williams functions are discontinuous across the backward extension of that tangent, which coincides with the crack only up to the last corner. Behind a corner the enrichment therefore places an attenuated jump on a line that is not the crack. The mismatch behind the corner is left to the minimization, and the extraction contour of Section~\ref{subsec:crack_tip_quantities} stays on the straight stretch that ends at the tip, where the two lines coincide. The magnitude of the mode I enrichment with its decay is shown in Fig.~\ref{fig:enrichment_decay}.

\subsection{Mechanical energy loss}
\label{subsec:mechanical_energy_loss}
For a fixed crack configuration and a trained temperature field, the mechanical network is obtained by minimizing the thermoelastic potential energy. Let
\begin{equation}
  \mathcal{Q}_u
  =
  \left\{
    \mathbf{x}_i^u,
    w_i^u
  \right\}_{i=1}^{N_u^q}
\end{equation}
denote the mechanical integration points and weights. This set is drawn independently of the thermal set $\mathcal{Q}_T$, on the same background elements with the same element weights, and is resampled in the same way. At each integration point, the strain tensor \eqref{eq:small_strain} is evaluated from the displacement field as
\begin{equation}
  \boldsymbol{\varepsilon}_i^n
  =
  \boldsymbol{\varepsilon}
  \left(
    \mathbf{u}(\mathbf{x}_i^u,t_n)
  \right)
  =
  \frac{1}{2}
  \left[
    \nabla \mathbf{u}(\mathbf{x}_i^u,t_n)
    +
    \nabla \mathbf{u}^{\mathrm{T}}(\mathbf{x}_i^u,t_n)
  \right].
  \label{eq:strain_tensor_neural}
\end{equation}
The elastic strain is then obtained by subtracting the thermal strain, as in \eqref{eq:elastic_strain},
\begin{equation}
  \boldsymbol{\varepsilon}_{\mathrm{e},i}^n
  =
  \boldsymbol{\varepsilon}_i^n
  -
  \boldsymbol{\varepsilon}_{\mathrm{th},i}^n,
  \qquad
  \boldsymbol{\varepsilon}_{\mathrm{th},i}^n
  =
  \alpha
  \left(
    T_i^n-T_0
  \right)
  \mathbf{I},
  \label{eq:elastic_strain_tensor_neural}
\end{equation}
where
\begin{equation}
  T_i^n
  =
  T(\mathbf{x}_i^u,t_n).
\end{equation}
The discrete mechanical loss is therefore written as
\begin{equation}
  \mathcal{L}_{u}^{n}
  =
  \sum_{i=1}^{N_u^q}
  w_i^u
  \,
  \psi_{\mathrm{e}}
  \left(
    \boldsymbol{\varepsilon}_{\mathrm{e},i}^n
  \right)
  -
  W_{\mathrm{ext}}^{n},
  \label{eq:mechanical_loss_vector}
\end{equation}
where $\psi_{\mathrm{e}}$ is the elastic strain energy density \eqref{eq:elastic_energy_density} of the thermoelastic material and $W_{\mathrm{ext}}^{n}$ is the external work \eqref{eq:external_work} evaluated with the discrete displacement. For displacement-controlled cases without body forces or prescribed tractions, the external work term $W_{\mathrm{ext}}^{n}$ vanishes.

Since the essential displacement conditions are built into the admissible approximation, no Dirichlet penalty terms are required. The traction-free crack-face condition remains a natural condition of the mechanical energy statement. The trainable mechanical parameters are
\begin{equation}
  \boldsymbol{\vartheta}_u
  =
  \left\{
    \boldsymbol{\theta}_u,
    a_I,
    a_{II}
  \right\},
  \label{eq:mechanical_trainable_set}
\end{equation}
which collect the network parameters $\boldsymbol{\theta}_u$ and the Williams amplitudes $a_I$ and $a_{II}$. The mechanical problem is therefore
\begin{equation}
  \boldsymbol{\vartheta}_u^n
  =
  \operatorname*{arg\,min}_{\boldsymbol{\vartheta}_u}
  \mathcal{L}_{u}^{n}.
  \label{eq:mechanical_minimization}
\end{equation}
During this minimization, the temperature field is fixed.

\subsection{Staggered solution procedure}
\label{subsec:staggered_training_procedure}

The staggered sequence in which the two fields are trained is the energy-based counterpart of the operator split solvers commonly used in phase-field modeling of brittle fracture~\cite{miehe2010cmame}. The complete cycle, including the crack growth decision of Section~\ref{sec:crack_propagation}, is summarized in the flow chart of Fig.~\ref{fig:solution_flowchart}.

At time step $t_n$, the crack $\Gamma_c$ is fixed and its embedding $\rho(\mathbf{x})$ is constructed. Given the previous temperature field $T(\mathbf{x},t_{n-1})$, the thermal parameters are obtained by minimizing
\begin{equation}
  \boldsymbol{\vartheta}_T^n
  =
  \operatorname*{arg\,min}_{\boldsymbol{\vartheta}_T}
  \mathcal{L}_T^n ,
  \label{eq:staggered_thermal_training}
\end{equation}
after which $T(\mathbf{x},t_n)$ is fixed and supplied as the thermal loading for the mechanical solve,
\begin{equation}
  \boldsymbol{\vartheta}_u^n
  =
  \operatorname*{arg\,min}_{\boldsymbol{\vartheta}_u}
  \mathcal{L}_u^n .
  \label{eq:staggered_mechanical_training}
\end{equation}
Within one step, the solution sequence can be summarized as
\begin{equation}
  \Gamma_c
  \rightarrow
  \rho(\mathbf{x})
  \rightarrow
  T(\mathbf{x},t_n)
  \rightarrow
  \mathbf{u}(\mathbf{x},t_n)
  \rightarrow
  \text{crack-tip quantities}.
  \label{eq:staggered_solution_sequence}
\end{equation}

Both minimizations are carried out with Adam~\cite{kingma2014adam}, followed by L-BFGS iterations on the last draw of the points only at the unloaded step 0 and in the stationary benchmark of Section~\ref{subsec:wang_verification}. The periodic resampling of Section~\ref{subsec:transient_thermal_energy_loss} replaces the objective between iterations, so a quasi-Newton refinement on a fixed objective, such as L-BFGS~\cite{liu1989lbfgs}, is not used within the resampling schedule, whereas Adam receives the change of the objective as gradient noise~\cite{zhang2026meshfree}. Each minimization runs until the mean loss over one resampling interval stops improving by a set relative amount, or until its iteration budget is spent, with the values given in Appendix~\ref{app:discretization}. Automatic differentiation supplies all spatial derivatives required in the thermal gradient, strains, stresses, and energy densities. At each new time step or crack increment, both networks are initialized from the converged parameters of the previous configuration. Since the fields change only incrementally, this warm start substantially reduces the number of iterations per step, in the same spirit as the transfer learning strategy of~\cite{goswami2020tl}.

Crack-tip quantities are extracted from the trained fields after the mechanical solve, and if the propagation criterion of Section~\ref{sec:crack_propagation} is met the crack is advanced, the embedding is rebuilt and the sequence is repeated on the new geometry. Throughout that repetition the previous temperature field $T^{n-1}$ is kept fixed, so the extensions within a load step repeat the solution at the same time level.

\section{Crack driving force and propagation}
\label{sec:crack_propagation}

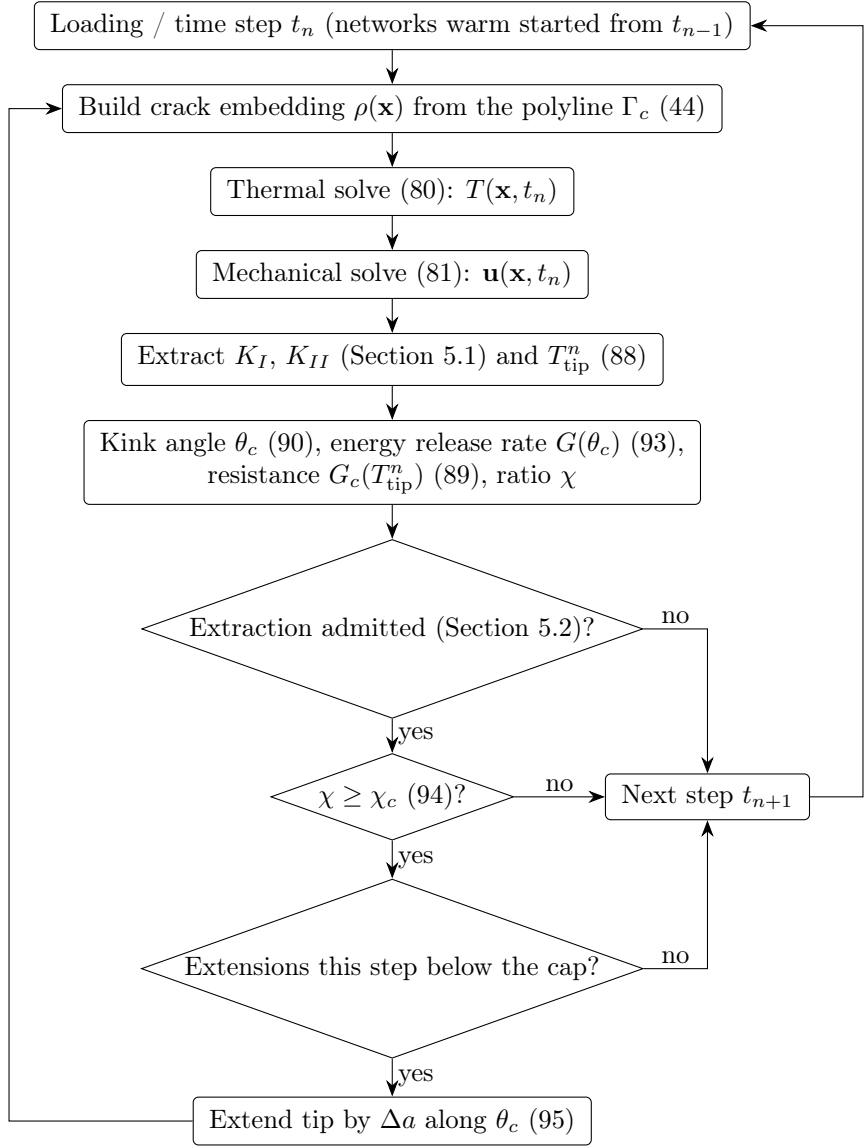
\begin{figure}[t]
\centering
\begin{tikzpicture}[
  font=\small,
  node distance=4.5mm and 12mm,
  box/.style={rectangle, rounded corners=2pt, draw, align=center, inner xsep=6pt, inner ysep=4pt},
  dec/.style={diamond, aspect=2.8, draw, align=center, inner sep=1pt},
  line/.style={-{Stealth[length=2.2mm]}}
]
\node[box] (step) {Loading / time step $t_n$ (networks warm started from $t_{n-1}$)};
\node[box, below=of step] (embed) {Build crack embedding $\rho(\mathbf{x})$ from the polyline $\Gamma_c$~\eqref{eq:rho_polyline}};
\node[box, below=of embed] (thermal) {Thermal solve~\eqref{eq:staggered_thermal_training}: $T(\mathbf{x},t_n)$};
\node[box, below=of thermal] (mech) {Mechanical solve~\eqref{eq:staggered_mechanical_training}: $\mathbf{u}(\mathbf{x},t_n)$};
\node[box, below=of mech] (extract) {Extract $K_I$, $K_{II}$ (Section~\ref{subsec:crack_tip_quantities}) and $T_{\mathrm{tip}}^{n}$~\eqref{eq:crack_tip_temperature}};
\node[box, below=of extract] (drive) {Kink angle $\theta_c$~\eqref{eq:max_hoop_angle}, energy release rate $G(\theta_c)$~\eqref{eq:kinked_energy_release_rate},\\ resistance $G_c(T_{\mathrm{tip}}^{n})$~\eqref{eq:temperature_dependent_gc}, ratio $\chi$};
\node[dec, below=of drive] (adm) {Extraction admitted (Section~\ref{subsec:extraction_admissibility})?};
\node[dec, below=of adm] (crit) {$\chi\geq\chi_c$~\eqref{eq:crack_propagation_condition}?};
\node[dec, below=5mm of crit] (cap) {Extensions this step below the cap?};
\node[box, below=5mm of cap] (grow) {Extend tip by $\Delta a$ along $\theta_c$~\eqref{eq:crack_tip_update}};
\node[box, right=of crit] (next) {Next step $t_{n+1}$};
\draw[line] (step) -- (embed);
\draw[line] (embed) -- (thermal);
\draw[line] (thermal) -- (mech);
\draw[line] (mech) -- (extract);
\draw[line] (extract) -- (drive);
\draw[line] (drive) -- (adm);
\draw[line] (adm) -- node[right, inner sep=2pt]{yes} (crit);
\draw[line] (crit) -- node[right, inner sep=2pt]{yes} (cap);
\draw[line] (cap) -- node[right, inner sep=2pt]{yes} (grow);
\draw[line] (crit) -- node[above, inner sep=2pt]{no} (next);
\draw[line] (cap.east) -| node[pos=0.25, above, inner sep=2pt]{no} (next.south);
\draw[line] (adm.east) -| node[pos=0.25, above, inner sep=2pt]{no} (next.north);
\path (embed.west) ++(-7mm,0) coordinate (loopback);
\draw[line] (grow.west) -| (loopback) -- (embed.west);
\draw[line] (next.east) -- ++(7mm,0) |- (step.east);
\end{tikzpicture}
\caption{Flow chart of the staggered solution and crack growth cycle. The ratio $\chi$ is formed before the admissibility checks of Section~\ref{subsec:extraction_admissibility}, since the first check acts on it. A reading that fails the checks authorizes no extension, except under the bounded release for a clipped sweep. While \eqref{eq:crack_propagation_condition} holds and the extensions in the step are below their cap, the crack is extended and the fields are retrained on the new embedding.}
\label{fig:solution_flowchart}
\end{figure}

The crack propagation treatment is set within Linear Elastic Fracture Mechanics (LEFM)~\cite{griffith1921,irwin1957}. Once the two fields have been trained for the current crack, the growth decision is taken from the trained fields alone, in the steps set out in the subsections below. The mixed-mode SIFs $K_I$ and $K_{II}$ are extracted at the tip. The crack-tip temperature is read from the thermal field and gives the fracture resistance $G_c$. The SIFs give the kink angle $\theta_c$ and the energy release rate $G(\theta_c)$ of a kink in that direction. The extraction is then checked for admissibility, and a reading that fails the check authorizes no extension, except under the bounded release for a clipped sweep of Section~\ref{subsec:extraction_admissibility}. Otherwise the crack is extended along $\theta_c$ whenever $G(\theta_c)$ reaches $G_c$, after which the embedding is rebuilt and the fields are retrained. Every extension has the same length, the crack increment $\Delta a$, which is a numerical parameter of the method and is given per example in Table~\ref{tab:discretization}. The cycle is summarized in the flow chart of Fig.~\ref{fig:solution_flowchart}.

\subsection{Extraction of the stress intensity factors}
\label{subsec:crack_tip_quantities}

The SIFs are extracted with the interaction integral~\cite{yau1980,shih1986}, in the form that holds for a body with a non-uniform temperature field~\cite{wilson1979,duflot2008,kckim2008}. Let $(x_1,x_2)$ denote the local tip coordinates of Section~\ref{subsec:local_crack_tip_coordinate}, let $\Gamma_r$ be a circle of radius $r$ around the tip with outward unit normal $\mathbf{n}$, and let $A_r$ be the disc that it encloses. Superposing the trained fields with a known auxiliary field $(\mathbf{u}^{\mathrm{aux}},\boldsymbol{\sigma}^{\mathrm{aux}})$ gives the interaction integral as the sum of a contour term and an area term,
\begin{equation}
  I
  =
  I_{\Gamma}
  +
  M^{\mathrm{th}} ,
  \label{eq:interaction_total}
\end{equation}
with
\begin{equation}
  I_{\Gamma}
  =
  \int_{\Gamma_r}
  \left[
    W^{(1,2)}
    \delta_{1j}
    -
    \sigma_{ij}
    \frac{\partial u_i^{\mathrm{aux}}}{\partial x_1}
    -
    \sigma_{ij}^{\mathrm{aux}}
    \frac{\partial u_i}{\partial x_1}
  \right]
  n_j
  \, d\Gamma ,
  \qquad
  W^{(1,2)}
  =
  \frac{1}{2}
  \left(
    \sigma_{kl}\varepsilon_{kl}^{\mathrm{aux}}
    +
    \sigma_{kl}^{\mathrm{aux}}\varepsilon_{\mathrm{e},kl}
  \right),
  \label{eq:interaction_integral}
\end{equation}
and
\begin{equation}
  M^{\mathrm{th}}
  =
  \int_{A_r}
  \sigma_{kk}^{\mathrm{aux}}
  \,
  \frac{\partial\varepsilon_{\mathrm{th}}}{\partial x_1}
  \,
  \mathrm{d}A ,
  \label{eq:thermal_domain_term}
\end{equation}
where summation over repeated indices is implied and all quantities are expressed in the local tip frame. The strain of the trained field that enters $I_{\Gamma}$ is the elastic strain $\boldsymbol{\varepsilon}_{\mathrm{e}}$ of \eqref{eq:elastic_strain}, with the stress from the thermoelastic law \eqref{eq:thermoelastic_stress}, whereas the auxiliary field is isothermal, so its strain carries no thermal part and is written without the subscript. In the area term, $\varepsilon_{\mathrm{th}}=\alpha(T-T_0)$ is the scalar thermal strain of \eqref{eq:thermal_strain}, carrying in plane strain the same effective coefficient $(1+\nu)\alpha$ as Section~\ref{subsec:constitutive}, and $\sigma^{\mathrm{aux}}_{kk}$ is the in-plane trace of the auxiliary stress. The area term is the contribution that Wilson and Yu~\cite{wilson1979} derived for the thermoelastic $J$-integral, written here in its interaction form. It arises since the elastic energy density depends on position through the temperature, and it vanishes for a uniform temperature, so that the isothermal cases reduce to the contour term. Its contribution is discussed in Appendix~\ref{app:thermal_area_term}. Choosing the auxiliary field as the pure mode I or pure mode II Williams solution with unit SIF yields
\begin{equation}
  K_I
  =
  \frac{E'}{2}
  I^{(I)},
  \qquad
  K_{II}
  =
  \frac{E'}{2}
  I^{(II)},
  \qquad
  J
  =
  \frac{K_I^2+K_{II}^2}{E'},
  \label{eq:sif_extraction}
\end{equation}
where $E'=E/(1-\nu^2)$ for plane strain and $E'=E$ for plane stress.

In finite element practice the interaction integral is converted into an equivalent domain integral over a ring of elements around the tip, with a weight function that vanishes on its outer boundary~\cite{shih1986}, since the finite element stresses are discontinuous across element boundaries and a line integral through the elements is inaccurate. Neural networks are not subject to this limitation, since their derivatives are obtained by automatic differentiation at any point. The contour term is therefore evaluated directly on the circle $\Gamma_r$ by the trapezoidal rule with $2880$ equally spaced points, with neither background elements nor a weight function, and the area term over the whole disc $A_r$ by the midpoint rule on a polar grid of $80$ radial by $360$ angular intervals. The trace $\sigma^{\mathrm{aux}}_{kk}$ is singular at the tip but integrable against the $r\,\mathrm{d}r\,\mathrm{d}\theta$ measure, so no core region is excluded. The accuracy of this evaluation is checked on the stationary benchmark of Section~\ref{subsec:wang_verification} and, at every load step, by the radius sweep of Section~\ref{subsec:extraction_admissibility}.

For a graded material the auxiliary fields are evaluated with the material properties at the tip, as detailed with the graded test of Section~\ref{subsec:sens_fgm}. When the tip is so close to a boundary that no contour fits between them, the SIFs are estimated instead from the displacement jump between the crack faces behind the tip. This estimate is a diagnostic. It has no sweep behind it, so it never passes the admissibility check of Section~\ref{subsec:extraction_admissibility} and never advances the crack.

\begin{figure}[htbp]
  \centering
  \includegraphics[width=\textwidth]{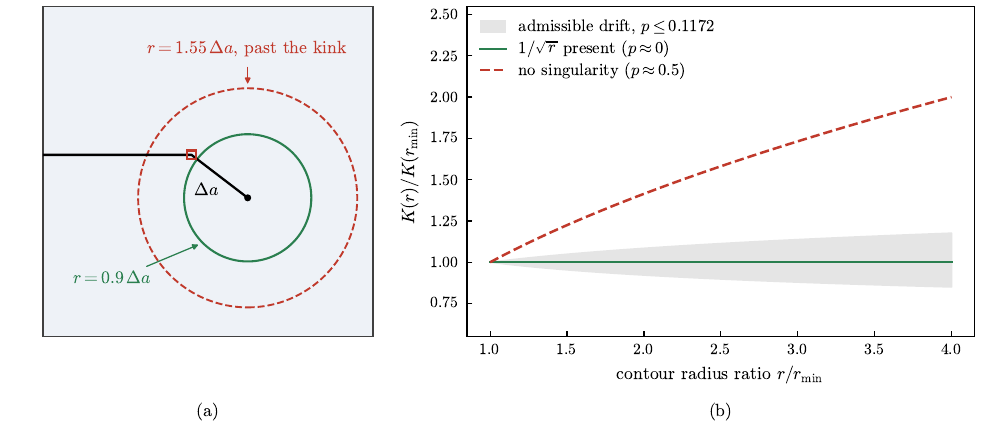}
  \caption{The two geometric rules of the extraction. (a) The nominal contour $r=0.9\,\Delta a$ stays inside the straight stretch that ends at the tip. A larger contour, such as $r=1.55\,\Delta a$, dashed, would cross the kink, and the auxiliary field, a straight-crack solution, would no longer match the crack it encloses. (b) Schematic of the radius sweep, flat for a field with the $1/\sqrt{r}$ singularity and drifting as $\sqrt{r}$ for a field without it, with the admissible drift $p<0.1172$, $15\%$ across a fourfold span, shaded.}
  \label{fig:contour_gates}
\end{figure}

The auxiliary field of \eqref{eq:interaction_integral} is the Williams solution for a straight crack, so the contour should not enclose a corner of the polyline. The stretch that ends at the tip is treated as straight as long as the segment directions stay within $5^{\circ}$ of the direction of the last segment. After a kink this stretch is one crack increment $\Delta a$ long, and the contour radius is kept below it, which ties the radius to the crack increment rather than to the specimen size. The nominal radius is $r=0.9\,\Delta a$ in every propagating example, as sketched in Fig.~\ref{fig:contour_gates}(a). A contour that crosses the corner does not fail visibly. It returns a finite value evaluated with an auxiliary field that does not match the crack behind the contour, and such a value can place the post-kink mode I SIF about a third below the kinked crack solution, arrest the crack and produce a spurious rise in the reaction force.

\subsection{Admissibility of an extraction}
\label{subsec:extraction_admissibility}

An extracted pair of SIFs enters the growth decision only if it passes the two checks below. A reading that fails either of them authorizes no extension and ends the growth cycle of that load step, and the criterion is evaluated again at the next load step. The number of load steps at which the checks refuse a reading is reported with each example.

The first check, called the convergence check below, acts on the criterion ratio $\chi$ of Section~\ref{subsec:crack_geometry_update}. Several extensions may be accepted within one load step, each followed by a partial retraining, and an extraction taken before the fields have settled is finite but does not yet describe the extended crack. An extraction is therefore rejected when its ratio exceeds a fixed multiple of the threshold, or exceeds the ratio of the previous extension of the same load step by more than a factor of two. The multiple is a set value, listed in Table~\ref{tab:discretization}, and in the tests below a settled and an unsettled extension differ by an order of magnitude in this ratio.

The second check is a sweep of the contour radius. A correctly extracted SIF does not depend on the radius, whereas a trained field that contains no $1/\sqrt{r}$ singularity returns a quantity that grows roughly as $\sqrt{r}$, so the radius dependence measures whether the extraction is admissible at all. The integral \eqref{eq:interaction_total} is evaluated on five radii, $0.5$, $0.75$, $1$, $1.5$ and $2$ times the nominal radius $r$ of Section~\ref{subsec:crack_tip_quantities}. A radius that would reach past the corner of the polyline, or beyond nine tenths of the distance from the tip to the nearest boundary, is dropped, so the set may hold fewer than five radii over a narrower span. The drift across the set is expressed as one exponent
\begin{equation}
  p
  =
  -\,\frac{
    \ln\!\left(1-d\right)
  }{
    \ln\!\left(r_{\max}/r_{\min}\right)
  },
  \label{eq:path_independence_exponent}
\end{equation}
where $d$ is the larger of the peak-to-peak variations of $K_I$ and $K_{II}$ across the set, divided by the largest SIF magnitude in it, so that a small mode II component cannot inflate the drift, and $r_{\max}$ and $r_{\min}$ are the largest and the smallest radius retained. A reading passes the sweep when $p<0.1172$, the value obtained by allowing a $15\%$ drift across a fourfold span, and the SIFs that then enter the criterion are the medians of $K_I$ and $K_{II}$ over the retained radii. A reading fails when fewer than two radii fit or when the SIF of larger magnitude changes sign across the set, and a field without the singularity is expected to give $p\approx0.5$ and to fail as well, as Fig.~\ref{fig:contour_gates}(b) illustrates. The sweep tests the singular structure of a reading rather than the completeness of the integral. Appendix~\ref{app:thermal_area_term} shows a reading without the area term that still passes on the stationary benchmark, where only the shape of the sweep, monotonic rather than flat, reveals the omission.

The dropping of radii has a consequence that the check allows for. A fresh kink leaves one increment of straight stretch, so the two largest radii are dropped and the retained set spans a factor of two only. A retained set that spans less than threefold is called clipped. Refusing every clipped sweep would stall the crack at each kink, since the straight stretch grows only when the crack moves. A reading that fails the sweep on a clipped set is therefore released to the growth decision, at most twice since the last sweep that passed over a wider span. A released reading advances the crack only if the criterion of Section~\ref{subsec:crack_geometry_update} is then met, and the readings released in this way and the extensions taken on them are counted with each example.

\subsection{Crack-tip temperature and fracture resistance}
\label{subsec:temperature_dependent_fracture_resistance}

The crack-tip temperature is read from the trained thermal field. The exact temperature is continuous at the tip but its gradient is singular there, so a value read at the tip itself depends on how well the network resolves that gradient. The temperature is therefore taken as the mean over a small ring centered on the tip,
\begin{equation}
  T_{\mathrm{tip}}^{n}
  =
  \frac{1}{N_T}
  \sum_{k=1}^{N_T}
  T\!\left(\mathbf{x}_{\mathrm{tip}}+r_T\,\mathbf{e}(\phi_k),\,t_n\right),
  \label{eq:crack_tip_temperature}
\end{equation}
with $N_T=24$ equally spaced angles $\phi_k$, $\mathbf{e}(\phi_k)$ the unit vector at that angle, ring points falling outside the specimen discarded, and the radius $r_T$ given per example in Table~\ref{tab:discretization}. The ring straddles both crack faces, so the mean averages the two face temperatures behind the tip. The comparison with an independent conduction solution in Section~\ref{subsec:sens_fgm} supports this choice over the point value.

The material resistance to crack growth may vary with temperature, which is incorporated through a temperature-dependent critical energy release rate~\cite{ruan2023thermomechanical},
\begin{equation}
  G_c(T)
  =
  G_{c0}
  \left[
    1
    -
    b_1
    \frac{
      T-T_0
    }{
      T_{\max}
    }
    +
    b_2
    \left(
      \frac{
        T-T_0
      }{
        T_{\max}
      }
    \right)^2
  \right],
  \label{eq:temperature_dependent_gc}
\end{equation}
evaluated at the crack-tip temperature \eqref{eq:crack_tip_temperature}, where $G_{c0}$ is the fracture resistance at the reference temperature $T_0$, $T_{\max}$ is a reference maximum temperature, and $b_1$ and $b_2$ are material parameters controlling the temperature dependence. We use $b_1=1.80$ and $b_2=1.10$ in the notched tension and shear tests, with $T_0$ and $T_{\max}$ taken as 300 K and 1000 K. With these values the bracketed factor stays above $0.26$ at every temperature, so the resistance remains positive. The cruciform test of Section~\ref{subsec:cruciform} uses a constant resistance, as stated there. For the functionally graded shear test, $G_{c0}$ is replaced by the linearly graded value $G_{c0}(x_{\mathrm{tip}})$ obtained from the interpolation \eqref{eq:sens_fgm_interpolation}, so that the spatial and thermal dependence of the resistance act multiplicatively.

\subsection{Growth direction and driving force}
\label{subsec:crack_growth_direction}

The growth direction is determined from the extracted SIFs by the maximum hoop stress criterion~\cite{erdogan1963}. The kink angle $\theta_c$, measured from the current tip tangent, is
\begin{equation}
  \theta_c
  =
  2
  \tan^{-1}
  \left(
    \frac{
      K_I
      -
      \sqrt{
        K_I^2+8K_{II}^2
      }
    }{
      4K_{II}
    }
  \right)
  \quad\text{for } K_{II}\neq 0,
  \qquad
  \theta_c=0
  \quad\text{for } K_{II}=0 ,
  \label{eq:max_hoop_angle}
\end{equation}
so that a crack under pure mode I advances along its current tangent. The magnitude of $\theta_c$ is capped at the value listed in Table~\ref{tab:discretization}, to avoid an excessive change of direction within a single crack increment. With $\beta$ denoting the angle between the current tip tangent and the global $x$ axis, the global growth direction is
\begin{equation}
  \mathbf{d}_c
  =
  \left\{
    \cos(\beta+\theta_c),
    \sin(\beta+\theta_c)
  \right\}^{\mathrm{T}} .
  \label{eq:global_crack_direction}
\end{equation}

The driving force is the energy released by an infinitesimal kink in the direction $\theta_c$, and not the rate of straight growth, since a crack that is about to turn does not release energy along its current tangent. Testing a mixed-mode criterion on the kink that the crack is about to take is the criterion of Nuismer~\cite{nuismer1975}. To first order, the SIFs at the tip of an infinitesimal kink at angle $\theta$ are~\cite{cotterell1980}
\begin{equation}
  \begin{aligned}
    k_I(\theta)
    &=
    \cos^3\!\left(\tfrac{\theta}{2}\right) K_I
    -
    3\cos^2\!\left(\tfrac{\theta}{2}\right)\sin\!\left(\tfrac{\theta}{2}\right) K_{II} ,
    \\[2pt]
    k_{II}(\theta)
    &=
    \cos^2\!\left(\tfrac{\theta}{2}\right)\sin\!\left(\tfrac{\theta}{2}\right) K_I
    +
    \cos\!\left(\tfrac{\theta}{2}\right)
    \left[
      1-3\sin^2\!\left(\tfrac{\theta}{2}\right)
    \right] K_{II} ,
  \end{aligned}
  \label{eq:kink_local_sifs}
\end{equation}
where the lower case SIFs $k_I$ and $k_{II}$ belong to the crack after the turn and $K_I$ and $K_{II}$ to the crack before it. The full expansion is given by Amestoy and Leblond~\cite{amestoy1992crack}. The corresponding energy release rate is
\begin{equation}
  G(\theta)
  =
  \frac{
    k_I^2(\theta) + k_{II}^2(\theta)
  }{
    E'
  } .
  \label{eq:kinked_energy_release_rate}
\end{equation}
Both relations are standard and are used unchanged. They are evaluated at $\theta=\theta_c$ with the extracted SIFs, and the resulting $G(\theta_c)$ is the driving force that enters the propagation criterion of Section~\ref{subsec:crack_geometry_update}. For pure mode I, $\theta_c=0$ and $G(\theta_c)$ equals the straight-ahead rate $J$ of \eqref{eq:sif_extraction}. For pure mode II the angle $|\theta_c|=\arccos(1/3)\approx70.53^\circ$ gives $k_I=(2/\sqrt{3})|K_{II}|$ and $k_{II}=0$, so that $G(\theta_c)=\tfrac{4}{3}K_{II}^2/E'$ and the turn releases a third more energy than straight growth. Where the cap of Table~\ref{tab:discretization} limits the angle, the capped angle is used for the direction and for the driving force alike, so the two cannot disagree. The choice of $G(\theta_c)$ over $J$ matters under thermal load. The straight-ahead rate $J$~\cite{rice1968} is even in both SIFs, so it does not distinguish a thermal stress that opens the crack from one of the same size that closes it, whereas $G(\theta_c)$ contains the product $K_IK_{II}$ through \eqref{eq:kink_local_sifs} and distinguishes the two states whenever $K_{II}\neq0$.

\subsection{Propagation criterion and crack update}
\label{subsec:crack_geometry_update}

The crack advances while the ratio $\chi$ of the driving force to the resistance meets or exceeds a prescribed threshold $\chi_c$,
\begin{equation}
  \chi
  \equiv
  \frac{
    G(\theta_c)
  }{
    G_c(T_{\mathrm{tip}}^{n})
  }
  \geq
  \chi_c .
  \label{eq:crack_propagation_condition}
\end{equation}
The classical Griffith condition is recovered by setting $\chi_c=1$, which is the value used in every example reported below. For a stationary straight crack under pure mode I with a tensile $K_I$, \eqref{eq:crack_propagation_condition} reduces to the familiar $K_I=\sqrt{E'G_c}$. The criterion does not by itself exclude growth from a closed state, which only a contact condition would, in line with the open crack assumption of Section~\ref{subsec:mechanical_equilibrium}. None of the quantities entering the criterion is calibrated against the results reported below.

When the criterion is met, the crack is advanced explicitly by appending a new control point to the crack polyline at
\begin{equation}
  \mathbf{x}_{\mathrm{tip}}^{new}
  =
  \mathbf{x}_{\mathrm{tip}}
  +
  \Delta a\,
  \mathbf{d}_c .
  \label{eq:crack_tip_update}
\end{equation}
The embedding function $\rho(\mathbf{x})$ is then reconstructed from the new polyline. Since both networks share it, the updated geometry is passed consistently to the two fields, which are retrained from a warm start. The extraction, the checks and the criterion are then repeated on the new configuration within the same load step, and the crack is extended again while the criterion holds, up to the maximum number of extensions per load step listed in Table~\ref{tab:discretization}. The cycle ends for that load step when the criterion is no longer met, an extraction is refused or that number is reached, and the analysis moves to the next load step.

Growth ends when the crack tip approaches the far boundary. Once the remaining ligament is thinner than the contour radius, the circle $\Gamma_r$ no longer lies inside the material and the extraction ceases to be reliable. The specimen is therefore declared separated once the tip comes within a separation margin $\delta$ of the boundary, after which no further SIFs are extracted and the reaction force is set to zero. This is a termination rule of the computation, and the last increment before the physical separation of the two halves is not resolved. The margin and the resulting definition of the failure displacement are given with each example.

\section{Numerical examples}
\label{section:numerical_examples}

The method is assessed on four problems. The first holds the crack stationary in a configuration whose SIF has a published value, and verifies the extraction of Section~\ref{sec:crack_propagation} on its own. The remaining three let the crack grow: a single-edge notched plate in tension, its functionally graded shear counterpart, and a notched cruciform specimen under combined thermal and mechanical loading. For the two notched tests the phase-field references are computed with an in-house finite element solver. The shear test is compared in addition with a sharp-crack finite element solution computed for this study, and the cruciform with published paths and SIFs. In the displacement-controlled cases the reaction force is computed as the derivative of the total potential energy with respect to the imposed boundary displacement, evaluated by automatic differentiation. In the comparison figures of the propagating examples the reference solutions are labeled by their crack description, sharp crack or phase field, and the temperature shading in the geometry sketches is schematic. The load steps at which the admissibility checks of Section~\ref{subsec:extraction_admissibility} refused a reading, and the readings released on a clipped sweep, are reported with the notched tests and, for the cruciform, in Appendix~\ref{app:discretization}. The numerical parameters of all four are collected in Appendix~\ref{app:discretization}.

\subsection{Stress intensity factor of a stationary crack under thermal load}
\label{subsec:wang_verification}

\begin{figure}[htbp]
  \centering
  \includegraphics[width=0.4805\textwidth]{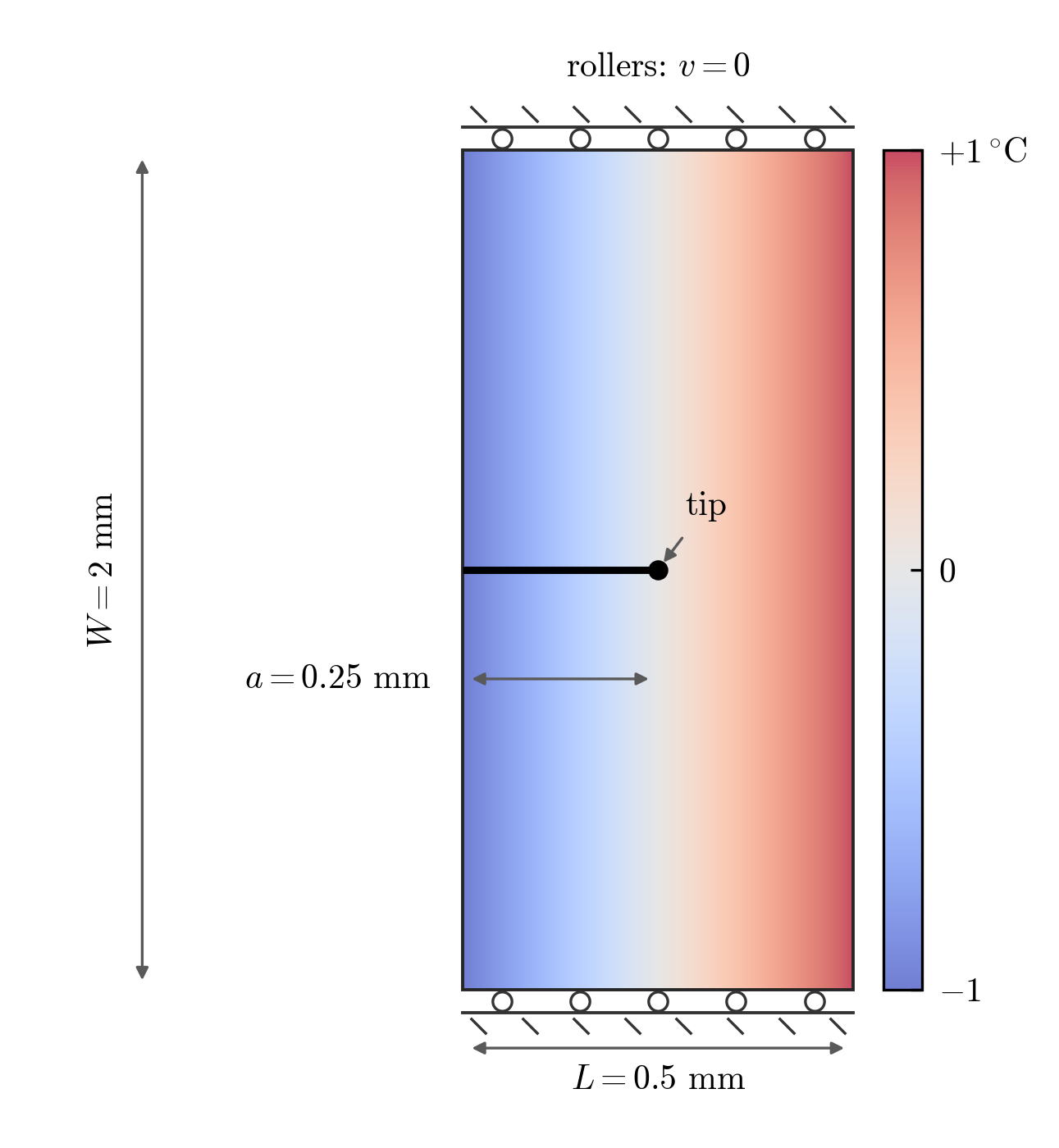}
  \caption{The thermal edge crack of Wang et al.~\cite{wang2023cmt}, with the imposed temperature difference drawn schematically. The rollers prevent the cooled face from contracting, which puts $\sigma_{yy}$ into tension across the crack.}
  \label{fig:wang_problem}
\end{figure}

Before any crack is allowed to move, the extracted SIF is checked against a published value. The configuration is the thermal edge crack of Wang et al.~\cite{wang2023cmt}, shown in Fig.~\ref{fig:wang_problem}: a strip of width $L=0.5$~mm and height $W=2$~mm, held at temperature changes of $-1$~K and $+1$~K relative to the stress-free reference on its two vertical faces, with rollers ($v=0$, $u$ free) on the top and bottom edges and an edge crack of length $a=0.25$~mm running from the cold face to the center line. The top and bottom edges are adiabatic, the two vertical faces are traction free, and the crack faces are adiabatic and traction free. Plane strain is assumed, the conduction problem is steady, as in the reference, and the strip is computed in its physical dimensions. Since the crack does not move, this example is trained on a single draw of the integration points, and its training ends with L-BFGS iterations after Adam, as Table~\ref{tab:discretization} records. The cooled face is prevented from contracting by the rollers, so $\sigma_{yy}$ goes into tension and the horizontal crack opens in pure mode I. The material data follow the reference, $E=218.4\times10^{3}$~MPa, $\nu=0.25$, $\alpha=1.67\times10^{-5}$~K$^{-1}$, and the extracted SIF is normalized by $K_0=E\alpha\,\Delta T_a\sqrt{\pi a}/(1-\nu)$, with $\alpha$ as quoted rather than its plane strain value $(1+\nu)\alpha$ and $\Delta T_a=1$~K the amplitude of the imposed change, for which the reference reports $K_I/K_0=0.500$. The temperature field in the strip is non-uniform, so the area term \eqref{eq:thermal_domain_term} of the interaction integral contributes to the extracted SIF.

\begin{figure}[htbp]
  \centering
  \includegraphics[width=\textwidth]{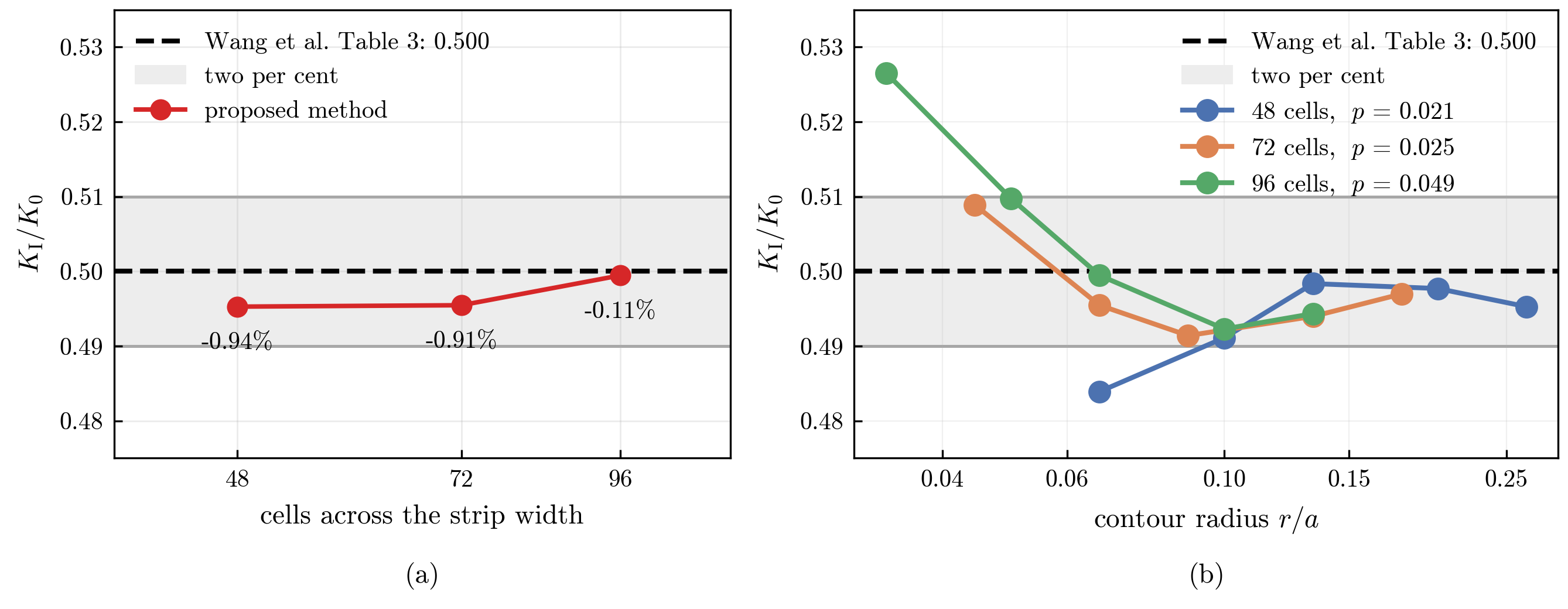}
  \caption{SIF of the stationary thermal edge crack, normalized by $K_0=E\alpha\,\Delta T_a\sqrt{\pi a}/(1-\nu)$ with $\alpha=1.67\times10^{-5}$~K$^{-1}$ and $\Delta T_a=1$~K. (a) The three grids against the published value $0.500$ of Wang et al.~\cite{wang2023cmt}, with a band of 2\%. (b) The radius sweep of each grid with its exponent $p$, all below the admissibility threshold $0.1172$. The innermost contour of the finest grid lies $5\%$ above the rest of its sweep.}
  \label{fig:wang_verification}
\end{figure}

Fig.~\ref{fig:wang_verification} collects the study. On grids of 48, 72 and 96 elements across the strip width, the extracted SIF is $K_I/K_0=0.4953$, $0.4955$ and $0.4995$, within $0.94\%$ of the published value on the coarsest grid and $0.11\%$ on the finest. The comparison is made over three grids rather than as a single reading, since the reference reports its own convergence study, with errors of $10.3\%$, $6.5\%$ and $0.6\%$ across three grids and path independence reached only on the refined one. Fig.~\ref{fig:wang_verification}(b) shows the radius sweep of Section~\ref{subsec:extraction_admissibility} at each resolution. Over a fourfold span of contour radii the reading drifts by $2.9\%$, $3.4\%$ and $6.5\%$, giving $p=0.021$, $0.025$ and $0.049$, so every sweep passes the admissibility check $p<0.1172$ of Section~\ref{subsec:extraction_admissibility}, the drift growing with refinement while the reading itself moves toward the published value.

\subsection{Single-edge notched tension test}
\label{subsec:sent}

A single-edge notched thermoelastic fracture benchmark considered by Tangella et al.~\cite{tangella2022hybrid} is adopted for the tensile test. This setting isolates a predominantly mode I fracture process and provides a direct check of the response under thermal loading.

\begin{figure}[htbp]
  \centering
  \includegraphics[width=\textwidth]{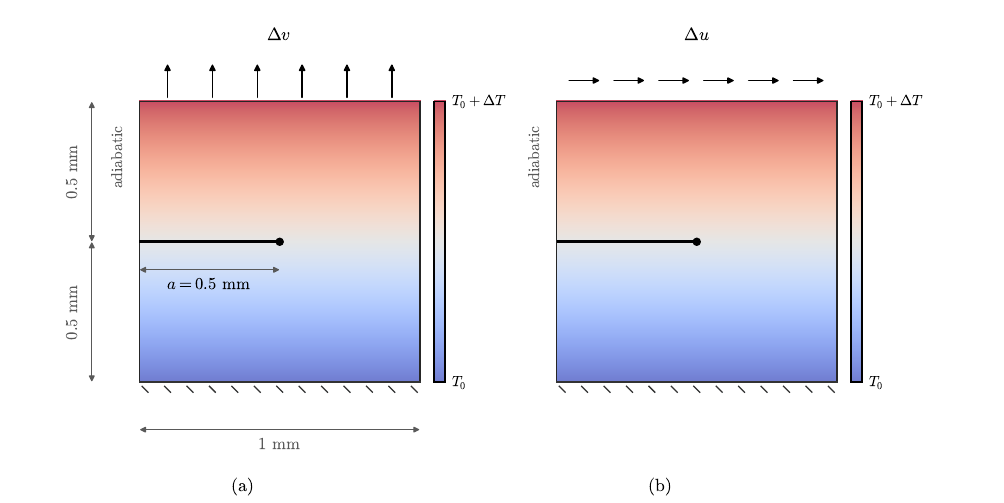}
  \caption{Geometry and boundary conditions of the single-edge notched specimen. (a) Tension, with a vertical displacement on the upper edge. (b) Shear, with a horizontal displacement on the upper edge. In both tests the lower edge is fixed and held at $T_0$, the upper edge is held at $T_0+\Delta T$ and the vertical edges are adiabatic. The temperature shading is schematic.}
  \label{fig:sen_geometry_bc}
\end{figure}

The geometry and boundary conditions are shown in Fig.~\ref{fig:sen_geometry_bc}(a). The specimen is a \(1~\mathrm{mm}\times1~\mathrm{mm}\) square plate with a horizontal edge notch of length \(0.5~\mathrm{mm}\) at mid-height, initially at \(T_0=300~\mathrm{K}\). The lower edge is fixed in both displacement components and held at \(T_0\), the upper edge receives a monotonically applied vertical displacement with its horizontal component held at zero and the temperature \(T_0+\Delta T\), and the two vertical edges are traction free and adiabatic. The reaction force is taken for a thickness of $1$~mm, so that it is in newtons. Three thermal cases are considered, \(\Delta T=-50~\mathrm{K}\), \(0~\mathrm{K}\) and \(+50~\mathrm{K}\), with the material parameters of Table~\ref{tab:sent_parameters}, which follow the reference benchmark and are used in a consistent mm--N--s--K unit system. Each case is solved over 106 load steps of \(10^{-5}~\mathrm{mm}\), with the temperature difference applied in equal increments over the first 20 steps and held from there on. The crack increment is \(\Delta a=0.025~\mathrm{mm}\), a fortieth of the specimen width, which by the rule of Section~\ref{subsec:crack_tip_quantities} puts the extraction contour at \(0.0225~\mathrm{mm}\).

\begin{table}[htbp]
  \centering
  \caption{Material parameters for the homogeneous single-edge notched tension test.}
  \label{tab:sent_parameters}
  \begin{tabular}{lll}
    \toprule
    Parameter & Symbol & Value \\
    \midrule
    Young's modulus & \(E\) & \(340~\mathrm{GPa}\) \\
    Poisson's ratio & \(\nu\) & \(0.22\) \\
    Critical energy release rate & \(G_c\) & \(42.47~\mathrm{J/m^2}\) \\
    Density & \(\varrho\) & \(2450~\mathrm{kg/m^3}\) \\
    Thermal conductivity & \(k\) & \(300~\mathrm{W/(m\,K)}\) \\
    Specific heat capacity & \(C_p\) & \(0.775~\mathrm{J/(kg\,K)}\) \\
    Thermal expansion coefficient & \(\alpha\) & \(8.0\times10^{-6}~\mathrm{K^{-1}}\) \\
    Reference temperature & \(T_0\) & \(300~\mathrm{K}\) \\
    \bottomrule
  \end{tabular}
\end{table}

The phase-field reference curves for both single-edge notched tests are computed with the solver of Appendix~\ref{app:references}, the AT2 model with regularization length $\ell=0.01~\mathrm{mm}$ on elements of size about $\ell/5$, with geometry, material data and boundary conditions matched to the corresponding run of the proposed method. On the tension test it reproduces the shift of initiation with the thermal load reported by Tangella et al.~\cite{tangella2022hybrid}, with peak loads $8\%$ to $13\%$ below theirs, a difference discussed in Appendix~\ref{app:references}. The reference and the proposed method differ in one respect. The reference uses the fracture energy of Table~\ref{tab:sent_parameters} as a constant, as the published statement of the test does, whereas the proposed method evaluates \eqref{eq:temperature_dependent_gc} at the tip temperature, which moves its critical SIF by $1.5\%$ between the cooled and the heated case, as reported below.

\begin{figure}[htbp]
  \centering
\begin{tikzpicture}
    \begin{axis}[
      width=0.74\textwidth,
      height=0.42\textwidth,
      xlabel={Displacement \((10^{-3}~\mathrm{mm})\)},
      ylabel={Force (N)},
      grid=both,
      xmin=-0.05,
      xmax=1.02,
      ymin=-5,
      ymax=125,
      xtick={0,0.2,0.4,0.6,0.8,1.0},
      tick label style={font=\small},
      label style={font=\small},
      legend style={
        font=\small,
        draw=none,
        fill=none,
        at={(0.5,-0.22)},
        anchor=north,
        /tikz/every even column/.append style={column sep=8pt}
      },
      legend columns=2,
      legend image post style={black},
      legend cell align=left
    ]

    \addplot[blue, dashed, thick]
      table[
        col sep=comma,
        x expr=\thisrowno{0}*1000,
        y index=1,
        skip first n=1
      ]
      {figures/data/sent_fem_m50.csv};
    \addlegendentry{phase field}

    \addplot[blue, solid, thick]
      table[
        col sep=comma,
        x expr=\thisrowno{0}*1000,
        y index=1,
        skip first n=1
      ]
      {figures/data/sent_xdem_m50.csv};
    \addlegendentry{proposed method}

    \addplot[black, dashed, thick]
      table[
        col sep=comma,
        x expr=\thisrowno{0}*1000,
        y index=1,
        skip first n=1
      ]
      {figures/data/sent_fem_0.csv};

    \addplot[black, solid, thick]
      table[
        col sep=comma,
        x expr=\thisrowno{0}*1000,
        y index=1,
        skip first n=1
      ]
      {figures/data/sent_xdem_0.csv};

    \addplot[red, dashed, thick]
      table[
        col sep=comma,
        x expr=\thisrowno{0}*1000,
        y index=1,
        skip first n=1
      ]
      {figures/data/sent_fem_p50.csv};

    \addplot[red, solid, thick]
      table[
        col sep=comma,
        x expr=\thisrowno{0}*1000,
        y index=1,
        skip first n=1
      ]
      {figures/data/sent_xdem_p50.csv};

    \end{axis}
  \end{tikzpicture}
  \caption{Force--displacement response of the single-edge notched tension test at \(\Delta T=-50\), \(0\) and \(+50~\mathrm{K}\), cooled in blue, isothermal in black and heated in red, with the phase-field reference dashed.}
  \label{fig:sent_fd}
\end{figure}
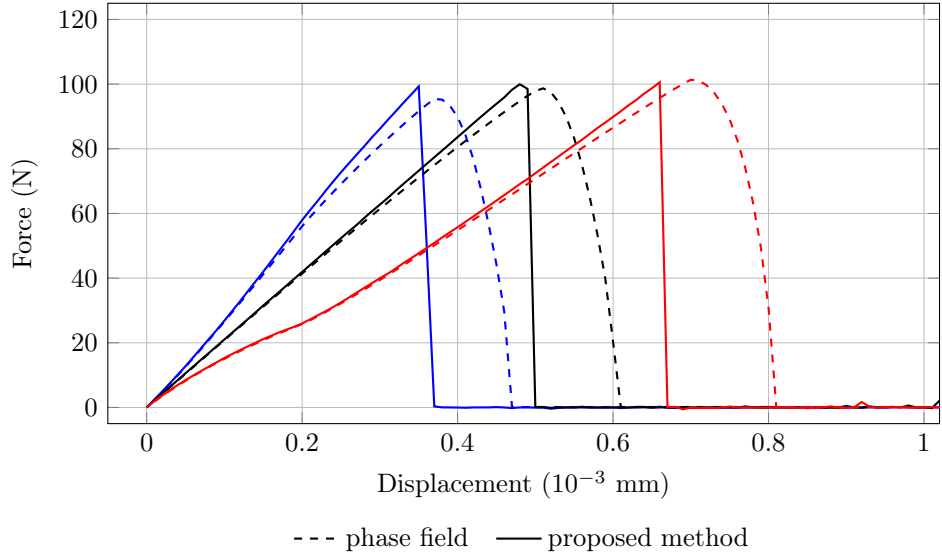

The force--displacement responses are compared with the finite element phase-field reference solutions in Fig.~\ref{fig:sent_fd}. The isothermal response remains linear until the crack initiates. Under thermal load the loading branch is steeper for the cooled specimen and shallower for the heated one, and the heated branch steepens once the temperature difference has been fully applied after 20 steps. The thermal load also moves the initiation point. Cooling brings it forward and heating delays it, which follows from the thermal strain. Expansion near the heated upper edge takes up part of the imposed displacement and reduces the elastic tensile strain at the notch, while cooling has the opposite effect. In this test the peak load falls at the last load step before the crack advances. The displacement at peak load moves from \(0.35\times10^{-3}~\mathrm{mm}\) at \(\Delta T=-50~\mathrm{K}\) to \(0.66\times10^{-3}~\mathrm{mm}\) at \(+50~\mathrm{K}\), an increase of 89\%, while the peak load itself changes by 1.3\% over the same range.

The peak load agrees with the reference to \(+4.1\%\), \(+1.3\%\) and \(-0.8\%\) at \(\Delta T=-50\), \(0\) and \(+50~\mathrm{K}\), with no systematic sign, and the displacement at which it is reached is 5.4\% to 5.9\% smaller than in the reference in all three cases. Repeats of all three cases move the peak by at most $1.8\%$ and initiation by at most one load step, which sets the resolution of this comparison at about $2\%$. After the peak the load drops within one or two load steps, and the ligament is fully separated 17\% to 21\% earlier in applied displacement. The post-peak branches differ by construction. A sharp crack whose resistance varies only with the tip temperature crosses the short remaining ligament as soon as the criterion is met, whereas the reference softens over a longer branch as its damage band approaches the far edge.

\begin{figure}[htbp]
  \centering
  \includegraphics[width=\textwidth]{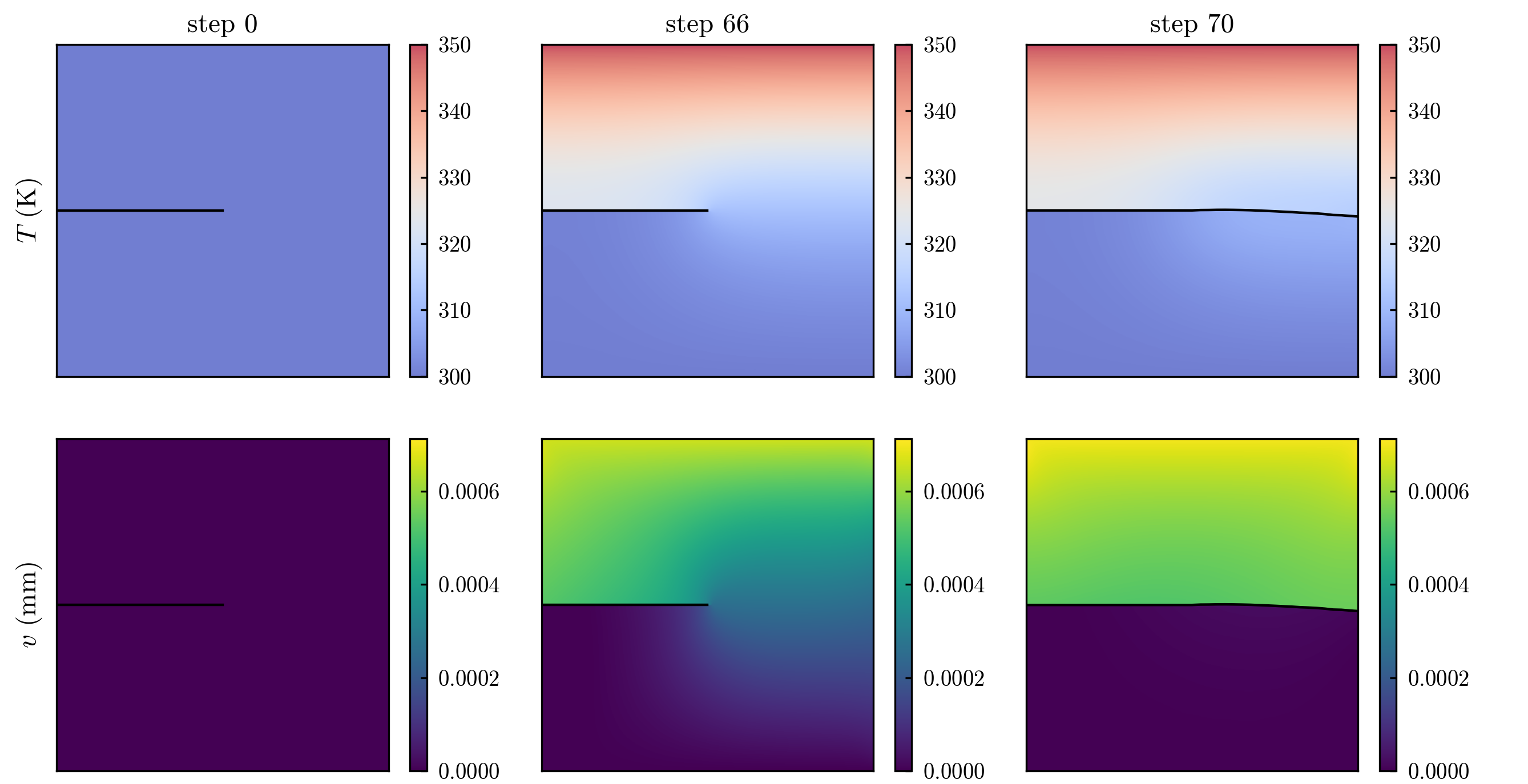}
  \caption{Temperature and vertical displacement in the single-edge notched tension test at \(\Delta T=+50~\mathrm{K}\), at load steps 0, 66 and 70 of 106. Step 66 is the last before the crack advances and step 70 follows separation. Each row shares one color scale. The region below the notch stays near the initial temperature since the crack faces are adiabatic.}
  \label{fig:sent_field_evolution}
\end{figure}

\begin{figure}[htbp]
  \centering
  \includegraphics[width=\textwidth]{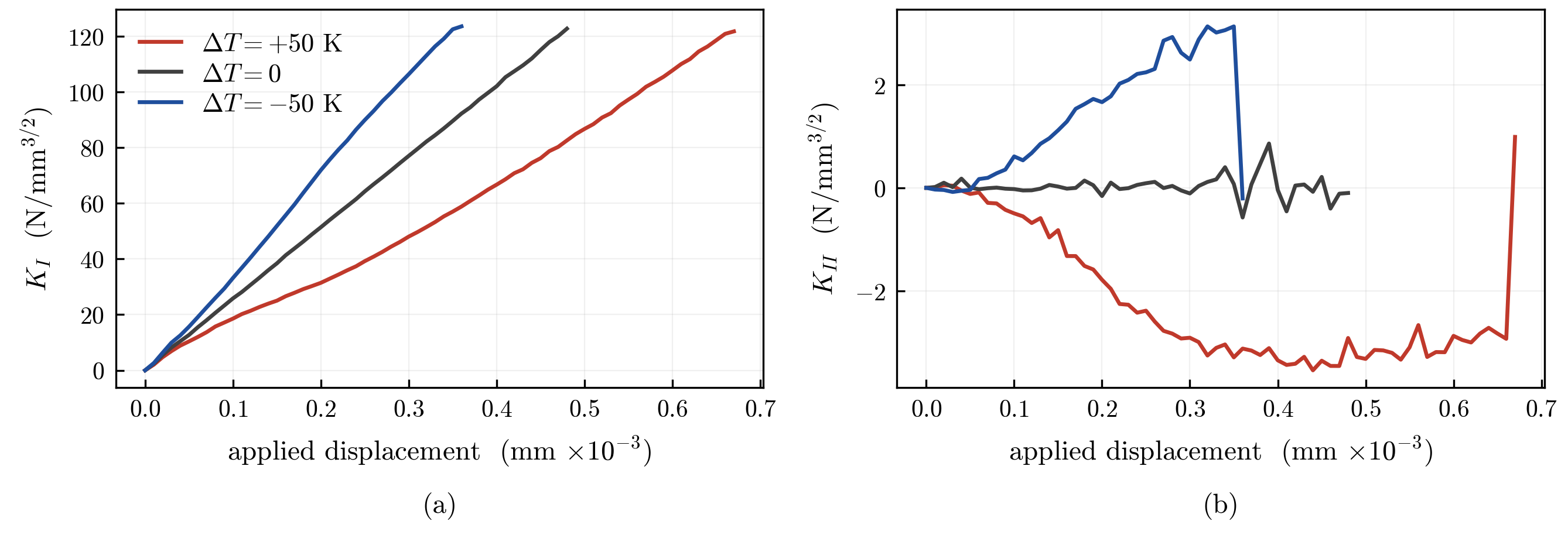}
  \caption{SIFs of the single-edge notched tension test against the applied displacement: (a) $K_I$, (b) $K_{II}$, with the legend of (a). The test is mode I to the accuracy of the extraction. Readings after the tip has reached the far edge, where no contour fits inside the material, are not plotted.}
  \label{fig:sent_sif}
\end{figure}

Fig.~\ref{fig:sent_field_evolution} shows the field evolution for the heating case with \(\Delta T=+50~\mathrm{K}\), from the same run as the curves of Figs.~\ref{fig:sent_fd} and \ref{fig:sent_sif}. The notch is unchanged until step 67, when the crack crosses the whole remaining ligament within that single step, so step 66 is the last step before growth, at the peak load, and step 70 is after separation. Since the ligament is crossed within one or two load steps, this test is run without a separation margin, and its failure displacement is that of the load step in which the crack completes the crossing. By step 66 the imposed temperature difference has fully developed and the isotherms are drawn down around the notch tip, since the adiabatic faces let the heat from the upper edge reach the region below the notch only around the tip. At step 70 the vertical displacement separates across the crack and the two halves of the specimen move independently.

The crack-tip quantities that drive this response are shown in Fig.~\ref{fig:sent_sif}. \(K_I\) rises with the applied displacement in all three cases. The isothermal curve is linear. The thermal curves are offset from it by the thermal stress at the notch, which keeps evolving after the ramp as the temperature diffuses through the transient, so their slopes differ from the isothermal one, the cooled specimen reaching a given SIF at the smallest displacement and the heated one at the largest, in the same order as the initiation points of Fig.~\ref{fig:sent_fd}. The extracted \(K_{II}\) stays below \(3.6~\mathrm{N/mm^{3/2}}\) throughout, against a \(K_I\) that passes \(120\), so the test is mode I to the accuracy of the extraction. The two thermal cases show a small mode II component of opposite signs, negative under heating and positive under cooling, the pattern the geometry suggests, since the imposed temperature difference is antisymmetric about the notch plane where the mechanical load is symmetric about it. The component is of the size of the extraction error. It is consistent with the slight deflection of the crack from the notch plane as it approaches the far edge, seen in the last column of Fig.~\ref{fig:sent_field_evolution} for the heated case, and Tangella et al.~\cite{tangella2022hybrid} likewise report crack paths deflected by the thermal load in this test. The crack advances when \(K_I\) reaches between \(120\) and \(123~\mathrm{N/mm^{3/2}}\) in all three cases, which is the pure mode I form \(K_I=\sqrt{E'G_c}\) of the criterion \eqref{eq:crack_propagation_condition}, since the kink angle of every accepted extension stays within about one degree of straight ahead. The critical SIF therefore follows from the imposed \(G_c\) and is a consistency check rather than an independent test. It does separate the two effects of the thermal load. The critical SIF moves by 1.5\% between the heated and the cooled case, through the temperature dependence of \(G_c\), while the peak-load displacement, the last before the crack advances, moves by 89\%.

The admissibility checks of Section~\ref{subsec:extraction_admissibility} refused no extraction at a load step where the crack could have advanced and released no reading, so they did not alter the response of this test.

\subsection{Single-edge notched shear test with functionally graded material}
\label{subsec:sens_fgm}

The same notched specimen as in Section~\ref{subsec:sent} is considered under shear loading, where the crack path is no longer expected to remain straight. The lower edge is fixed in both displacement components, a horizontal displacement is applied monotonically on the upper edge with its vertical component held at zero, and the two vertical edges are traction free. The resulting mixed-mode crack driving force provides a direct test of the predicted propagation direction.

A functionally graded material is introduced in this test, so that the pointwise graded properties of Section~\ref{subsec:constitutive} and the graded extraction of Section~\ref{subsec:crack_tip_quantities} are exercised. The test is the thermoelastic shear test of Tangella et al.~\cite{tangella2022hybrid}, which their paper reports for a homogeneous material and their released code~\cite{tangella2022code} implements in graded form, and the graded properties of Table~\ref{tab:sens_fgm_parameters} are taken from that code. No solution of the graded form is published, so both references for this test are computed here. The properties vary linearly in the horizontal direction, a generic material parameter \(m\) being interpolated as
\begin{equation}
  m(x)
  =
  m_1
  +
  \frac{x}{L}
  \left(
    m_2-m_1
  \right),
  \qquad
  0\leq x\leq L ,
  \label{eq:sens_fgm_interpolation}
\end{equation}
where \(L=1~\mathrm{mm}\) and \(m_1\) and \(m_2\) are the values at the left and right edges. In the extraction the auxiliary fields and \eqref{eq:sif_extraction} use the material properties at the current tip, and the stresses on the contour the pointwise graded properties. The terms of the interaction integral that involve the material gradient~\cite{kckim2008} are omitted, since across the contour of this test the graded properties vary by about $1\%$, which is the order of those terms.

\begin{table}[htbp]
  \centering
  \caption{Material parameters for the functionally graded single-edge notched shear test, adopted from the implementation of Tangella et al.~\cite{tangella2022code}. The values at $x=0$ are those of the homogeneous tension test in Table~\ref{tab:sent_parameters}.}
  \label{tab:sens_fgm_parameters}
  \begin{tabular}{llll}
    \toprule
    Parameter & Symbol & \(x=0\) & \(x=L\) \\
    \midrule
    Young's modulus & \(E\) & \(340~\mathrm{GPa}\) & \(450~\mathrm{GPa}\) \\
    Poisson's ratio & \(\nu\) & \(0.22\) & \(0.22\) \\
    Critical energy release rate & \(G_c\) & \(42.47~\mathrm{J/m^2}\) & \(120~\mathrm{J/m^2}\) \\
    Density & \(\varrho\) & \(2450~\mathrm{kg/m^3}\) & \(4000~\mathrm{kg/m^3}\) \\
    Thermal conductivity & \(k\) & \(300~\mathrm{W/(m\,K)}\) & \(350~\mathrm{W/(m\,K)}\) \\
    Specific heat capacity & \(C_p\) & \(0.775~\mathrm{J/(kg\,K)}\) & \(0.775~\mathrm{J/(kg\,K)}\) \\
    Thermal expansion coefficient & \(\alpha\) & \(8.0\times10^{-6}~\mathrm{K^{-1}}\) & \(8.0\times10^{-6}~\mathrm{K^{-1}}\) \\
    Reference temperature & \(T_0\) & \(300~\mathrm{K}\) & \(300~\mathrm{K}\) \\
    \bottomrule
  \end{tabular}
\end{table}

The shear test is solved for \(\Delta T=-30~\mathrm{K}\), \(0~\mathrm{K}\), and \(+30~\mathrm{K}\) over 141 load steps of $2.5\times10^{-5}$~mm, with the temperature difference applied in equal increments over the first 12 steps and held from there on, and with the same crack increment \(\Delta a=0.025~\mathrm{mm}\) as the tension test.

\begin{figure}[htbp]
  \centering
\begin{tikzpicture}
    \begin{axis}[
      width=0.74\textwidth,
      height=0.42\textwidth,
      xlabel={Applied horizontal displacement \((10^{-3}~\mathrm{mm})\)},
      ylabel={Force (N)},
      grid=both,
      xmin=-0.05,
      xmax=3.8,
      ymin=-5,
      ymax=140,
      xtick={0,0.5,1.0,1.5,2.0,2.5,3.0},
      tick label style={font=\small},
      label style={font=\small},
      legend style={
        font=\small,
        draw=none,
        fill=none,
        at={(0.5,-0.22)},
        anchor=north,
        /tikz/every even column/.append style={column sep=8pt}
      },
      legend columns=3,
      legend image post style={black},
      legend cell align=left,
      unbounded coords=discard
    ]

    \addplot[blue, dashed, thick]
      table[
        col sep=comma,
        x expr=\thisrowno{0}*1000,
        y index=1,
        skip first n=1
      ]
      {figures/data/sens_fgm_fem_m30.csv};
    \addlegendentry{phase field}

    \addplot[blue, solid, thick]
      table[
        col sep=comma,
        x expr=\thisrowno{0}*1000,
        y index=1,
        skip first n=1
      ]
      {figures/data/sens_fgm_xdem_m30.csv};
    \addlegendentry{proposed method}

    \addplot[black, dashed, thick, forget plot]
      table[
        col sep=comma,
        x expr=\thisrowno{0}*1000,
        y index=1,
        skip first n=1
      ]
      {figures/data/sens_fgm_fem_0.csv};

    \addplot[black, solid, thick, forget plot]
      table[
        col sep=comma,
        x expr=\thisrowno{0}*1000,
        y index=1,
        skip first n=1
      ]
      {figures/data/sens_fgm_xdem_0.csv};

    \addplot[red, dashed, thick, forget plot]
      table[
        col sep=comma,
        x expr=\thisrowno{0}*1000,
        y index=1,
        skip first n=1
      ]
      {figures/data/sens_fgm_fem_p30.csv};

    \addplot[red, solid, thick, forget plot]
      table[
        col sep=comma,
        x expr=\thisrowno{0}*1000,
        y index=1,
        skip first n=1
      ]
      {figures/data/sens_fgm_xdem_p30.csv};

    \addplot[only marks, mark=star, mark size=4.5pt, line width=0.9pt, blue]
      coordinates {(1.025, 89.03)};
    \addlegendentry{sharp crack, initiation}
    \addplot[only marks, mark=star, mark size=4.5pt, line width=0.9pt, black, forget plot]
      coordinates {(1.150, 102.32)};
    \addplot[only marks, mark=star, mark size=4.5pt, line width=0.9pt, red, forget plot]
      coordinates {(1.275, 115.83)};

    \end{axis}
  \end{tikzpicture}
  \caption{Force--displacement response of the graded single-edge notched shear test at \(\Delta T=-30\), \(0\) and \(+30~\mathrm{K}\), with the colors and line styles of Fig.~\ref{fig:sent_fd}. The stars mark initiation in the sharp-crack finite element solution, which holds the crack stationary.}
  \label{fig:sens_fd}
\end{figure}
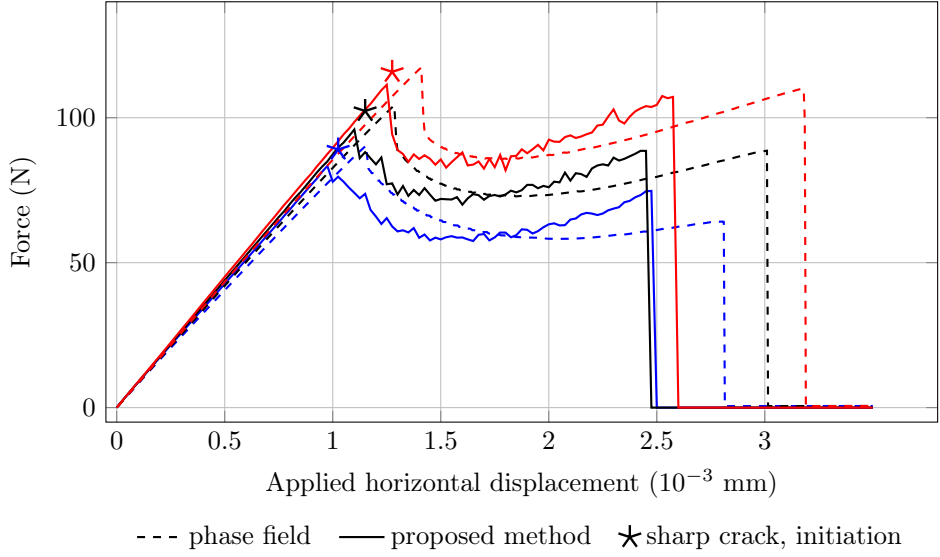

The phase-field reference for this test is the solver of Section~\ref{subsec:sent} with the graded properties of Table~\ref{tab:sens_fgm_parameters}. It represents the notch by setting the damage variable to one along it, so its notch is a damaged band of width about $\ell$ rather than a slit, and it differs from the proposed method before either crack has grown. To separate this effect from any difference between the solvers, a second reference was computed with a sharp crack. It is a standard finite element solution of the same problem, with the notch as a seam of the mesh and the crack held at its initial length, which suffices up to initiation. Its SIFs are obtained without the interaction integral, so that it shares no part of the extraction under test, and the criterion of Section~\ref{subsec:crack_geometry_update} then gives the load step at which the crack first grows, marked by stars in Fig.~\ref{fig:sens_fd}. The procedure and its check on the stationary benchmark of Section~\ref{subsec:wang_verification} are given in Appendix~\ref{app:references}.

The force--displacement responses are shown in Fig.~\ref{fig:sens_fd}. Against the sharp-crack solution the loading ramp of the proposed method agrees to within 1\% to 2\%, and initiation falls within one load step in all three thermal cases, 2\% of the applied displacement at that point. This lies inside the run-to-run scatter of $0$ to $2$ load steps in initiation and $0.2\%$ to $4.5\%$ in peak load reported in Appendix~\ref{app:discretization}. Since the sharp-crack solution applies the same criterion and the same temperature-dependent toughness, the agreement tests the trained fields and the extraction. All three solutions reproduce the influence of the thermal load, the peak load rising monotonically with \(\Delta T\), cooling advancing fracture and heating delaying it.

The phase-field reference reaches its peak five load steps after the sharp-crack solution initiates in every case, and five to six after the proposed method. Two properties of the regularized description delay it. Its AT2 crack density function departs from the undamaged stress--strain response before the critical stress is reached~\cite{kristensen2021}, so damage and compliance accumulate ahead of the peak while a sharp crack stays elastic. Its discretization raises the effective toughness to $G_c(1+h/2\ell)$~\cite{bourdin2000}, with $h$ its element size, which raises a Griffith peak load by the square root of that factor. Its fracture energy is also held constant, where the proposed method and the sharp-crack solution evaluate it at the tip temperature.

The sharp-crack solution also supplies the temperature field, and with it a test of the choice of Section~\ref{subsec:temperature_approximation_energy} to leave the temperature without tip enrichment. At load step 47 of the heated case, four steps before initiation, the adiabatic crack splits the field by $18.6$~K of the $30$~K imposed, the largest difference between the two face temperatures along the notch, against $18.5$~K in the reference. On a $501\times501$ grid, excluding the row of the notch itself, the trained temperature agrees with the finite element field to $0.08$~K root mean square. Farther than $0.05$~mm from the tip the largest pointwise deviation is $0.43$~K, which matches the discretization scatter of the reference itself, measured by repeating it on a mesh twice as fine. Closer to the tip the deviations are larger, and there the ring-sampled tip temperature of \eqref{eq:crack_tip_temperature} agrees with the reference to $0.24$~K.

\begin{figure}[htbp]
  \centering
  \includegraphics[width=\textwidth]{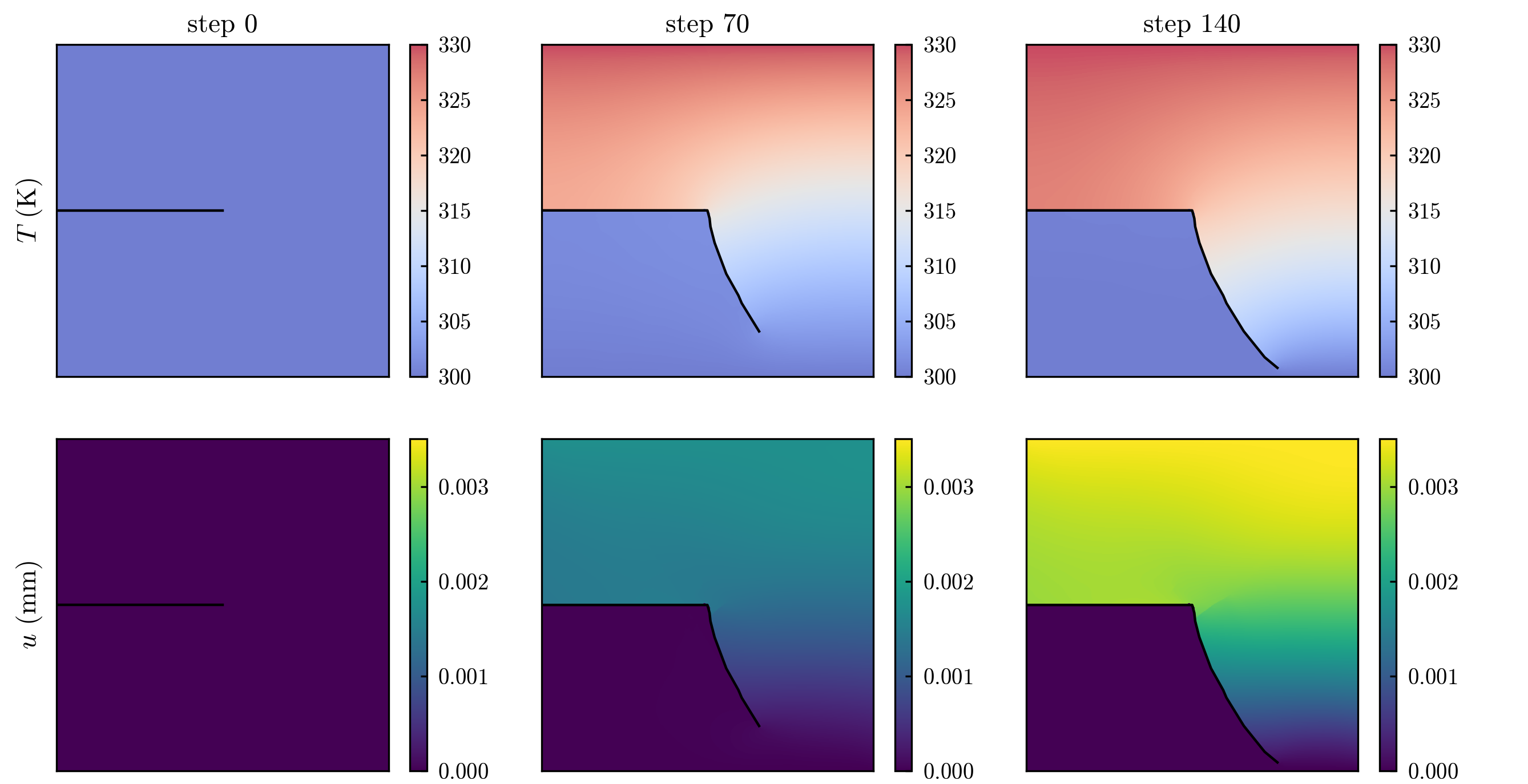}
  \caption{Temperature and horizontal displacement in the graded single-edge notched shear test at \(\Delta T=+30~\mathrm{K}\), at load steps 0, 70 and 140 of 141. Each row shares one color scale. The region below the crack stays near the initial temperature since the crack faces are adiabatic.}
  \label{fig:sens_fgm_crack_evolution}
\end{figure}

Fig.~\ref{fig:sens_fgm_crack_evolution} shows the fields of the heated case from the same run as Fig.~\ref{fig:sens_fd}. The crack leaves the notch at its tip and turns downward. The region below the crack is connected to the heated upper edge only around the tip, so it stays near the initial temperature and the isotherms wrap around the advancing tip, while the displacement jumps across the faces.

\begin{figure}[htbp]
  \centering
  \includegraphics[width=\textwidth]{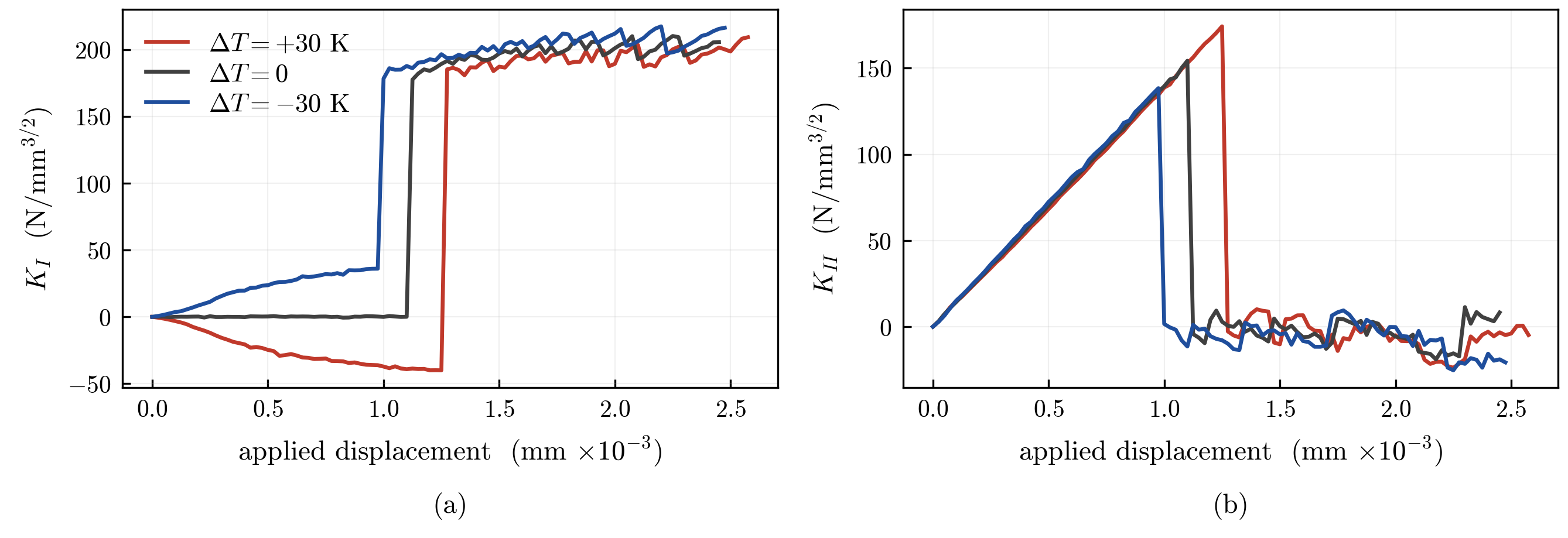}
  \caption{SIFs of the graded shear test against the applied displacement: (a) $K_I$, (b) $K_{II}$, with the legend of (a). $K_{II}$ grows while the notch is stationary and collapses when the first extension turns out of its plane, where $K_I$ takes over. Readings after the tip has reached the separation margin $\delta$ of Section~\ref{subsec:crack_geometry_update} are not plotted.}
  \label{fig:sens_sif}
\end{figure}

The downward kink is not prescribed. It follows from the extracted SIFs through the maximum hoop stress criterion of Section~\ref{subsec:crack_growth_direction}, and the resulting path is consistent with the phase-field reference. The SIFs are shown in Fig.~\ref{fig:sens_sif}. Before initiation the specimen is shear dominated. $K_{II}$ grows linearly with the applied displacement to between $140$ and $175~\mathrm{N/mm^{3/2}}$ at the last step before growth, while $K_I$ stays small and is offset by the thermal load, to about $-40~\mathrm{N/mm^{3/2}}$ in the heated case and about $+36$ in the cooled one, since heating expands the material across the notch and cooling contracts it. The first extension leaves the notch plane at the kink angle of this mixed-mode state and turns the crack onto the plane where the shear nearly vanishes. $K_{II}$ drops to between $-25$ and $+11~\mathrm{N/mm^{3/2}}$ and $K_I$ rises to between $177$ and $217$, so growth is mode I dominated from there. The criterion ratio $\chi$ then stays close to $\chi_c$ for the rest of the run, as a Griffith criterion should in a displacement-controlled test, and the thermal ordering of initiation is visible in both panels, cooling reaching the threshold first.

In the heated case $K_I$ is negative before initiation. With the traction-free faces of Section~\ref{subsec:mechanical_equilibrium} this means that the faces interpenetrate slightly near the tip. The embedding admits a jump of either sign, and only a contact condition, keeping the faces from crossing and transmitting a compressive traction between them, would exclude a negative normal jump. The reading that authorized the first extension was taken in that state, $K_I=-40$ and $K_{II}=178~\mathrm{N/mm^{3/2}}$ at load step $51$, with $\chi=1.03$ and a kink angle of $-75^{\circ}$, and the positive $K_I$ of Fig.~\ref{fig:sens_sif} is recorded after the crack has turned. Contact would relieve the compressive $K_I$, which in the computation is what delays the heated case relative to the isothermal one, so under contact that delay would shrink and the first kink would lie nearer the pure mode II angle. The sharp-crack reference rests on the same traction-free faces, so the comparison with it is made under this assumption.

The admissibility checks of Section~\ref{subsec:extraction_admissibility} refused the reading at three to five load steps per case, and two extensions of the heated case were taken on a clipped sweep.

The computed curves end $11\%$ to $19\%$ earlier than the phase-field reference separates, in fractions of its separation displacement, but the two ends are different events. The computation stops when the tip comes within the separation margin of Section~\ref{subsec:crack_geometry_update}, here $\delta=0.03~\mathrm{mm}$, the last admitted extraction being taken at about $0.032~\mathrm{mm}$ from the boundary with the contour at full size, whereas the reference continues to full separation. The crack covered the $0.03~\mathrm{mm}$ preceding the margin in $18$, $15$ and $19$ load steps, and crossing the margin at the same rate would take as many again, $15\%$ to $18\%$ of the displacement at the margin, so the margin alone could account for a difference of this size. Arrival at the margin is quantized by the crack increment, which is why the cooled and isothermal cases stop within one load step of each other.

\subsection{Notched cruciform test under combined thermo-mechanical loading}
\label{subsec:cruciform}

The notched cruciform specimen has served as a benchmark of thermoelastic crack growth since Prasad et al.~\cite{prasad1994} introduced it and computed it with the dual boundary element method. It has since been solved with enriched finite elements~\cite{duflot2008,nguyenbui2017}, meshfree and smoothed formulations~\cite{wang2015tafm,chen2016esfem}, a moving mesh technique~\cite{greco2021}, and phase-field models~\cite{mandal2021fracture,chen2024sbfem}. This history makes it the one propagating example whose predicted paths can be laid over a body of independent published solutions.

\begin{figure}[htbp]
  \centering
  \includegraphics[width=\textwidth]{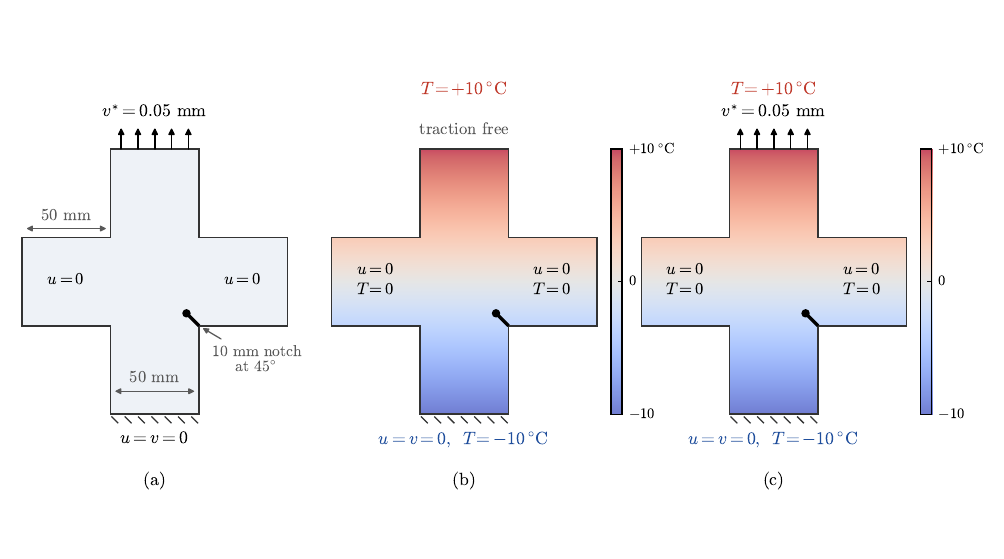}
  \caption{Geometry of the notched cruciform specimen and the three load cases. (a) Case I, the top edge displacement $v^{*}$ alone, isothermal. (b) Case II, the temperatures alone, with the top edge traction free. (c) Case III, both. In all three the side arms are held at the reference temperature with $u=0$ and the bottom edge is fixed. Dimensions are marked on (a). The temperature shading is schematic.}
  \label{fig:cruciform_geometry_bc}
\end{figure}

The specimen and its boundary data are shown in Fig.~\ref{fig:cruciform_geometry_bc}. It is a cruciform plate with a $10~\mathrm{mm}$ notch at $45^{\circ}$ from its lower right re-entrant corner, and three load cases are run, following the benchmark's standard program. Case I applies the mechanical load on the top edge alone, isothermally. Case II applies the temperatures alone, $+10~^{\circ}\mathrm{C}$ on the top arm and $-10~^{\circ}\mathrm{C}$ on the bottom, with the top edge left traction free, as the published treatments leave it. Case III applies both loads at once. In every case the side arms are held at the reference temperature with $u=0$, and the bottom edge is fixed.

The published solutions do not all solve the same problem. They share the geometry and the three load cases, and fall into two families that differ in how the mechanical load is applied and in the material data. The original statement of Prasad et al.~\cite{prasad1994} applies a traction of $10~\mathrm{Pa}$ on the top edge, with $\nu=0.3$ and $\alpha=1.67\times10^{-5}~^{\circ}\mathrm{C}^{-1}$, and grows the crack by a prescribed number of increments, so it involves no fracture resistance. The restatement of Mandal et al.~\cite{mandal2021fracture} prescribes a displacement of $0.05~\mathrm{mm}$ on the top edge instead, with $\nu=0.2$, $\alpha=6.0\times10^{-4}~^{\circ}\mathrm{C}^{-1}$ and a fracture energy, in plane stress. We compute the second statement, with its material data and its plane stress condition unchanged, since the propagation criterion of Section~\ref{subsec:crack_geometry_update} requires the fracture energy that only this statement supplies. The differences in Poisson's ratio, thermal expansion and plane condition among the solutions compared below are therefore differences between the two published statements, not choices made here.

The material parameters are listed in Table~\ref{tab:cruciform_parameters}, quoted exactly as Mandal et al.\ state them, in their consistent N--m--$^{\circ}$C system. The vanishing mass density makes the conduction problem of Section~\ref{subsec:thermal} steady at each load step, which matches their treatment of the thermal field. The tensile strength \(f_t\) enters only their phase-field model, and the proposed method uses the fracture energy \(G_f\) as a constant \(G_{c0}\), without the temperature dependence of \eqref{eq:temperature_dependent_gc}, which the published statement does not contain. Each case is run for 101 load steps over which the imposed displacement and the imposed temperatures rise linearly from zero to their final values, so that in case III their ratio is the same at every step. At most one crack increment of \(\Delta a=4~\mathrm{mm}\) is taken per step, nearly the same fraction of the specimen width as in the single-edge notched tests, so that the growth is paced by the load stepping in the same way as the phase-field solution it is compared with~\cite{mandal2021fracture}. A step whose ratio stays above the threshold after its extension keeps that surplus for the next step, with the consequence for the criterion ratio recorded in Appendix~\ref{app:discretization}. The test is run without a separation margin, and growth stops when the tip reaches the boundary of the specimen. The specimen occupies part of its square bounding box, and the corner regions outside the arms receive no integration points. Each case is run three times from different network initializations. The integration points are drawn independently in every run, so the three runs are independent repeats, and the spread quoted below is the spread over them.

\begin{table}[htbp]
  \centering
  \caption{Material parameters for the notched cruciform test.}
  \label{tab:cruciform_parameters}
  \begin{tabular}{lll}
    \toprule
    Parameter & Symbol & Value \\
    \midrule
    Young's modulus & \(E\) & \(218.4\times10^{3}~\mathrm{N/m^2}\) \\
    Poisson's ratio & \(\nu\) & \(0.2\) \\
    Fracture energy & \(G_f\) & \(2.0\times10^{-4}~\mathrm{N/m}\) \\
    Tensile strength & \(f_t\) & \(120~\mathrm{N/m^2}\) \\
    Thermal expansion coefficient & \(\alpha\) & \(6.0\times10^{-4}~^{\circ}\mathrm{C}^{-1}\) \\
    Density & \(\varrho\) & \(0.0~\mathrm{kg/m^3}\) \\
    Thermal conductivity & \(k\) & \(1.0~\mathrm{W/(m\,^{\circ}C)}\) \\
    Specific heat capacity & \(C_p\) & \(1.0~\mathrm{N\,m/(kg\,^{\circ}C)}\) \\
    Reference temperature & \(T_0\) & \(0~^{\circ}\mathrm{C}\) \\
    \bottomrule
  \end{tabular}
\end{table}

Table~\ref{tab:cruciform_references} lists the published solutions used below with the statement each solves. When only one load acts, as in cases I and II, its amplitude cancels out of the growth direction. The two statements then still differ in Poisson's ratio, in the plane condition and in the traction against the displacement on the top edge, but their published single-load paths agree with one another within the spreads quoted below, so all six solutions are drawn for these cases. When both loads act, as in case III, the ratio of the mechanical to the thermal load decides the path, the two families differ in thermal expansion by a factor of $36$ and in how the mechanical load enters, and only the two displacement-controlled solutions are drawn. Merging the families into one band for case III would produce a spread of nearly $40^{\circ}$ that no single problem exhibits.

\begin{table}[htbp]
  \centering
  \caption{Published solutions of the notched cruciform specimen and the statement of the problem that each solves. Every source that states Young's modulus gives $E=218.4\times10^{3}~\mathrm{N/m^2}$. The traction on the top edge is $10$~Pa and the displacement $0.05$~mm. An asterisk marks data that the source does not state, taken from Prasad et al.~\cite{prasad1994}, to whose statement of the problem the source refers, and a dash marks what neither states. The last column gives the load cases in which the path is compared, and F marks the two solutions whose SIFs are compared.}
  \label{tab:cruciform_references}
  \footnotesize
  \setlength{\tabcolsep}{4.5pt}
  \begin{tabular}{@{}llllllll@{}}
    \toprule
    Solution & Crack & Method & Load & $\nu$ & $\alpha$ ($^{\circ}\mathrm{C}^{-1}$) & Plane & Used in \\
    \midrule
    Wang~\cite{wang2015tafm} & sharp & meshfree & traction & $0.3^{\ast}$ & $1.67\times10^{-5\,\ast}$ & -- & I, II \\
    Nguyen et al.~\cite{nguyenbui2017} & sharp & enriched elements & traction & $0.3^{\ast}$ & $1.67\times10^{-5\,\ast}$ & -- & I, II \\
    Chen et al.~\cite{chen2016esfem} & sharp & smoothed elements & traction & $0.3$ & $1.67\times10^{-5}$ & strain & I, II, F \\
    Greco et al.~\cite{greco2021} & sharp & moving mesh & traction & $0.3$ & $1.67\times10^{-5}$ & strain & I, II, F \\
    Mandal et al.~\cite{mandal2021fracture} & phase field & finite elements & displacement & $0.2$ & $6.0\times10^{-4}$ & stress & I, II, III \\
    Chen et al.~\cite{chen2024sbfem} & phase field & scaled boundary elements & displacement & $0.2$ & $6.0\times10^{-4}$ & -- & I, II, III \\
    \midrule
    Proposed method & sharp & & displacement & $0.2$ & $6.0\times10^{-4}$ & stress & \\
    \bottomrule
  \end{tabular}
\end{table}

\begin{figure}[htbp]
  \centering
  \includegraphics[width=\textwidth]{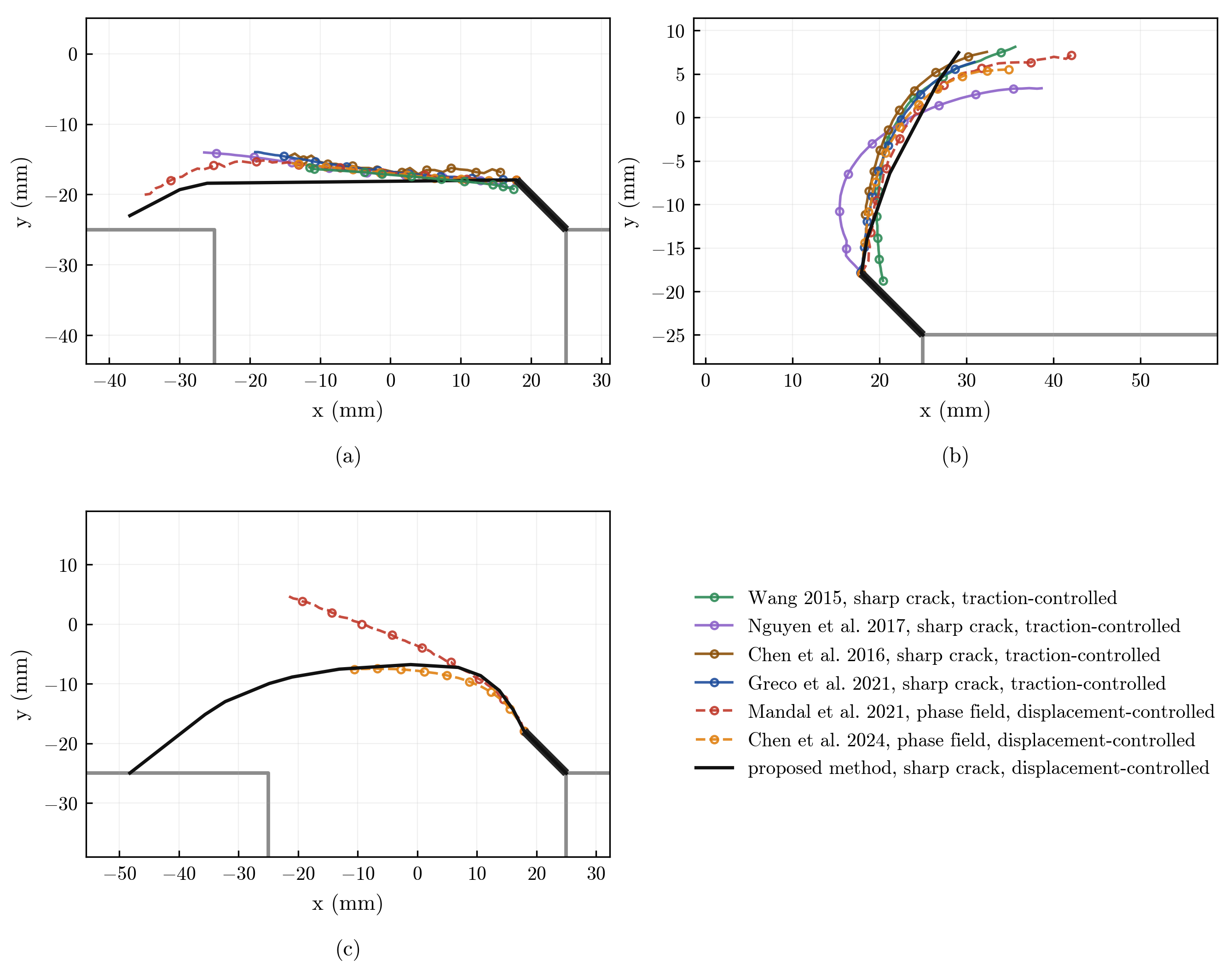}
  \caption{Crack paths of the three load cases against the published solutions, labeled by crack description. (a) Case I, mechanical. (b) Case II, thermal. (c) Case III, both, with the two displacement-controlled references only, since the traction-controlled family solves a different combined problem. One of the three repeats is drawn per case, the other two staying within $1.6~\mathrm{mm}$ of it.}
  \label{fig:cruciform_paths}
\end{figure}

Fig.~\ref{fig:cruciform_paths} lays the computed paths over the published ones. The three repeats of a case stay within $1.6~\mathrm{mm}$ of one another, under half the crack increment, so one is drawn. To compare the paths by one number, each is summarized by its chord angle, the angle from the global $x$ axis of the line from the notch tip to the point where the path crosses a circle of radius $12$~mm about it, three crack increments from the tip and a distance that every published path reaches. The angle measures the initial direction of the path and not its whole course. Table~\ref{tab:chord_angles} collects the angles. In case I the three repeats fall $1.5^{\circ}$ to $1.7^{\circ}$ above the published span and in case II $2.4^{\circ}$ to $4.1^{\circ}$ below it, the spans themselves being $5.7^{\circ}$ and $18.9^{\circ}$ wide, and in case III they fall $1.3^{\circ}$ to $3.2^{\circ}$ below the closer of the two displacement-controlled references, which are $5.4^{\circ}$ apart. Beyond the chord radius the figure shows how far the agreement extends along each path.

\begin{table}[htbp]
  \centering
  \caption{Chord angles of the crack paths at $12$~mm from the notch tip, measured from the global $x$ axis. The proposed method is given as the first of three repeats with the range over the three, the published solutions as the span of those drawn in Fig.~\ref{fig:cruciform_paths}, and the last column gives the distance of the repeats from the nearest published value.}
  \label{tab:chord_angles}
  \begin{tabular}{lccc}
    \toprule
    Case & Proposed method & Published & Outside the published span \\
    \midrule
    I, mechanical & $180.6^{\circ}$ ($180.5^{\circ}$ to $180.7^{\circ}$) & $173.3^{\circ}$ to $179.0^{\circ}$ & $1.5^{\circ}$ to $1.7^{\circ}$ above \\
    II, thermal & $74.1^{\circ}$ ($72.9^{\circ}$ to $74.6^{\circ}$) & $77.0^{\circ}$ to $95.9^{\circ}$ & $2.4^{\circ}$ to $4.1^{\circ}$ below \\
    III, both & $128.7^{\circ}$ ($128.7^{\circ}$ to $130.5^{\circ}$) & $131.9^{\circ}$ and $137.3^{\circ}$ & $1.3^{\circ}$ to $3.2^{\circ}$ below \\
    \bottomrule
  \end{tabular}
\end{table}

\begin{figure}[htbp]
  \centering
  \includegraphics[width=\textwidth]{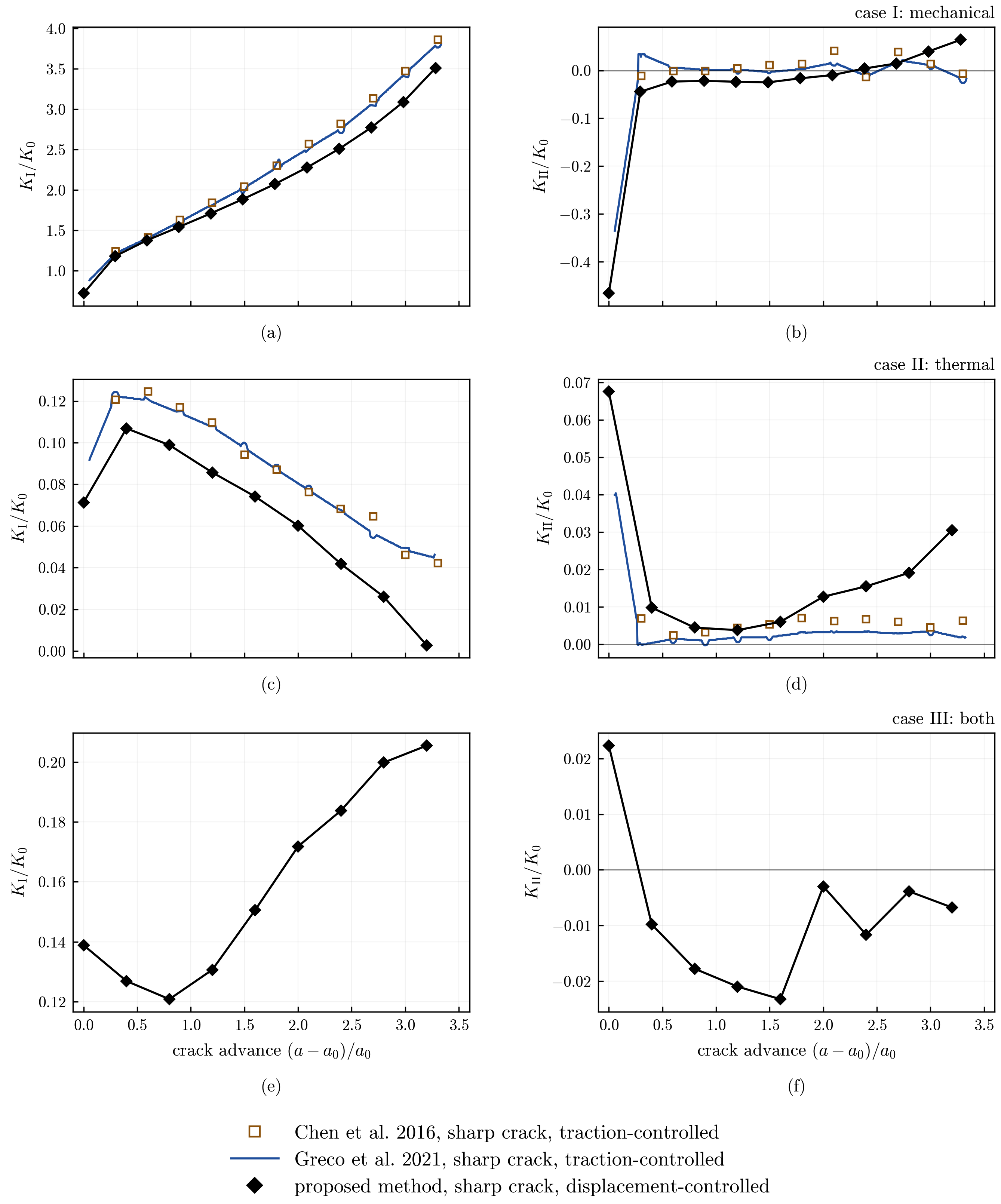}
  \caption{Both SIFs against the crack advance $(a-a_0)/a_0$, read along the path the proposed method grows for itself, as the published curves are read along theirs. Case I is read with the crack held at each position of that path under the final load, cases II and III from the propagating runs. Markers give the median over the three repeats, which lie within $3\%$ of each other in case I, $11\%$ in case II and $9\%$ in case III, except at $(a-a_0)/a_0=1.6$ in (c), where one repeat passed the radius sweep at a single load step, and at $2.8$ and $3.2$ in case II, where two repeats and one passed the radius sweep. Case III has no reference, since the only published SIFs under both loads are for a cruciform with cooled side arms.}
  \label{fig:cruciform_K}
\end{figure}

The SIFs are compared as well, in an exploratory way. Published SIFs exist only for the traction-controlled statement, each tabulated along the path of its own solution, and those of the proposed method are likewise read along its own path. In case I the normalization below needs the reaction force at each crack length, which the propagating runs do not record, so the crack is held at each position of the path it grew, the final load of the case is applied, the fields are trained and the SIFs are extracted. In cases II and III the SIFs are those of the propagating runs, at the load steps whose radius sweep passed, with the median taken over the steps spent at one crack length. Each family is normalized by its own load measure $K_0$, which is $\sigma_{\mathrm{eff}}\sqrt{\pi a_0}$ in the mechanical case, where $\sigma_{\mathrm{eff}}$ is the reaction force on the loaded arm divided by the arm width at that crack length, and $(1-\nu_r)E\alpha\,\Delta T\sqrt{\pi a_0}$ in the thermal case and in case III, with $\Delta T$ the amplitude at that load step, $a_0$ the notch length and the crack advance $a-a_0$ counted along the path. The factor $1-\nu_r$, with $\nu_r=0.3$ the Poisson's ratio of the traction-controlled statement, brings the thermal SIFs of the plane stress computation to the plane strain scale of the published curves. The families still differ in Poisson's ratio, in plane condition, in how the load enters and in path, so part of any residual discrepancy belongs to the difference in problem statement rather than to the method, and the path comparison above remains the primary validation.

Fig.~\ref{fig:cruciform_K} collects both modes for all three cases, with the curves of Chen et al.~\cite{chen2016esfem} and Greco et al.~\cite{greco2021} overlaid where comparable ones exist. In case I the extracted $K_I$ lies below the published curves over most of the growth history. The ratio at matched crack length averages $0.92$ against Chen et al. and $0.95$ against Greco et al., $0.93$ over both, and ranges from $0.89$ to $1.04$ along the history, with the average moving by $0.26\%$ between repeats. Mode II stays small, as it does for a crack that follows the maximum hoop stress direction, except at the notch before the first kink. Away from the notch $K_{II}/K_I$ lies between $-0.038$ and $+0.018$ with a median magnitude of $0.013$, against published values between $-0.009$ and $+0.030$.

In case II the extracted $K_I$ rises and declines with the published curves but lies below them, by $0.015$ in $K_I/K_0$ at $(a-a_0)/a_0=0.4$, about $0.02$ up to $2.0$ and about $0.03$ at $2.8$, so that it declines more steeply and its ratio to the published curves falls from $0.88$ to $0.48$ over that range. At $2.8$ the readings come from two repeats and at $3.2$ from one. $K_{II}/K_0$ stays between $0.004$ and $0.019$ against published values below $0.007$. The residual grows with crack length, and since the two families differ in Poisson's ratio, in plane condition and in path at once, the comparison does not attribute it to one of them. For case III no published SIFs exist for this combination of loads. Its three repeats agree in $K_I$ to within $9\%$.

\FloatBarrier

\section{Conclusions}
\label{sec:conclusions}

We have presented an extended deep energy method for thermo-mechanical crack propagation in which the crack is represented explicitly, as a polyline, and advanced by a criterion of classical fracture mechanics. The transient heat conduction equation is recast as an incremental functional and minimized in a staggered manner with the thermoelastic potential energy. Each field is represented by its own network, and the two networks share one embedding of the crack polyline, which admits the displacement jump and the temperature jump across the crack faces, the adiabatic condition on the faces following from the thermal functional itself. The displacement network is enriched with the leading asymptotic terms at the tip. The integration points are redrawn during training and densified near the tip. At each load step the SIFs are extracted from the trained fields by the interaction integral, the fracture energy is evaluated at the tip temperature, and the crack is extended when the criterion is met.

Under thermal load the interaction integral has to include the area term of Wilson and Yu in addition to the contour term. Without it, the SIF read from the same trained field changes with the contour radius, and with it, the reading on the stationary benchmark is path independent to between 3\% and 7\% across a fourfold span of radii. Every extraction is also taken as a sweep over contour radii, and a reading that drifts across the sweep by more than a declared tolerance authorizes no extension, so that the crack does not advance on SIFs the fields do not support.

The method was assessed on four benchmark problems. On the stationary edge crack under thermal load the extracted SIF agrees with the published value to $0.11\%$ on the finest of three grids. On the single-edge notched tension test the peak load agrees with the phase-field reference to within $4.1\%$ in the three thermal cases, with no systematic sign, and the displacement at peak load increases by 89\% from the cooled to the heated case. On the graded shear test the crack paths agree with the phase-field reference, and initiation agrees to within one load step with an independent sharp-crack finite element solution in all three thermal cases. On the notched cruciform specimen the predicted chord angles fall $1.3^{\circ}$ to $4.1^{\circ}$ outside the nearest published curve, where the published solutions themselves spread over $5.4^{\circ}$ to $18.9^{\circ}$ per case. The SIFs of the cruciform were compared in an exploratory way with published curves that solve the traction-controlled statement of the benchmark rather than the displacement-controlled one computed here. Read along the paths the method grows, the mechanical case gives a ratio averaging $0.93$ over the growth history, and in the thermal case $K_I$ follows the published decline below it by an offset that grows from $0.015$ to $0.03$ in units of the load measure. Run-to-run scatter was measured on every example and is reported beside each result.

The scope of the study is limited in several respects. All examples are two-dimensional and quasi-static. A single crack is followed, and neither branching nor merging is treated. The coupling is one way for a fixed crack, so no mechanical dissipation returns to the thermal problem, and the mechanical solution enters the conduction problem only through the crack it advances. The crack faces are taken open and traction free, so a loading state that closes the crack would require a contact condition. In the heated shear case the first extension was taken from such a state, and contact would shorten the delay of initiation that the computation shows. The crack advances by a prescribed increment along a direction evaluated once per extension, so a path is resolved to that increment, and the last increment before separation is not resolved, which in the tension test means the whole post-peak branch. The extraction contour is tied to the increment and has to span about two background elements, which sets a lower limit on the increment for a given grid. The tolerance of the radius sweep is a chosen value rather than a derived bound. For the cruciform specimen the published solutions solve two statements of the problem that share the geometry but not the loads, and no published SIFs exist for the combined load case, so that case is compared on its path only. A propagating case takes from $3.5$ to $12$ hours on one GPU, as reported in Appendix~\ref{app:discretization}, and no propagating reference solution was timed on the same task.

Extending the method to multiple and merging cracks, to crack nucleation, to three-dimensional problems, and to a two-way coupling in which the mechanical field feeds back into the thermal one is left to further work.

\section*{Data availability}
Data will be made available on request.

\section*{Code availability}
The repository \url{https://github.com/HannnZH/Extended-DEM-ThermoMechanical-Crack} contains the solver, the phase-field and sharp-crack finite element reference solvers and the scripts that reproduce the numerical examples.

\section*{Acknowledgements}
This research was undertaken with the assistance of resources from the Gadi supercomputer~\cite{nci2019gadi} at the National Computational Infrastructure (NCI Australia), an NCRIS enabled capability supported by the Australian Government, and from the Katana cluster~\cite{unsw2010katana} at UNSW Sydney, supported by Research Technology Services.

\appendix
\counterwithin{table}{section}
\counterwithin{figure}{section}
\section{Numerical parameters, cost and sensitivity}
\label{app:discretization}

The numerical parameters of the four examples are collected in Table~\ref{tab:discretization}. Each solve runs at most its stated number of Adam iterations and stops earlier when the mean loss over the last $500$ iterations has fallen by less than $10^{-4}$ of the mean over the $500$ before it on three consecutive checks, made every $250$ iterations, so that each mean spans one resampling interval. The partial retrainings after an extension, $1000$ thermal and $3000$ mechanical iterations in the notched tests and $1500$ and $3000$ in the cruciform, run to their count. The network weights are initialized randomly from the seed listed in Table~\ref{tab:discretization}. The seed does not govern the draw of the integration points, so the repeats of the notched tests, run with the same seed, differ only in that draw, whereas those of the cruciform also differ in the seed.

\begin{table}[htbp]
    \centering
    \caption{Numerical parameters of the four examples. Layer widths run from the input $(x,y,\rho)$ to the output. The thermal network has a sinusoidal first layer with $\omega_0=30$, the displacement network hyperbolic tangents throughout. All four examples share nine integration points per background element, the thermal network $3$--$50$--$50$--$50$--$1$ and the learning rate $10^{-3}$. The stationary example has no crack increment and no redraw. Lengths are in millimeters, the notched tests being computed on the unit square of side $1$~mm and the cruciform in its physical dimensions, and $\ell_\rho$ of \eqref{eq:implemented_localisation_weight} is given in millimeters or in units of the background element size $h$. The cruciform entries marked $\dagger$ are those of cases I and III, and case II uses the displacement network $3$--$100$--$100$--$100$--$100$--$2$ with $\alpha_{\rho}=20$.}
    \label{tab:discretization}
    \small
    \begin{tabular}{@{}>{\raggedright\arraybackslash}p{4.5cm}>{\raggedright\arraybackslash}p{2.6cm}>{\raggedright\arraybackslash}p{2.6cm}>{\raggedright\arraybackslash}p{2.6cm}>{\raggedright\arraybackslash}p{2.8cm}@{}}
        \toprule
        & Stationary crack & Tension & Graded shear & Cruciform \\
        & Section~\ref{subsec:wang_verification} & Section~\ref{subsec:sent} & Section~\ref{subsec:sens_fgm} & Section~\ref{subsec:cruciform} \\
        \midrule
        Background elements across the width & 48, 72, 96 & 100 & 100 & 150 \\
        Resampling interval & none & 500 iterations & 500 iterations & 500 iterations \\
        Displacement network & 3--50--50--50--2 & 3--50--50--50--2 & 3--50--50--50--2 & 3--50--50--50--2$^{\dagger}$ \\
        Adam iterations per load step, thermal / mech. & 8000 & 3000 / 3000 & 3000 / 3000, 5000 / 5000 at step 1 & 2000 / 3000 \\
        Unloaded step 0, Adam then L-BFGS, thermal / mech. & -- & 5000 / 7000, then 2500 / 3000 & 3000 / 3000, then 3000 / 1500 & 5000 / 7000, then 2500 / 3000 \\
        L-BFGS iterations after Adam, loaded steps & 2000 & -- & -- & -- \\
        Time increment $\Delta t$ & $1~\mathrm{s}$, steady & $10^{-8}~\mathrm{s}$ & $2.5\times10^{-8}~\mathrm{s}$ & $1~\mathrm{s}$, steady \\
        Crack increment $\Delta a$ & none & $0.025~\mathrm{mm}$ & $0.025~\mathrm{mm}$ & $4~\mathrm{mm}$ \\
        Extraction contour $r$ & $3.2h$ ($0.033$, $0.022$, $0.017$~mm) & $0.9\,\Delta a$ & $0.9\,\Delta a$ & $0.9\,\Delta a$ \\
        Kink angle cap & -- & $60^{\circ}$ & $80^{\circ}$ & $60^{\circ}$ \\
        Tip densification radius, factor & -- & $0.08~\mathrm{mm}$, $8$ & $0.08~\mathrm{mm}$, $8$ & $3~\mathrm{mm}$, $48$ \\
        Tip temperature ring $r_T$ ($N_T=24$) & -- & $2\times10^{-3}~\mathrm{mm}$ & $2\times10^{-3}~\mathrm{mm}$ & $0.25\,\Delta a$ \\
        Embedding decay $\alpha_{\rho}$, $\ell_{\rho}$ ($p_{\rho}=1$) & $1$, $5h$ & $20$, $1$~mm & $50$, $1$~mm & $1.2^{\dagger}$, $6$~mm \\
        Endpoint overlap $\delta_c$ (mm) & $0$ & $0.0125$ & $0.0125$ & $2$ \\
        Extraction face offset (mm) & $10^{-4}$ & $5\times10^{-4}$ & $5\times10^{-4}$ & $3\times10^{-3}$ \\
        Flux face offset (mm) & $10^{-4}$ & $10^{-4}$ & $10^{-4}$ & $5\times10^{-2}$ \\
        Flux weight $\lambda_q$, samples $N_q$ & $10^{-2}$, $200$ & $10^{-6}$, $150$ & $10^{-6}$, $150$ & $10^{-8}$, $200$ \\
        Enrichment decay $m_u$ (mm$^{-1}$, $q_u=1$) & $5$ & $5$ & $5$ & $0.05$ \\
        Extensions per load step, at most & -- & 35 & 50 & 1 \\
        Ratio cap of the convergence check & -- & 5 & -- & 100 \\
        Seeds & 2025 & 2025 & 2025 & 2025, 1, 2 \\
        \bottomrule
    \end{tabular}
\end{table}

The stationary crack and the cruciform have zero mass density, so their conduction problem is steady and their time increment is written as unity. The SIFs of the cruciform, $150~\mathrm{mm}$ across, are read on a contour of $3.6~\mathrm{mm}$. Cases I and III use the displacement network of the notched tests with a decay length $\ell_{\rho}/\alpha_{\rho}=5h$ of the embedding, and case II a larger network, $30\,902$ parameters against $5402$, with $\ell_{\rho}/\alpha_{\rho}=0.3h$.

The stationary benchmark is the one example trained under load on a single draw of its points, $82\,944$ to $331\,776$ on its three grids, with $2000$ L-BFGS iterations after Adam, and the agreement of its SIF with the published value is the check that no spurious layer of the kind described in Section~\ref{subsec:transient_thermal_energy_loss} formed there.

The flux weight $\lambda_q$ multiplies the mean squared normal flux $(k\,\partial T/\partial n)^2$ over $N_q$ points on the crack faces in physical units, so its natural scale is set by $k$ and by the imposed temperature difference over the size of the specimen, which is why it differs by orders of magnitude between the examples. It was set in each example so that the term remains a regularizer behind the natural condition of the thermal functional. At convergence its share of the thermal loss is $1.5\times10^{-7}$ of the conduction energy on the stationary crack and $2.5\times10^{-6}$ on the heated shear test at load step $47$, where the root mean square normal flux on the faces is $6\%$ of the flux scale in the body.

The implemented embedding of Section~\ref{subsec:crack_embedding} departs from the conditions \eqref{eq:crack_function_condition_jump} and \eqref{eq:crack_function_condition_smooth} in three local ways. First, the embedded jump closes only $\delta_c$ beyond the tip, so \eqref{eq:crack_function_condition_smooth} does not hold in a strip of width $\delta_c$ ahead of the tip, whose effect on the temperature is measured in Appendix~\ref{app:references}. For $\delta_c=0$, the setting of the stationary benchmark, the jump closes at the tip and the enrichment alone supplies the opening. Second, near a corner the sign change of $\tilde{d}$ departs from the polyline, by up to $0.004$~mm within one increment either side of the $74^{\circ}$ corner at which the crack of Fig.~\ref{fig:crack_embedding} leaves the notch and by up to $0.0005$~mm at corners of under $10^{\circ}$. Within those stretches two face samples at the offsets of Table~\ref{tab:discretization} can fall on one side, and the flux term then acts on one face only, which is why Fig.~\ref{fig:crack_embedding}(c) samples the jump at $0.006$~mm. Third, the projected arc length has a discontinuous gradient on the lines normal to a segment through its endpoints, where the projection \eqref{eq:segment_projection} and the end terms of \eqref{eq:arc_length_extension} change slope. The networks inherit these gradient discontinuities on curves of zero area, which the redrawn integration points meet with probability zero and the extraction contour crosses at isolated points.

Away from the corners the face offsets attenuate the sampled jump through the normal decay of the embedding, by the factor $\exp[-\alpha_{\rho}(\epsilon_c/\ell_{\rho})^{p_{\rho}}]$, which with the values of Table~\ref{tab:discretization} reduces the sampled jump by at most $2.5\%$ in the notched tests and $15\%$ for the flux offset of the cruciform.

The stationary benchmark trains and extracts in about $17$ minutes on a laptop GPU at its finest grid. The shear cases take $7.3$ hours isothermal and $11.5$ to $11.9$ hours under thermal load for their $141$ load steps on one V100, the tension cases $8.4$ to $9.0$ hours for their $106$, and a cruciform case $3.5$ to $4$ hours for its $101$ on one H200. Both networks are retrained at every load step and after every extension, so that a shear case comprises about $160$ staggered solves of $2.5$ to $4.5$ minutes each. In the sensitivity study below, $16$ and $4$ integration points per element instead of $9$ changed the run time by $+9\%$ and $-2\%$, and network widths between $30$ and $70$ by $2.4\%$, against $2.1\%$ between repeats of the same configuration, so the cost is set by the number of optimizer iterations rather than by the size of the representation. No propagating reference solution was timed on the same task. The sharp-crack reference of Appendix~\ref{app:references} covers its three thermal cases to initiation in $90$ seconds on one laptop CPU core but holds the crack stationary, and the cost of the phase-field references was not measured.

The three thermal cases of the graded shear test were each run twice, and Table~\ref{tab:scatter} gives the load step at which the crack first advances. The pairs initiate $2$, $1$ and $0$ load steps apart at $\Delta T=0$, $-30$ and $+30~\mathrm{K}$, with peak loads $4.5\%$, $2.4\%$ and $0.2\%$ apart. Every run lies within one load step of the sharp-crack solution.

\begin{table}[htbp]
    \centering
    \caption{Load step at which the crack first advances in the graded shear test, two runs per case, against the two references of Section~\ref{subsec:sens_fgm}. One load step is $2.5\times10^{-5}~\mathrm{mm}$ of applied displacement. The phase-field entry is the step of its peak load, since it has no discrete initiation. The two runs differ only in the draw of the integration points.}
    \label{tab:scatter}
    \small
    \begin{tabular}{@{}lccc@{}}
        \toprule
        & $\Delta T=-30~\mathrm{K}$ & $\Delta T=0$ & $\Delta T=+30~\mathrm{K}$ \\
        \midrule
        Proposed method & 40 & 45 & 51 \\
        Proposed method, repeat & 41 & 47 & 51 \\
        Sharp-crack finite elements & 41 & 46 & 51 \\
        Phase field & 46 & 51 & 56 \\
        \bottomrule
    \end{tabular}
\end{table}

The isothermal shear test was rerun with one numerical parameter changed at a time, and Table~\ref{tab:sens_sensitivity} gives the outcomes against the reported run. The repeat of that run sets the scale, $4.5\%$ on the peak and two load steps at initiation. Every variant that grows a crack peaks $2.0\%$ to $4.1\%$ above the baseline, within the $4.5\%$ of the repeat. The variants initiate at step $46$ or $47$ against $45$, and with the increment doubled the crack reaches the separation margin after $11$ or $12$ extensions against $23$, over nearly the same crack length. Halving the increment with the grid unchanged puts the contour at $1.1$ background elements, and the radius sweep then refuses every reading. The contour has to span about two background elements, so refining the increment requires refining the grid with it.

\begin{table}[htbp]
  \centering
  \caption{Sensitivity of the isothermal shear test to its numerical parameters, one change per row. Peak forces are read inside the growth window and given relative to the baseline. The repeat differs only in the draw of the integration points and sets the run-to-run scatter against which the other rows are read.}
  \label{tab:sens_sensitivity}
  \begin{tabular}{lccc}
    \toprule
    Variation & Peak & First growth & Extensions \\
    \midrule
    Baseline & $95.9$~N & step $45$ & $23$ \\
    Repeat, integration point draw resampled & $+4.5\%$ & $47$ & $23$ \\
    Increment halved to $0.0125$~mm, contour at $1.1$ elements & no growth & -- & $0$ \\
    Increment doubled to $0.05$~mm, radii $0.9\,\Delta a$ & $+2.0\%$ & $46$ & $11$ \\
    Increment doubled, contour pinned at $0.0225$~mm & $+2.8\%$ & $46$ & $12$ \\
    Both networks narrowed to width $30$ & $+2.9\%$ & $46$ & $23$ \\
    Both networks widened to $70$ & $+4.1\%$ & $47$ & $23$ \\
    Quadrature coarsened to $4$ points per element & $+3.1\%$ & $47$ & $23$ \\
    Quadrature refined to $16$ points per element & $+2.8\%$ & $46$ & $23$ \\
    \bottomrule
  \end{tabular}
\end{table}

Table~\ref{tab:cruciform_checks} gives the readings of the nine cruciform runs refused by the two checks of Section~\ref{subsec:extraction_admissibility} and those released on a clipped sweep. A sweep refusal follows either a drift beyond the threshold or a load step at which no contour fitted between the tip and the boundary. The convergence check has a cap of $100$ in the cruciform rather than $5$, since one extension per load step leaves the ratio above the threshold between steps by construction, and it refused only readings taken after the growth window, when the tip had reached the boundary region. A repeat of case III with that check removed grows over the same window of steps $29$ to $49$, with $18$ extensions against $18$ to $19$ and a chord angle $0.7^{\circ}$ from the reported run, inside the $1.9^{\circ}$ scatter over repeats, since the readings the check had refused are refused by the sweep instead.

\begin{table}[htbp]
  \centering
  \caption{Readings refused and released in the nine cruciform runs over $101$ load steps each, the reported run of each case first. The sweep refusals are split into a drift beyond the threshold and no fitting contour. The last column gives the readings released on a clipped sweep and, after the comma, the extensions taken on them, and an asterisk marks an entry that includes one extension taken on a single radius near the boundary.}
  \label{tab:cruciform_checks}
  \begin{tabular}{llcccccc}
    \toprule
    Case & Run & Extensions & \multicolumn{2}{c}{Sweep refusals} & Convergence & Released, extended \\
    & & & drift & no contour & refusals & \\
    \midrule
    I & reported & 14 & 6 & 37 & 0 & 0, 0 \\
    I & repeat 1 & 14 & 16 & 37 & 0 & 0, 0 \\
    I & repeat 2 & 15 & 7 & 36 & 0 & 1, 1$^{\ast}$ \\
    II & reported & 7 & 37 & 0 & 0 & 2, 0 \\
    II & repeat 1 & 7 & 33 & 0 & 0 & 3, 0 \\
    II & repeat 2 & 9 & 29 & 0 & 0 & 2, 0 \\
    III & reported & 19 & 3 & 0 & 51 & 2, 2$^{\ast}$ \\
    III & repeat 1 & 18 & 4 & 2 & 50 & 0, 0 \\
    III & repeat 2 & 18 & 2 & 0 & 52 & 1, 1 \\
    \bottomrule
  \end{tabular}
\end{table}

Two further checks on the stationary benchmark of Section~\ref{subsec:wang_verification} test the construction. With the enrichment amplitude started from five values between $-1.22$ and $+2.43$, one of the wrong sign, the extracted SIF lies between $0.4921$ and $0.4979$, a spread of $1.2\%$, while the trained amplitudes finish $7.8\%$ apart, so the SIF is set by the minimization and not by the initialization. Recomputing the three grids on a different GPU model moves no value by more than $0.7\%$.

\section{Reference solutions}
\label{app:references}

The phase-field reference of Sections~\ref{subsec:sent} and \ref{subsec:sens_fgm} is the staggered finite element solver that served as the reference in Zhang et al.~\cite{zhang2026meshfree}, extended to the thermoelastic case, on a uniform mesh of $512\times512$ bilinear quadrilateral elements of size about $\ell/5$, finer than the $\ell/2$ of common practice~\cite{miehe2010}, with $789\,507$ degrees of freedom for the displacement and the phase field together. It uses the AT2 crack density function in its second-order form with the hybrid split of Ambati et al.~\cite{ambati2015}, in which the full energy is degraded in the equilibrium equation and the tensile part of the spectral decomposition drives the phase field. Irreversibility is enforced through the history field of Miehe et al.~\cite{miehe2010cmame}, and the pre-existing crack is imposed by setting the phase field $\phi$ to one on the nodes within $1.5$ element sizes of the notch line. Within each load increment the temperature is solved first, by one backward Euler step of the transient conduction equation on the real clock of the loading, with the conductivity degraded by $(1-\phi)^2$ so that the crack faces are adiabatic and with the phase field of the previous increment, and is then held while the displacement and the phase field are alternated to a tolerance on the change of the phase field. The thermal strain uses the plane strain coefficient $(1+\nu)\alpha$, as the proposed method does. The fracture energy is constant in temperature and, in the graded test, linear in $x$ as Table~\ref{tab:sens_fgm_parameters} states it.

The solver was checked against the published curves of Tangella et al.~\cite{tangella2022hybrid} for the tension test before use. Its peak loads are $13\%$, $8\%$ and $9\%$ lower at $\Delta T=-50$, $0$ and $+50~\mathrm{K}$ and its displacements at peak $16\%$, $4\%$ and $3\%$ earlier, read from their published figure. The published solution regularizes over $\ell=0.005$~mm at $h=\ell/2$, against $\ell=0.01$~mm at $\ell/5$ here, and the toughness inflation $G_c(1+h/2\ell)$ of that coarser ratio accounts for $6.6\%$ of the peak difference on its own. Against the same published curves the peaks of the proposed method lie $7\%$ to $10\%$ below, close to the $12\%$ by which that inflation raises a peak load at $h=\ell/2$.

The sharp-crack finite element reference of Section~\ref{subsec:sens_fgm} is a separate implementation, with quadratic triangles for the displacement and linear ones for the temperature on a structured mesh of $80$ elements across the specimen, the notch a seam of doubled nodes, the transient conduction advanced by backward Euler with the time step and the loading history of the proposed method, and the crack held at its initial length throughout.

The reference does not use the interaction integral of Section~\ref{subsec:crack_tip_quantities}, and it obtains the SIFs in two steps. First, the potential energy $\Pi$ is computed on the mesh with the notch at its length $a_0$ and on two meshes with the notch extended straight ahead by $\Delta a$ and by $\Delta a/2$, the temperature being transferred to the extended meshes. The energy released per unit extension, $[\Pi(a_0)-\Pi(a_0+\Delta a)]/\Delta a$, approximates the straight-ahead energy release rate $G$, and the two increments are combined by Richardson extrapolation to remove the first-order error of that difference. $G$ gives the magnitude $\sqrt{K_I^2+K_{II}^2}=\sqrt{E'G}$. Second, the ratio of $K_I$ to $K_{II}$ is the ratio of the normal to the tangential opening of the crack faces, each fitted against $\sqrt{r}$ between $2$ and $12$ element sizes behind the tip. The signs follow the Williams convention in a frame with $x$ along the notch toward the tip, which gives $K_{II}>0$ under the applied shear, and $K_I$ takes the sign of the normal opening, negative in a thermal state that closes the faces. From there the two SIFs enter the criterion exactly as in the proposed method, through the kink angle and the energy release rate of the kink $G(\theta_c)$ of Section~\ref{subsec:crack_growth_direction}, compared with $G_c(T_{\mathrm{tip}})$ of \eqref{eq:temperature_dependent_gc} at the temperature of the tip node, which is single in the seam. The load step at which the ratio first reaches unity is the initiation marker of Fig.~\ref{fig:sens_fd}. Extending the notch costs little for the finite element reference, whereas in the proposed method it would add two trainings to every extraction.

The implementation was checked before use. On the stationary thermal edge crack of Section~\ref{subsec:wang_verification}, with $32$, $64$ and $128$ elements across the strip, the same implementation returns $K_I/K_0=0.4808$, $0.4889$ and $0.4928$, which extrapolate to $0.4966$ against the published $0.500$, a difference of $0.7\%$, and the ratio of successive differences is $2.02$, the first-order convergence a crack-tip singularity gives. On the shear specimen its elastic stiffness extrapolates to $88\,802~\mathrm{N/mm}$ from linear elements and to $88\,778$ from quadratic ones, the two agreeing to $0.03\%$. Doubling the $80$ elements of the graded thermal runs leaves the load step at which the crack first advances unchanged and moves the reaction force by $0.1\%$ and $G(\theta_c)$ by $0.3\%$.

The temperature comparison of Section~\ref{subsec:sens_fgm} extends to the crack faces. The jump between the two face temperatures, read one grid spacing either side of the notch, agrees with the reference to $1.4$~K on the faces except over the last crack increment, $0.025$~mm, behind the tip. This region reaches closer to the tip than the pointwise comparison of Section~\ref{subsec:sens_fgm}, which leaves out the last $0.05$~mm. Ahead of the tip, where the exact field is continuous, the same difference read across the extended notch line is $1.9$~K in the trained field against $0.36$~K in the reference within the endpoint overlap $\delta_c$ of Section~\ref{subsec:crack_embedding}, and $0.5$~K against $0.28$~K beyond it. The excess within the overlap is the discontinuity that the overlap prolongs past the tip by construction, and beyond it the trained field is continuous but steeper than the reference.

\section{Stress intensity factor extraction under thermo-mechanical loading}
\label{app:thermal_area_term}

The area term \eqref{eq:thermal_domain_term} of the interaction integral is the interaction form of the area contribution that Wilson and Yu~\cite{wilson1979} derived for the thermoelastic $J$-integral. It has no counterpart in the isothermal contour form, and it is omitted in some works of the literature~\cite{ghaffari2016xfem}, so its effect is examined here.

When the area term is omitted, the extraction does not fail visibly. The SIFs stay finite and drift with the contour radius, which is the dependence that the radius sweep of Section~\ref{subsec:extraction_admissibility} measures. The area integral carries no weight function. An evaluation in domain form multiplies the integrand by a ramp that falls to zero on the outer boundary of the integration domain, whereas the quantity needed here is the difference between a vanishing contour and $\Gamma_r$, which is the integral over the enclosed disc with unit weight.

\begin{figure}[htbp]
  \centering
  \includegraphics[width=0.62\textwidth]{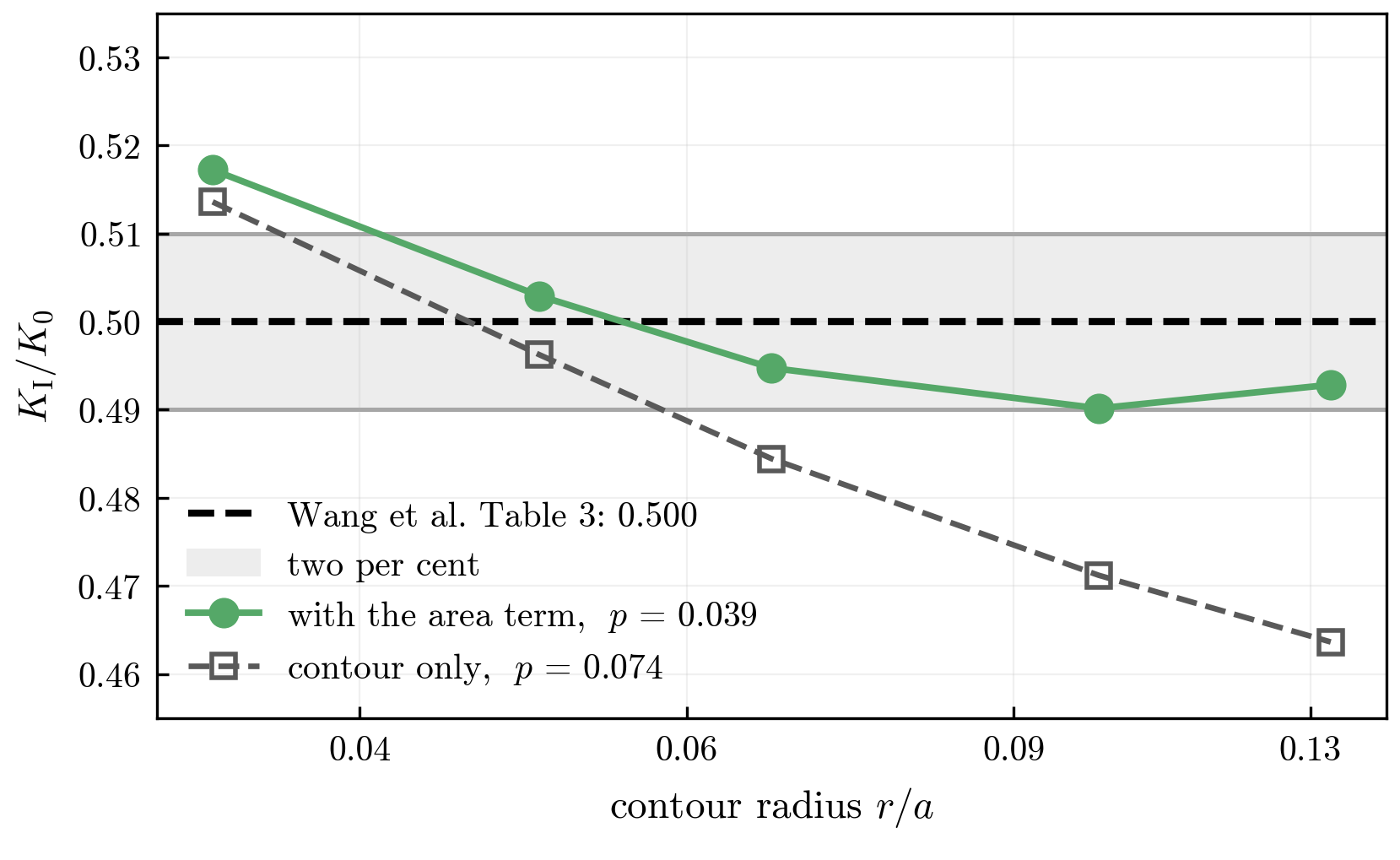}
  \caption{The area term isolated on the stationary benchmark. A field trained separately on the finest grid is read with the area term \eqref{eq:thermal_domain_term} and from the contour alone. With the area term the sweep flattens inside the band, and from the contour alone the reading falls monotonically. The dashed line is the published value of Wang et al.~\cite{wang2023cmt}, the shaded band 2\%.}
  \label{fig:mth_ablation}
\end{figure}

The effect of the area term was isolated on the stationary benchmark of Section~\ref{subsec:wang_verification}. A field trained separately on the finest grid was read twice, with the area term and from the contour alone, and Fig.~\ref{fig:mth_ablation} shows both sweeps. With the area term the sweep flattens at $K_I/K_0=0.495$, $1.1\%$ below the published value, with $p=0.039$. Its difference from the reading of Section~\ref{subsec:wang_verification} on the same grid lies within the $1.2\%$ spread over initial enrichment amplitudes reported in Appendix~\ref{app:discretization}. From the contour alone the reading falls monotonically, by $9.7\%$ from the innermost contour to the outermost, and the value reduced from the sweep drops to $0.484$ with $p=0.074$. The two readings agree to $0.7\%$ on the smallest contour and separate to $5.9\%$ on the largest, since the omitted quantity is an area integral and grows with the contour. On this benchmark the contour-only reading still passes the admissibility check, and the shape of the sweep, flat or monotonic, is what separates the two. Where the thermal load contributes more to the SIFs the difference is larger. On the trained field of the thermal case of the cruciform of Section~\ref{subsec:cruciform} at mid load, taken from the reported run, the contour-only reading gives $p=0.54$, more than four times the threshold, and the reading with the area term $0.18$. The run itself recorded $p=0.176$ at that step and did not advance the crack.

\bibliographystyle{unsrt}
\bibliography{Bib}

\end{document}